\documentclass{article}

\PassOptionsToPackage{table}{xcolor}

\usepackage[preprint]{neurips_2025}

\usepackage[utf8]{inputenc} 
\usepackage[T1]{fontenc}    
\usepackage[colorlinks,linkcolor=blue,urlcolor=blue,citecolor=blue]{hyperref}
\usepackage{url}            
\usepackage{booktabs}       
\usepackage{amsfonts}       
\usepackage{nicefrac}       
\usepackage{microtype}      

\usepackage{xspace}
\usepackage{stmaryrd}
\usepackage{amsthm}
\usepackage{amsmath}
\usepackage{amssymb}
\usepackage{mathtools}
\usepackage{times}
\usepackage[pdftex]{graphicx}
\graphicspath{{figures/}}
\usepackage{wrapfig}
\usepackage{subcaption}
\usepackage{multirow}
\usepackage{multicol}
\usepackage{makecell}

\usepackage{tabularx} 

\usepackage{ragged2e} 
\usepackage{algorithm}
\usepackage{algorithmic}
\usepackage{arydshln}
\usepackage{threeparttable}
\usepackage[most]{tcolorbox}

\newcommand{\repourl}{\url{https://github.com/LinXueyuanStdio/RETuning}}

\newcommand{\huggingfaceurl}{\url{https://huggingface.co/collections/linxy/retuning-68c999be4d9ac2834c64fd00}}

\newcommand{\dataseturl}{\url{https://huggingface.co/datasets/linxy/Fin-2024}}

\usepackage{amsmath,amsfonts,bm}

\def\eqref#1{equation~\ref{#1}}

\def\1{\bm{1}}

\DeclareMathAlphabet{\mathsfit}{\encodingdefault}{\sfdefault}{m}{sl}
\SetMathAlphabet{\mathsfit}{bold}{\encodingdefault}{\sfdefault}{bx}{n}

\definecolor{Primary1}{RGB}{106,62,171} 
\definecolor{Primary2}{RGB}{244,244,249} 

\definecolor{Secondary1}{RGB}{179,167,225} 
\definecolor{Secondary2}{RGB}{199,179,230} 
\definecolor{Secondary3}{RGB}{50,43,68} 

\definecolor{Highlight1}{RGB}{232,183,76} 
\definecolor{Highlight2}{RGB}{176,143,243} 
\definecolor{Highlight3}{RGB}{126,162,255} 

\definecolor{Background1}{RGB}{241,230,247} 
\definecolor{Background2}{RGB}{44,28,53} 

\definecolor{Font1}{RGB}{44,28,53} 
\definecolor{Font2}{RGB}{244,236,232} 

\definecolor{Best1}{RGB}{198,70,17} 
\definecolor{Best2}{RGB}{228,163,4} 
\definecolor{Best3}{RGB}{84,130,53} 

\definecolor{RED}{RGB}{198,70,17}
\definecolor{GREEN}{RGB}{84,130,53}
\definecolor{GREY}{RGB}{128,128,128}

\definecolor{golden}{RGB}{255, 215, 0} 
\definecolor{lightblue}{RGB}{147, 170, 177}  
\definecolor{lightgreen}{RGB}{182, 124, 98}  

\hypersetup{
 colorlinks=False,
 linkcolor=blue,
 citecolor=blue,
 urlcolor=blue}

\definecolor{Green}{rgb}{0.0, 0.2, 1.0}
\definecolor{Red}{rgb}{0.8, 0.0, 0.0}
\newtcolorbox{takeaway}{
  colback=white,
  colframe=green!75!black,
  fonttitle=\bfseries,
  title=,
  boxrule=1pt,
  arc=4pt,
  auto outer arc,
  width=\linewidth,
  breakable,
}
\newcommand{\hidecomments}[1]{}

\title{\textcolor{Primary1}{RETuning}: Upgrading Inference-Time Scaling for Stock Movement Prediction with Large Language Models}
\author{
  Xueyuan Lin$^{1,2,3}*$ \quad Cehao Yang$^{1,2}*$ \quad Ye Ma$^{3}$ \quad Ming Li$^{3}$ \quad Rongjunchen Zhang$^{3}$ \\
  \textbf{Yang Ni$^{1}$ \quad Xiaojun Wu$^{1,2}$ \quad Chengjin Xu$^{2,4}$ \quad Jian Guo$^{2}\dagger$ \quad Hui Xiong$^{1}\dagger$} \\
  $^1$ The Hong Kong University of Science and Technology (Guangzhou)\\
  $^2$ IDEA Research \quad $^3$ Hithink RoyalFlush Information Network Co., Ltd \quad $^4$ DataArc Tech Ltd\\
  \small\texttt{\{xlin058,cyang289,yni002\}@connect.hkust-gz.edu.cn, xionghui@ust.hk} \\
  \small\texttt{\{maye,zhangrongjunchen\}@myhexin.com, lm92@mail.ustc.edu.cn}\\
  \small\texttt{\{xuchengjin,wuxiaojun,guojian\}@idea.edu.cn}
}

\begin{document}

\maketitle
\footnotetext[1]{Equal contribution.}
\footnotetext[2]{Corresponding Authors.}

\begin{abstract}
Recently, large language models (LLMs) have demonstrated outstanding reasoning capabilities and inference-time scaling on mathematical and coding tasks. However, their application to financial tasks—especially the most fundamental task of stock movement prediction—remains underexplored. We study a three-class classification problem (up, hold, down) and, by analyzing existing reasoning responses, observe that:
(1) LLMs are easily swayed by contextual viewpoints, tending to follow analysts' opinions rather than exhibit a systematic, independent analytical logic in their chain-of-thoughts (CoTs).
(2) LLMs often list summaries from different sources without weighing adversarial evidence, yet such counterevidence is crucial for reliable prediction.
It shows that the model does not make good use of its reasoning ability to complete the task.
To address this, we propose \textbf{R}eflective \textbf{E}vidence \textbf{}uning (\textbf{RETuning}), a cold-start method prior to reinforcement learning, to enhance prediction ability. While generating CoT, \textbf{RETuning} encourages dynamically constructing an analytical framework from diverse information sources, organizing and scoring evidence for price up or down based on that framework—rather than on contextual viewpoints—and finally reflecting to derive the prediction. This approach maximally aligns the model with its learned analytical framework, ensuring independent logical reasoning and reducing undue influence from context.
We also build a large-scale dataset spanning all of 2024 for 5,123 A-share stocks, with long contexts (32K tokens) and over 200K samples. In addition to price and news, it incorporates analysts' opinions, quantitative reports, fundamental data, macroeconomic indicators, and similar stocks.
Experiments on this new dataset show that, as a cold-start method, \textbf{RETuning} successfully unlocks the model's reasoning ability in the financial domain. During reinforcement learning, response length steadily increases under the designed curriculum setting. Furthermore, inference-time scaling still works even after 6 months or on out-of-distribution stocks, since the models gain valuable insights about stock movement prediction.
\end{abstract}

\section{Introduction}\label{sec:introduction}
\textbf{Stock Movement Prediction (SMP)} is one of the most fundamental and consequential tasks in finance. It not only directly affects the interests of individual investors but also plays a central role in algorithmic trading~\citep{Mahfooz2022ImprovingST,Ta2018PredictionAP}, financial risk control~\citep{Adyatma2022TheIS,Vui2013ARO}, and intelligent research platforms~\citep{Shi2020StockMP}. In recent years, Large Language Models (LLMs)~\citep{brown2020language,Achiam2023GPT4TR,Qwen,LLaMA} have demonstrated remarkable reasoning capabilities in domains such as code generation and mathematical problem-solving~\citep{deepseek-r1,o1,uesato2022solving}. This has sparked growing interest in exploring whether such models can also excel in financial tasks. However, it remains an open question whether LLMs' strength in \textit{reasoning} and \textit{inference-time scaling} can be effectively harnessed for stock price prediction.

In the traditional planning phase of a financial agent, the model has access to a wide range of information sources—news articles, analyst opinions, research reports, quantitative factor analyzes, and more. Despite this richness, making reliable and interpretable predictions remains a major challenge. On one hand, \textcolor{Primary1}{LLMs often exhibit strong prior biases due to the optimistic slant of their training data}, which is skewed toward long positions and excludes contrarian views for political or regulatory reasons. On the other hand, these models tend to \textcolor{Primary1}{lack the ability to construct independent reasoning frameworks, reconcile conflicting information, and perform reflective analysis}—capabilities that are essential for robust financial decision-making.

To address these challenges, we propose a novel modeling paradigm that treats stock movement prediction as a \textit{generative reasoning task}. By processing all textual information sources end-to-end, this approach aims to simulate the thought process of a human trader, ultimately generating structured and interpretable predictions. Two key innovations underpin this paradigm.

First, we introduce \textbf{Reflective Evidence Tuning (RETuning)}, which instills LLMs to dynamically construct reasoning frameworks based on diverse information sources, collect and evaluate evidence for potential price directions (up, down, or hold), and reflect on the evidence before making a final prediction. This structured approach is a cold-start training mechanism prior to reinforcement learning (RL). It enables models to avoid merely summarizing or echoing external viewpoints and instead follow an internally consistent logic, improving both interpretability and accuracy.

Second, we explore the role of \textbf{inference-time scalability}, a technique that has shown promise in mathematical and programming tasks~\citep{s1,S*}. Specifically, we investigate whether \textit{majority voting} can significantly improve predictive accuracy in financial domain. Although widely successful elsewhere, its efficacy in stock movement prediction has not yet been systematically examined.

To support this research, we construct a large-scale, high-quality dataset that reflects the complexity and information density of real-world financial environments. Covering the full year of 2024 across over 4,000 A-share stocks, this dataset integrates six heterogeneous information sources: news, fundamentals, analyst opinions, quantitative factor reports, macroeconomic context, and stocks of similar trends. With over 200,000 samples and an average input length of up to 32K tokens, it overcomes the limitations of prior datasets that were outdated and lacked information diversity~\citep{StockNet,CMIN,EDT}. The details of the dataset construction are discussed in Appendix~\ref{sec:appendix:dataset_details}.

Empirical results show that \textbf{RETuning} effectively enhances reasoning structure and improves predictive performance over strong baselines. It also generalizes beyond stock movement prediction, yielding significant improvements in other financial tasks, and demonstrates strong performance under inference-time scaling and out-of-distribution settings.

Our contribution can be summarized as follows:
(1) We build a large-scale, long-context financial dataset with diverse evidence sources beyond price and news, which fill the gap that existing datasets are outdated and lack information diversity.
(2) We introduce \textbf{RETuning}, synthesizing cold-start responses that guide LLMs to construct and reflect on an analytical framework for stock movement prediction. It allows significant inference-time scalability of LLMs in the prediction task.
(3) We empirically show that \textbf{RETuning} unlocks prediction ability and generalizes beyond stock movement prediction. We believe this research lays the groundwork for deploying trustworthy, reasoning-driven LLMs in real-world financial applications.

\section{Related Work}\label{sec:relatedwork}

\paragraph{Stock Movement Prediction with LLMs.}
Recent work has focused on exploring various types of information sources for stock movement prediction with LLMs. Several studies emphasize the importance of stock-related news in revealing fundamental market insights~\citep{vargas2018deep,li2021modeling}. Meanwhile, other research highlights the significance of understanding the relationships between companies and industries~\citep{feng2019temporal, hsu2021fingat}. Recent studies have also provided empirical evidence of the impact of public sentiment on market trends, with researchers working to extract sentiment and keyphrases from news and social media data~\citep{nguyen2015sentiment, hao2021predicting}. However, these works often focus on a single type of information source, such as news or sentiment, and do not fully leverage the potential of LLMs to integrate multiple heterogeneous sources. In contrast, our work aims to construct a comprehensive dataset that incorporates diverse information sources, including news, fundamentals, analyst opinions, quantitative reports, macroeconomic indicators, and similar stocks, to enhance the predictive capabilities of LLMs in stock movement prediction.

\paragraph{Inference-Time Scaling for LLMs.}
Inference-time scaling methods~\citep{snell2024scaling} enhance LLM performance by leveraging additional computation during generation, broadly categorized into three strategies. \textbf{Repeated sampling} improves diversity and accuracy via parallel candidate generation, utilizing verification strategies like majority voting~\citep{li2024agentsneed,lin-etal-2024-just,wang2023selfconsistency,toh2024votescountprogramsverifiers} or best-of-N (BoN) selection with verifiers~\citep{stiennon2020learning,cobbe2021trainingverifierssolvemath,nakano2022webgptbrowserassistedquestionansweringhuman,li-etal-2023-making,liu2025pairwise}, while efficiency optimizations prune low-scoring paths early~\citep{zhang2024accelerating,qiu2024treebonenhancinginferencetimealignment,sun2024fast,manvi2024adaptiveinferencetimecomputellms,ye-ng-2024-preference}. \textbf{Self-correction} iteratively refines outputs using feedback from tools, external models, or self-critique~\citep{shinn2023reflexionlanguageagentsverbal,gou2024critic,li2024confidencemattersrevisitingintrinsic,song2025progcoprogramhelpsselfcorrection}, though its efficacy depends on feedback reliability~\citep{olausson2024is,huang2024large,wang2024reasoningtokeneconomiesbudgetaware,yang2024confidencevscritiquedecomposition}. \textbf{Tree searching} combines parallel and sequential scaling via algorithms like MCTS or A*~\citep{NEURIPS2023_271db992,xie2023selfevaluation,long2023largelanguagemodelguided,chari2025pheromonebasedlearningoptimalreasoning} guided by value functions~\citep{xu2023traingainunleashmathematical,hao-etal-2023-reasoning,chen2024alphamathzeroprocesssupervision,zhang2024llamaberrypairwiseoptimizationo1like}. Training techniques distill these scaling benefits into more efficient models~\citep{gao2023scaling,hou2024doesrlhfscaleexploring,gulcehre2023reinforcedselftrainingrestlanguage,zhang2024o1codero1replicationcoding}.
However, these methods have not been systematically applied to financial tasks, particularly stock movement prediction. Our work aims to fill this gap by exploring how inference-time scaling can be effectively utilized in this domain.

\section{Preliminaries}\label{sec:preliminary}

\subsection{Strict Controlled Dataset for Stock Movement Prediction}\label{sec:preliminary:1}

\begin{figure*}
  \centering
  \includegraphics[width=0.95\textwidth]{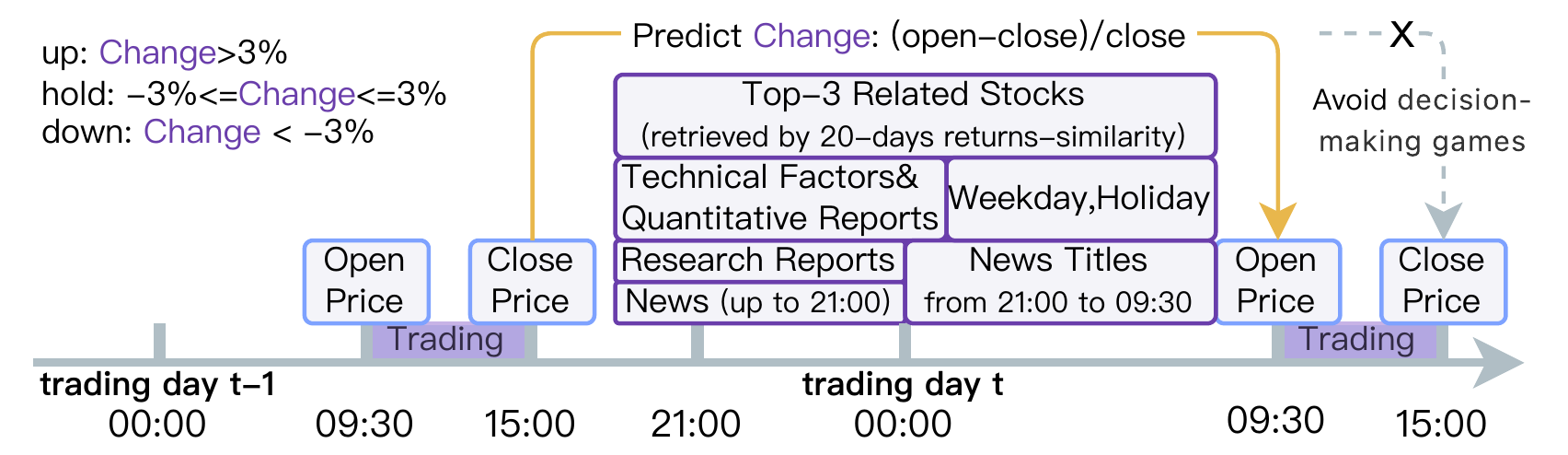}
  \caption{Overview of data sample for one stock at trading day t.}
  \label{fig:dataset}
\end{figure*}

We aim to evaluate the ability of LLMs to predict stock movements based on diverse information sources. To this end, we construct a strict controlled dataset named \textbf{Fin-2024}, which covers the entire year of 2024, including 5,123 A-share stocks and 209,063 samples. Each sample is designed with a long context window of 32K tokens. Figure~\ref{fig:dataset} illustrates the data sample for one stock at trading day $t$. LLMs will be trained from January to November, and then evaluated on December. In addition, we also collect data \textbf{Fin-2025[June]} on June 2025 for long-horizon prediction evaluation. The dataset construction process is detailed in Appendix~\ref{sec:appendix:dataset_details}.



\paragraph{Information Sources}
The dataset consists of diverse information sources that have been proven valuable in machine learning-based quantitative trading research. These include: (1) \textcolor{Primary1}{\textbf{News articles}} providing real-time market updates and company-specific \& sector-specific information, (2) \textcolor{Primary1}{\textbf{Fundamental reports}} reflecting company financial health and performance, (3) \textcolor{Primary1}{\textbf{Analyst opinions}} offering professional market insights, (4) \textcolor{Primary1}{\textbf{Quantitative reports}} containing technical analysis and market indicators, (5) \textcolor{Primary1}{\textbf{Macroeconomic indicators}} showing broader economic trends, and (6) \textcolor{Primary1}{\textbf{Similar stocks information}} for comparative analysis. The information are primarily textual, friendly for LLMs.

\paragraph{Prediction Target}
We define the prediction target as the price movement between the current trading day's opening price and the previous trading-day's closing price. This setting is less common in pre-training data compared to closing price-based movements, which helps prevent the model from exploiting memorization.
The setting also avoids the model from decision-making games in trading periods, which is hard to capture in the given context.
Based on price change, we classify the stock movement into three classes: \textbf{up} for change > 3\%, \textbf{down} for < -3\%, and \textbf{hold} for else.
The three-class classification scheme requires more significant signals for price movements than binary classification.
The \textbf{hold} class serves as a \textcolor{Primary1}{decoy}.
If the model never learns to distinguish between \textbf{up} and \textbf{down}, it would indicate a shortage on the model's ability to make price movement predictions.

\paragraph{Evaluation Protocol}
During evaluation, we require the model to simultaneously predict both the price change percentage and direction to assess its instruction-following capability and verify the consistency between the two predictions, as the direction should align with the predicted change percentage.

\subsection{Observation on Existing Models}\label{sec:preliminary:2}


\begin{minipage}[t]{0.49\textwidth}
    \centering
    \includegraphics[width=\textwidth]{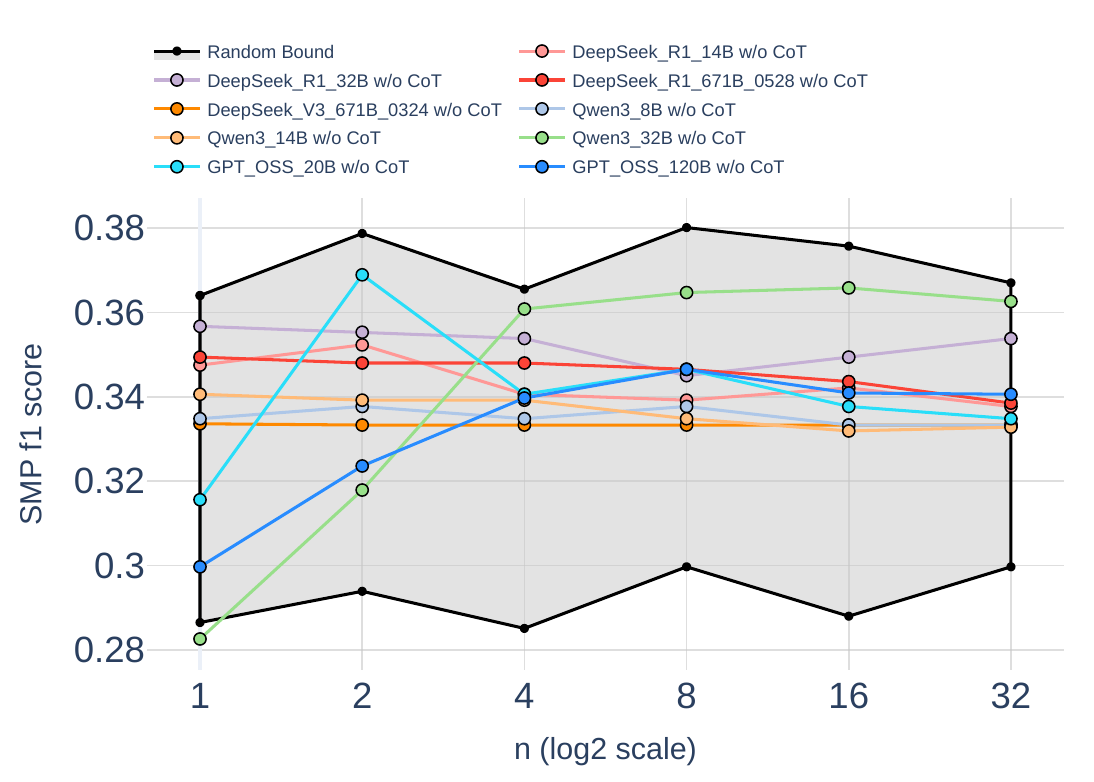}
    \captionof{figure}{Existing LLMs without RETuning CoT perform no better than random guessing in stock movement prediction, and most (except Qwen3 32B) fail to scale at inference time.}
    \label{fig:baselines}
\end{minipage}
\hfill
\begin{minipage}[t]{0.49\textwidth}
    \centering
    \includegraphics[width=\textwidth]{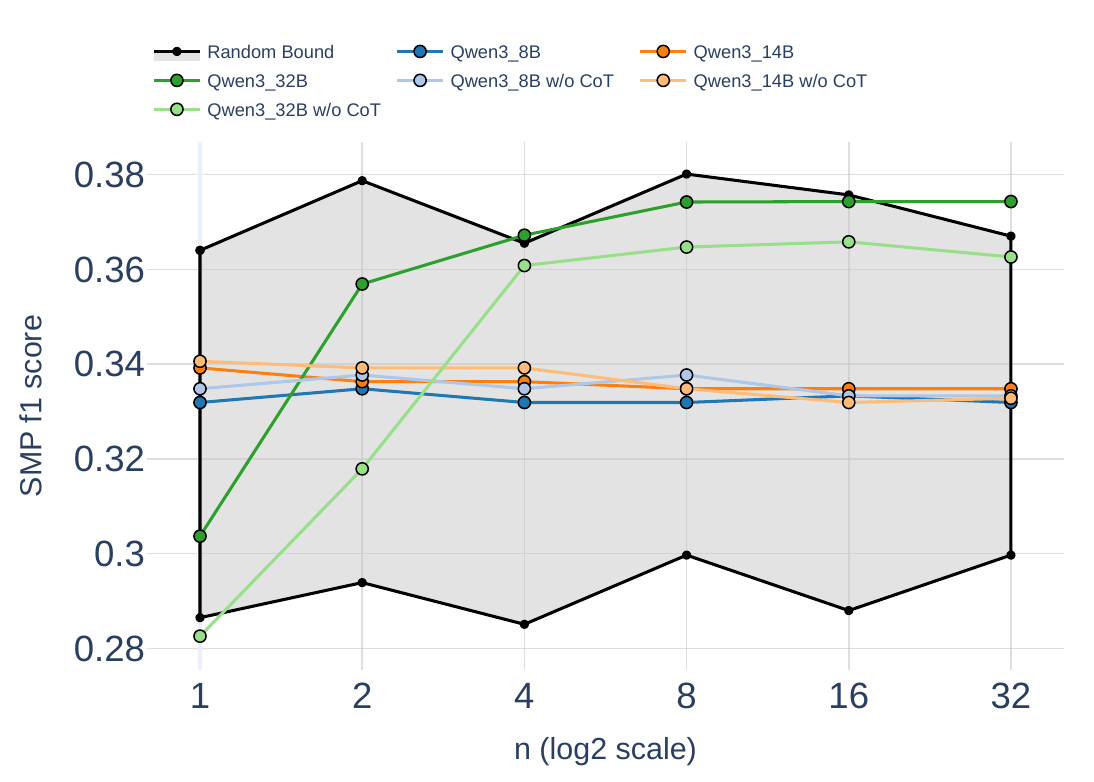}
    \captionof{figure}{Abaltion on CoT prompting. The proposed CoT in Section~\ref{sec:method:1} can help improve the prediction performance on Qwen3 32B, but not smaller models. Others lose to random guessing.}
    \label{fig:ablation_on_CoT}
\end{minipage}

To understand the limitations of existing models in stock movement prediction, we analyze their performance on \textbf{Fin-2024[December]} in Section~\ref{sec:preliminary:1}. We use 32 different random seeds to uniformly sample the prediction (hold, up, and down) to construct the random bound (grey area in the figure). Any results that are covered by the random bound will be regarded failing to make trust-worthy prediction. The results reveal two key issues:

Firstly, by investigating existing LLMs' performance in Figure~\ref{fig:baselines}, we observe that current LLMs are almost randomly guessing the prediction result. And the most models cannot scale their ability of prediction at inference time.
Secondly, we prompt the model with fine-grained CoT (Section~\ref{sec:method:1}) to inject knowledge of financial analysis into the reasoning process. We further analyze the responses of the models, as shown in Figure~\ref{fig:ablation_on_CoT}. We find that CoT can help improve the model's prediction performance on Qwen3 32B, but not other models.

To understand why these models fail, we further sample multiple responses for case study (Appendix~\ref{sec:appendix:case_study:cot_prompting}) and find that:
the outputs of these LLMs tend to be vague, detached from the prediction target, and biased toward the \texttt{hold} class on label-balanced datasets.
Thus, we propose to utilize \textcolor{Primary1}{supervised fine-tuning} to induce coherent, task-specific reasoning to change the output distribution of these models, as discussed in the following section.

\section{Reflective Evidence Tuning (RETuning)}\label{sec:method}

\begin{figure*}
	\centering
	\includegraphics[width=0.95\textwidth]{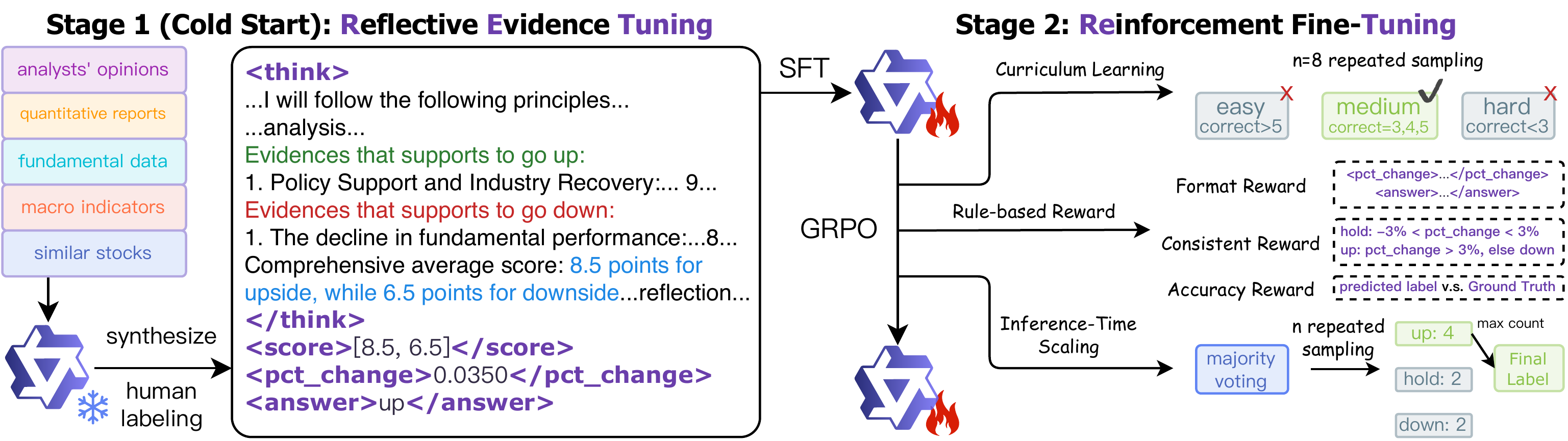}
	\caption{Two-stage stock movement prediction model training framework: \textbf{Stage 1} (Cold Start) uses multi-source data with human labeling and synthesis pipeline for \textcolor{Primary1}{R}eflective \textcolor{Primary1}{E}vidence \textcolor{Primary1}{Tuning}; \textbf{Stage 2} applies \textcolor{Primary1}{Re}inforcement Fine-\textcolor{Primary1}{Tuning} with curriculum learning, reward shaping, and inference-time scaling for final label determination.}
	\label{fig:method}
\end{figure*}

This section introduces \textbf{Reflective Evidence Tuning (RETuning)}, a two-stage framework designed to unlock the latent reasoning ability of LLMs in stock movement prediction tasks. As illustrated in Figure~\ref{fig:method}, RETuning comprises: (1) an SFT stage to \textit{cold-start generative reasoning modeling}, and (2) \textit{rule-based reinforcement learning} for performance refinement and alignment.

\subsection{Stock Movement Prediction via Generative Reasoning Modeling}\label{sec:method:1}




\begin{wrapfigure}{r}{0.38\textwidth}
  \vspace{-15pt}
  \centering
  \begin{subfigure}[t]{0.37\textwidth}
    \centering
    \includegraphics[width=\textwidth]{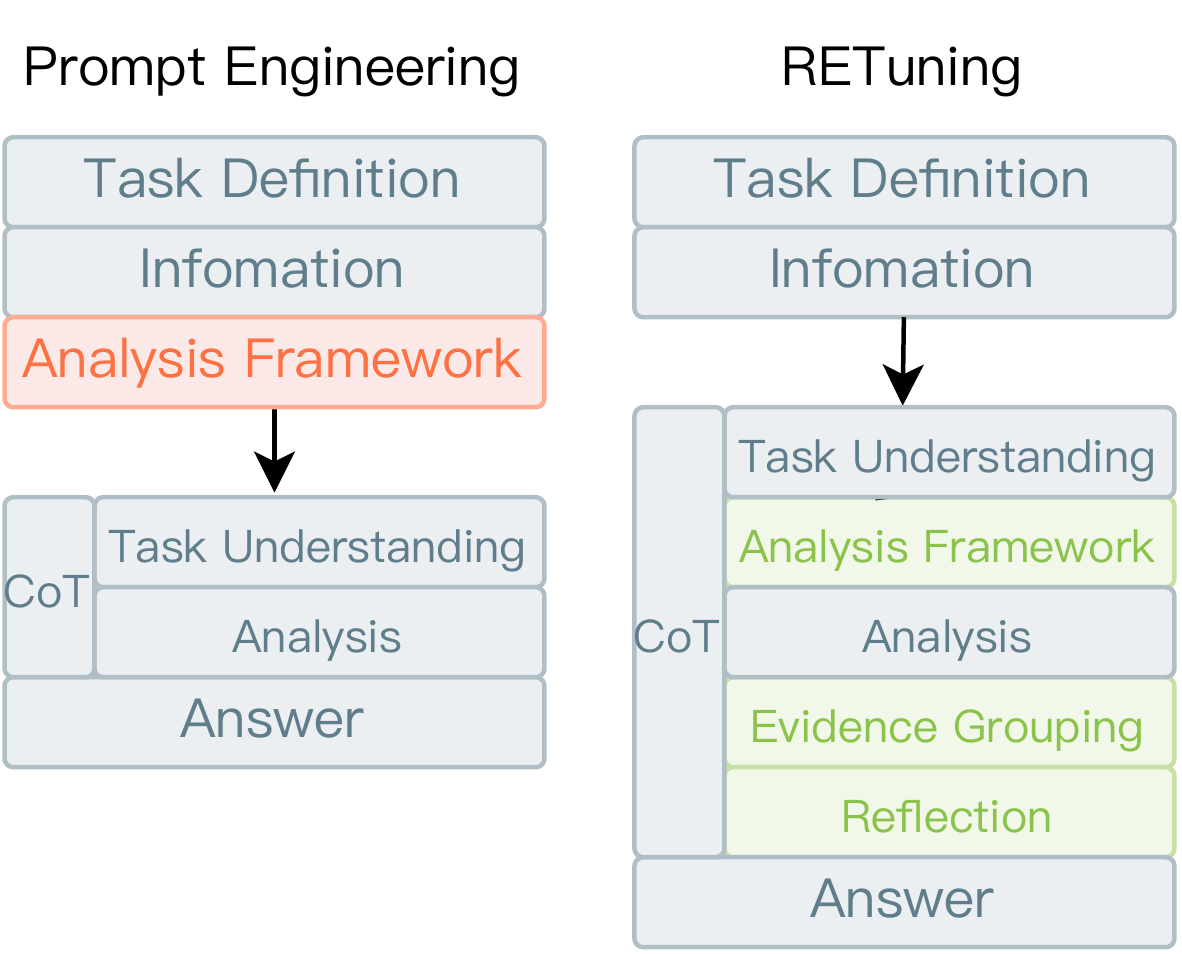}
    \caption{\textbf{RETuning} guides the model to generate a principle, collect evidence, and reflect before making a prediction.}
    \label{fig:data_synthesis}
  \end{subfigure}


  \begin{subfigure}[t]{0.37\textwidth}
    \centering
    \includegraphics[width=\textwidth]{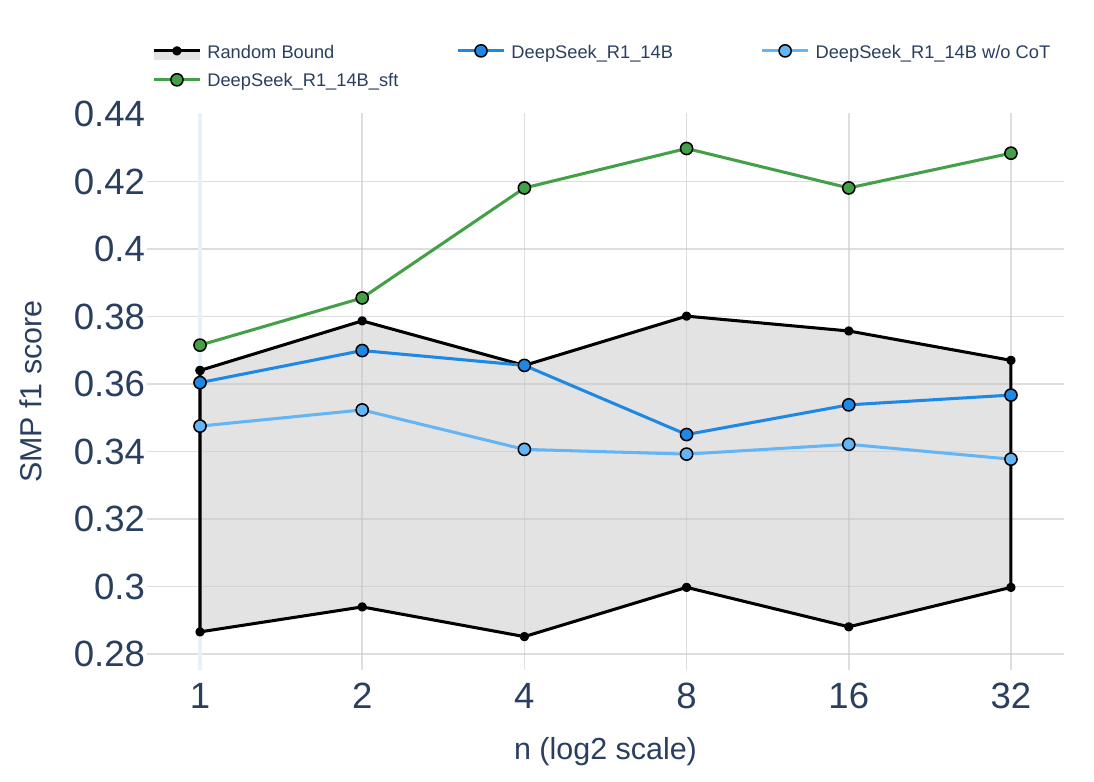}
    \caption{DeepSeek\_R1\_14B\_SFT scales prediction performance via repeated sampling.}
    \label{fig:scaling_on_prediction_performance}
  \end{subfigure}
  \vspace{-25pt}
\end{wrapfigure}

We frame the SMP task as a \textcolor{Primary1}{\textit{generative reasoning problem}}, in which the LLM leverage its reasoning ability to make predictions. It must construct an analytical framework, extract and score evidence from heterogeneous sources, and reflect before reaching a conclusion. As is shown in Figure~\ref{fig:data_synthesis}, this contrasts with zero-shot settings, where models superficially summarize and avoid making grounded predictions.

Then we employ supervised fine-tuning (SFT) to instill this reasoning structure into the model. The SFT dataset is constructed through a semi-automated pipeline (Appendix~\ref{sec:appendix:sft_data_synthesis}) that uses DeepSeek-R1 (671B) as the backbone model to rejected sampling to synthesize 118 golden cold-start items.

By grounding each prediction in a dynamically built framework, \textbf{RETuning} promotes robust and context-aware reasoning. This reduces susceptibility to dominant context bias and improves the model’s ability to rationally weigh adversarial evidence—crucial for reliable financial forecasting.
By fine-tuning on this structured reasoning process, we find that the model \textcolor{Primary1}{DeepSeek\_R1\_14b\_SFT can scale its prediction performance via repeated sampling far beyond random guessing}, as shown in Figure~\ref{fig:scaling_on_prediction_performance}. It indicates that the model already possesses a certain level of predictive capability. We can further leverage this weak predictive power to assess the difficulty of predicting samples, thereby enhancing the overall predictive performance of the model more efficiently.

\subsection{Rule-based Reinforcement Learning}\label{sec:method:2}
To further align model outputs with desired reasoning behavior, we introduce a rule-based reinforcement learning (RL) stage. Rather than relying on simple correctness-based rewards—which are noisy and statistically uninformative in financial prediction—we design more principled signals through \textbf{reward shaping} and \textbf{curriculum learning}.

\paragraph{Reward Shaping}




We design a \textbf{multi-faceted reward function} to capture both the \textcolor{Primary1}{structural} and \textcolor{Primary1}{semantic} correctness of model outputs.
First, a \textbf{\underline{Format Score}} ensures that the response adheres to the expected structured format, maintaining clarity and consistency.
The \textbf{\underline{Accuracy Score}} focuses on whether the model correctly predicts the directional movement—\texttt{up}, \texttt{down}, or \texttt{hold}.
Lastly, the \textbf{\underline{Consistency Score}} encourages logical alignment between the predicted percentage change and the stated directional label. Final score is given by $ R = \alpha \cdot \text{Format} + \beta \cdot \text{Accuracy} + \gamma \cdot \text{Consistency} $, where $\alpha, \beta, \gamma$ are hyperparameters.
This design mitigates the issue of misleading signals from noisy.

\paragraph{Curriculum Learning}








\begin{wrapfigure}{r}{0.45\textwidth}
  \vspace{-10pt}
  \centering
  \includegraphics[width=0.43\textwidth]{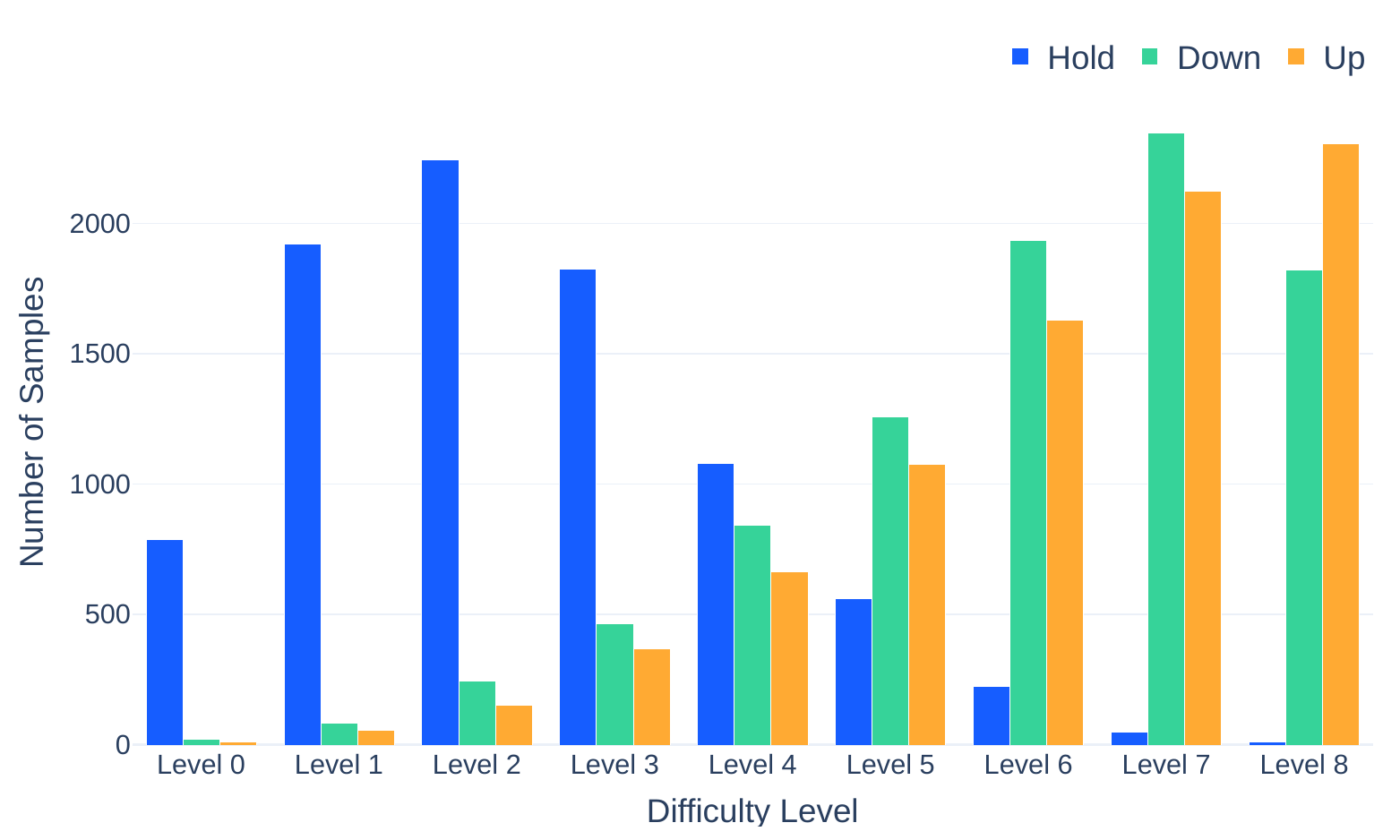}
  \caption{Difficulty distribution given by the cold-started model (DeepSeek\_R1\_14B\_SFT).}
  \label{fig:difficulty_distribution}
  \vspace{-20pt}
\end{wrapfigure}

Not all samples are equally difficult: \textbf{hold} predictions are often trivial, while confident \textbf{up/down} predictions require strong signal integration. To make training more efficient and targeted, we propose a curriculum learning strategy:

We use \textcolor{Primary1}{the cold-started model} to generate $N$ (=8 in practice) predictions for each training sample. The difficulty of a sample is measured by counting how many of these predictions are incorrect. Based on this difficulty score, we categorize examples into three groups: \textbf{easy} (correct $\in [\frac{2}{3}N, N)$ ), \textbf{medium} (correct $\in [\frac{1}{3}N, \frac{2}{3}N)$ ), and \textbf{hard} (correct $\in [0, \frac{1}{3}N)$ ).

In Figure~\ref{fig:difficulty_distribution}, we observe that a clear correlation between difficulty levels and labels.
Low-difficulty samples are mostly dominated by \textbf{hold} predictions, which tend to be either spurious or too simple to be informative. High-difficulty samples, on the other hand, often involve \textbf{up} or \textbf{down} predictions but with weak or noisy signals. In contrast, medium-difficulty samples tend to reflect realistic market complexities and require non-trivial reasoning.
To ensure the model focuses on meaningful learning signals, we discard both low and high-difficulty examples and train only on medium-difficulty ones, progressing in order of increasing difficulty.

\paragraph{Inference-Time Scaling}
We apply \textcolor{Primary1}{majority voting} on predicted labels over $n$ repeated generations with temperature 0.6. The final decision is given by: $ \hat{y} = \arg\max_{y \in \{\texttt{up}, \texttt{down}, \texttt{hold}\}} \sum_{i=1}^{n} \mathbf{1}[y_i = y] $

\section{Experiment and Results}\label{sec:exp}





We conduct several experiments to ascertain the effectiveness of \textbf{RETuning}, with the aim to gain insights into the following:
(1) Basically, can we improve the stock movement prediction performance of LLMs? How does \textbf{RETuning} compare with existing methods? What insights does the model learn from the data?
(2) Can we scale the prediction performance of LLMs at inference time?
(3) What are the key factors that contribute to the success of \textbf{RETuning}?
(4) Does the enhanced prediction ability contribute to other financial tasks?

\subsection{Experiment Setup}\label{sec:exp:1}
\paragraph{Datasets.}
We use the data from January to November in \textbf{Fin-2024} for training, and the December data \textbf{Fin-2024[December]} for testing. We also use the \textbf{Fin-2025[June]} dataset to evaluate whether the model persists its scaling ability on prediction performance after 6 months. Besides, we evaluate the generalization ability on \textbf{BizFinBench}~\citep{lu2025bizfinbench0}, a comprehensive financial benchmark covering 10 tasks, including Anomalous Event Attribution (AEA), Financial Time Reasoning (FTR), Financial Tool Usage (FTU), Financial Numerical Computation (FNC), Financial Knowledge QA (FQA), Financial Data Description (FDD), Emotion Recognition (ER), Stock Price Prediction (SP), Financial Named Entity Recognition (FNER). The details of the datasets are shown in Appendix~\ref{sec:appendix:dataset_details}.

\paragraph{Evaluation and Metrics.}
We care about the generalization ability and consider three types of out-of-distribution (OOD) settings: \textbf{OOD\_Stock}, \textbf{OOD\_Date}, and \textbf{OOD\_Stock\&Date}. We choose 50 stocks in random as the OOD stocks and the last month of 2024 as the OOD dates. The 50 stocks are also OOD stocks in \textbf{Fin-2025[June]}.
We adopt the standard metrics \textcolor{Primary1}{F1-score} because it balances precision and recall, making it suitable for our multi-class classification task.

\paragraph{Implementation and Baselines.}
The models are trained on up to 4*8 H100 GPUs. Rollout $n$ is set to $8$.
Default results are obtained by greedy decoding.
For inference-time scaling, we use $k\in\{1,2,4,8,16,32\}$ and temperature=$0.6$.
More implementation details are shown in Appendix~\ref{sec:appendix:implementation_details}.
Based on DeepSeek\_R1\_14B\_Instruct (originally DeepSeek-R1-Distill-Qwen-14B~\citep{deepseek-r1}), we apply \textbf{RETuning} and get DeepSeek\_R1\_14B\_SFT and DeepSeek\_R1\_14B\_SFT\_GRPO, which are after SFT stage and after SFT + GRPO stages, respectively. We also implement DeepSeek\_R1\_32B\_SFT and DeepSeek\_R1\_32B\_SFT\_GRPO based on DeepSeek\_R1\_32B\_Instruct.
We compare to several strong baselines, including:
LLMFactor~\citep{wang-etal-2024-llmfactor}, Fino1~\citep{Fino1}, Fin-R1~\citep{Fin-R1}, CMIN~\citep{CMIN} and StockNet~\citep{StockNet}.
We also report results of several state-of-the-art open-weight LLMs:
DeepSeek~\citep{deepseek-r1} (R1-7B, R1-14B, R1-32B, R1-671B, V3-671B), Qwen3~\citep{yang2025qwen3technicalreport} (8B, 14B, 32B), GPT-OSS~\citep{openai2025gpt0oss0120b} (20B, 120B).

\subsection{Results and Analysis}\label{sec:exp:2}

\begin{minipage}[b]{0.49\textwidth}
    \centering
    \captionof{table}{Results of different methods on \textbf{Fin-2024[December]} benchmarks. w/ CoT means using the CoT prompting in Section~\ref{sec:preliminary:2}. The relative improvements (\%) over the baselines are shown in parentheses. The best results are in \textbf{bold}.}
    \resizebox{\linewidth}{!}{
        \begin{tabular}{ll}
        \toprule
        \textbf{Model} &  \textbf{F1 Score} \\
        \midrule\midrule
        \multicolumn{2}{c}{\textit{Results of Public Models}} \\
        Random Guessing       & 0.3333              \\
        LLMFactor~\citep{wang-etal-2024-llmfactor} & 0.3345              \\
        Fino1~\citep{Fino1}          & 0.0622              \\
        Fin-R1~\citep{Fin-R1}          & 0.2543              \\
        CMIN~\citep{CMIN}           & 0.3275              \\
        StockNet~\citep{StockNet}        & 0.3081              \\
        \midrule
        \multicolumn{2}{c}{\textit{Results with CoT Ablation}} \\
        Qwen3\_8B~\citep{yang2025qwen3technicalreport}       & 0.3348                   \\
        $\quad$ w/ CoT  & 0.3319 (\textcolor{RED}{-0.87\%}) \\
        Qwen3\_14B~\citep{yang2025qwen3technicalreport}      & 0.3406                   \\
        $\quad$ w/ CoT  & 0.3392 (\textcolor{RED}{-0.41\%}) \\
        Qwen3\_32B~\citep{yang2025qwen3technicalreport}      & 0.2826                   \\
        $\quad$ w/ CoT  & 0.3037 (\textcolor{GREEN}{+7.47\%}) \\
        GPT\_OSS\_20B~\citep{openai2025gpt0oss0120b}   & 0.3156                   \\
        $\quad$ w/ CoT  & 0.3249 (\textcolor{GREEN}{+2.95\%}) \\
        GPT\_OSS\_120B~\citep{openai2025gpt0oss0120b}  & 0.2997                   \\
        $\quad$ w/ CoT  & 0.3436 (\textcolor{GREEN}{+14.65\%}) \\
        DeepSeek\_R1\_671B\_0528 & 0.3333       \\
        $\quad$ w/ CoT           & 0.3494 (\textcolor{GREEN}{+4.83\%}) \\
        DeepSeek\_V3\_671B\_0324 & 0.3336       \\
        $\quad$ w/ CoT           & 0.3456 (\textcolor{GREEN}{+3.60\%}) \\
        \midrule
        \multicolumn{2}{c}{\textit{Results of Our Models}} \\
        DeepSeek\_R1\_14B\_Instruct                     & 0.3475 (\textcolor{GREY}{baseline})      \\
        $\quad$ w/ CoT                                  & 0.3604 (\textcolor{GREEN}{+3.71\%})      \\
        $\quad$ DeepSeek\_R1\_14B\_GRPO (\textbf{Ours})  & 0.3377 (\textcolor{RED}{-2.25\%})      \\
        $\quad$ DeepSeek\_R1\_14B\_SFT (\textbf{Ours})  & 0.3715 (\textcolor{GREEN}{+6.91\%})      \\
        $\quad$ DeepSeek\_R1\_14B\_SFT\_GRPO (\textbf{Ours}) & \textbf{0.4196} (\textcolor{GREEN}{+20.75\%}) \\
        DeepSeek\_R1\_32B\_Instruct                     & 0.3567 (\textcolor{GREY}{baseline})      \\
        $\quad$ w/ CoT                                  & 0.3589 (\textcolor{GREEN}{+0.62\%})      \\
        $\quad$ DeepSeek\_R1\_32B\_GRPO (\textbf{Ours})  & 0.3683 (\textcolor{GREEN}{+3.22\%})      \\
        $\quad$ DeepSeek\_R1\_32B\_SFT (\textbf{Ours})  & 0.3639 (\textcolor{GREEN}{+2.02\%})      \\
        $\quad$ DeepSeek\_R1\_32B\_SFT\_GRPO (\textbf{Ours}) & 0.4071 (\textcolor{GREEN}{+14.13\%}) \\
        \bottomrule
        \end{tabular}
    }
    \label{fig:compare_results}
\end{minipage}
\hfill
\begin{minipage}[b]{0.49\textwidth}
    \centering
    \includegraphics[width=\textwidth]{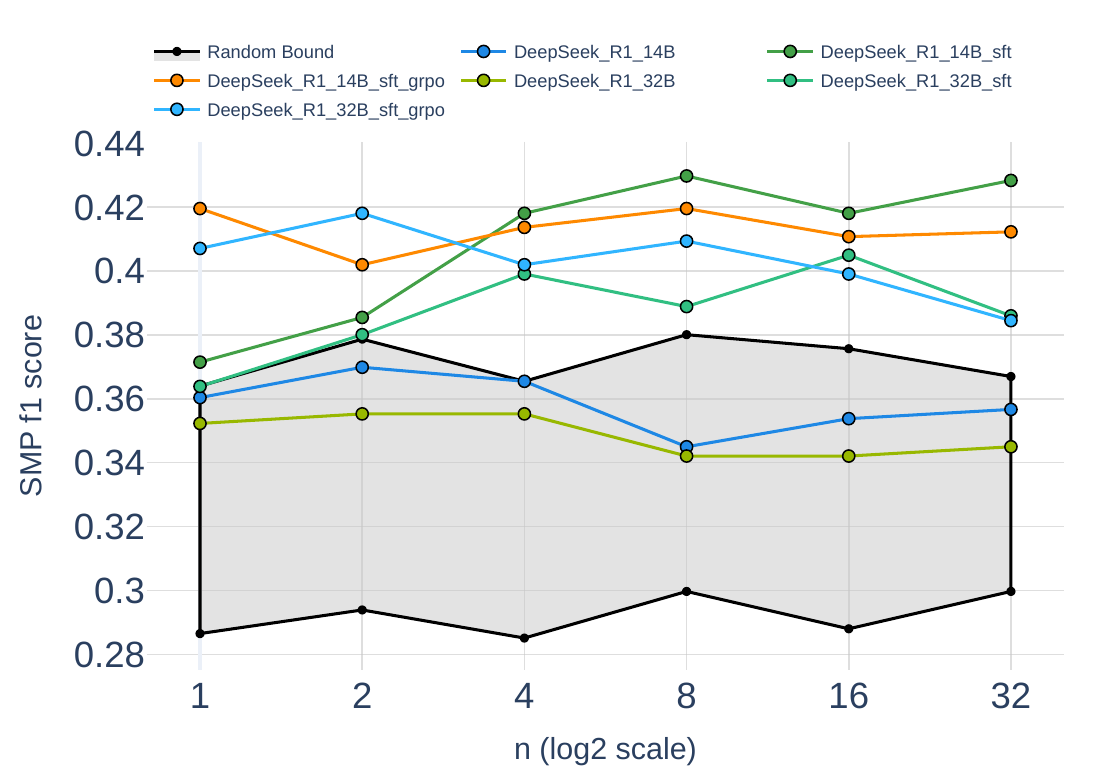}
    \captionof{figure}{Inference-time scalability results on \textbf{Fin-2024[December]}. SFT model already has prediction ability, and GRPO further refines it.}
    \label{fig:main_results}
    \centering
    \includegraphics[width=\textwidth]{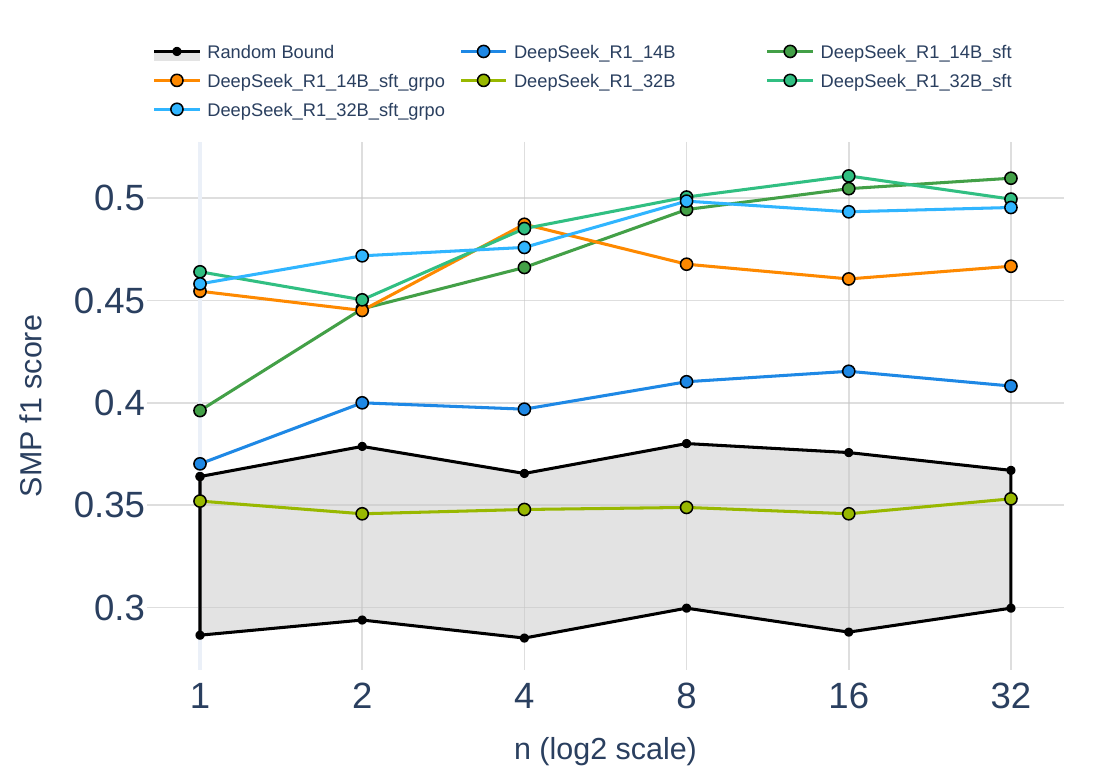}
    \captionof{figure}{Inference-time scalability results on \textbf{Fin-2025[June]}. Finetuned models continue to scale via repeated sampling even after 6 months.}
    \label{fig:June_results}
\end{minipage}



\paragraph{RETuning vs baselines.}
The results of different methods on \textbf{Fin-2024[December]} benchmarks are shown in Table~\ref{fig:compare_results}.
We observe that RETuning (SFT + GRPO) significantly outperforms all baselines, including state-of-the-art open-weight LLMs (DeepSeek, Qwen3, GPT-OSS) and public models specifically designed for stock movement prediction (LLMFactor, Fino1, Fin-R1, CMIN, StockNet). For instance, DeepSeek\_R1\_14B\_SFT\_GRPO achieves an F1 score of 0.4196, which is a 20.75\% relative improvement over its instruct baseline (0.3475) and surpasses the best public model (GPT-OSS-120B w/ CoT at 0.3436) by 22.15\%. Similarly, DeepSeek\_R1\_32B\_SFT\_GRPO attains an F1 score of 0.4071, marking a 14.13\% relative improvement over its instruct baseline (0.3567) and outperforming the best public model by 18.55\%.

\paragraph{Can stock movement prediction benefit from inference-time scaling?}
Yes, but the gains from inference-time scaling are limited. Figure~\ref{fig:main_results} and Figure~\ref{fig:June_results} present the inference-time scalability results on \textbf{Fin-2024[December]} and \textbf{Fin-2025[June]}, respectively.
We observe monotonic or near-monotonic improvements up to $n{\approx}8$--$16$, after which returns plateau and can even regress for some settings. RL (GRPO) makes test-time scaling less necessary by improving one-sample quality, yet does not increase the peak accuracy.

\paragraph{Can predictive ability generalize to unseen stocks, future dates, or both?}
Yes.
We evaluate out-of-distribution (OOD) robustness along two axes: unseen stocks and forward-in-time generalization.
The dataset \textbf{Fin-2024[December]} consists of \textbf{OOD\_Stock}, \textbf{OOD\_Date}, and \textbf{OOD\_Stock\&Date} cases, where RETuning maintains or increases F1 score as the number of inference-time samples $n$ grows (Figure~\ref{fig:main_results}).
On \textbf{Fin-2025[June]} (future dates only), RETuning preserves its advantage and continues to benefit from moderate repeated sampling (Figure~\ref{fig:June_results}), indicating strong temporal and cross-ticker generalization.

To further determine how the model scales on different OOD cases, we group the results by OOD split and present in Figure~\ref{fig:ood_stock_f1_score}\ref{fig:ood_date_f1_score}\ref{fig:ood_stock_date_f1_score}. The scaling is significant in \textbf{OOD\_Stock}, then is \textbf{OOD\_Date}. \textbf{OOD\_Stock\&Date} is the hardest cases to scale up, but it still outperforms baselines. We leave detailed analysis in Appendix~\ref{sec:appendix:detailed_results:group_by_ood}.

\begin{minipage}[b]{0.32\textwidth}
    \centering
    \includegraphics[width=\textwidth]{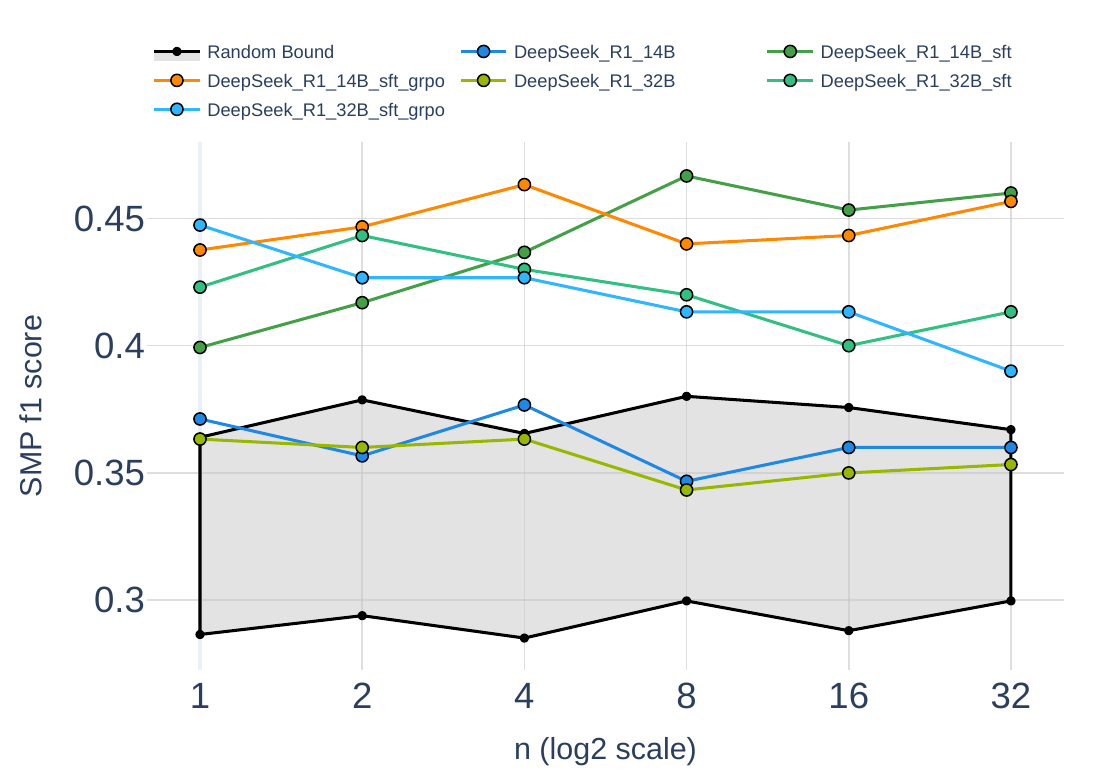}
    \captionof{figure}{\textbf{OOD\_Stock} results on \textbf{Fin-2024[December]}}
    \label{fig:ood_stock_f1_score}
\end{minipage}
\hfill
\begin{minipage}[b]{0.32\textwidth}
    \centering
    \includegraphics[width=\textwidth]{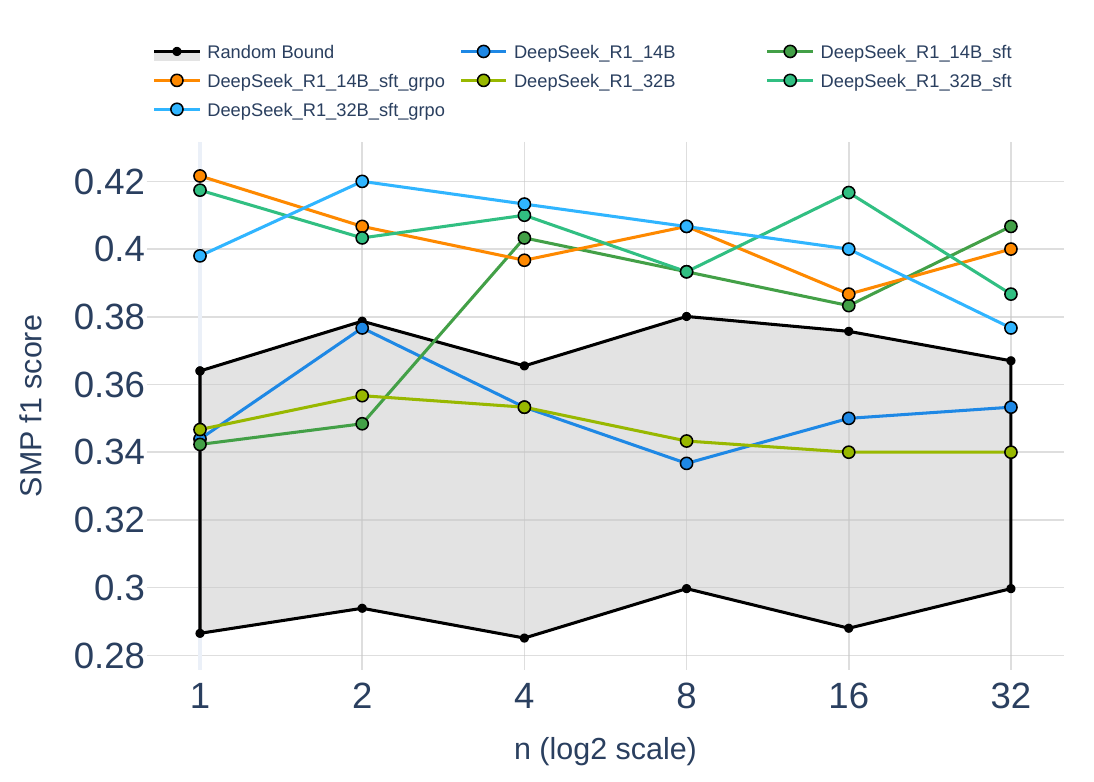}
    \captionof{figure}{\textbf{OOD\_Date} results on \textbf{Fin-2024[December]}}
    \label{fig:ood_date_f1_score}
\end{minipage}
\hfill
\begin{minipage}[b]{0.32\textwidth}
    \centering
    \includegraphics[width=\textwidth]{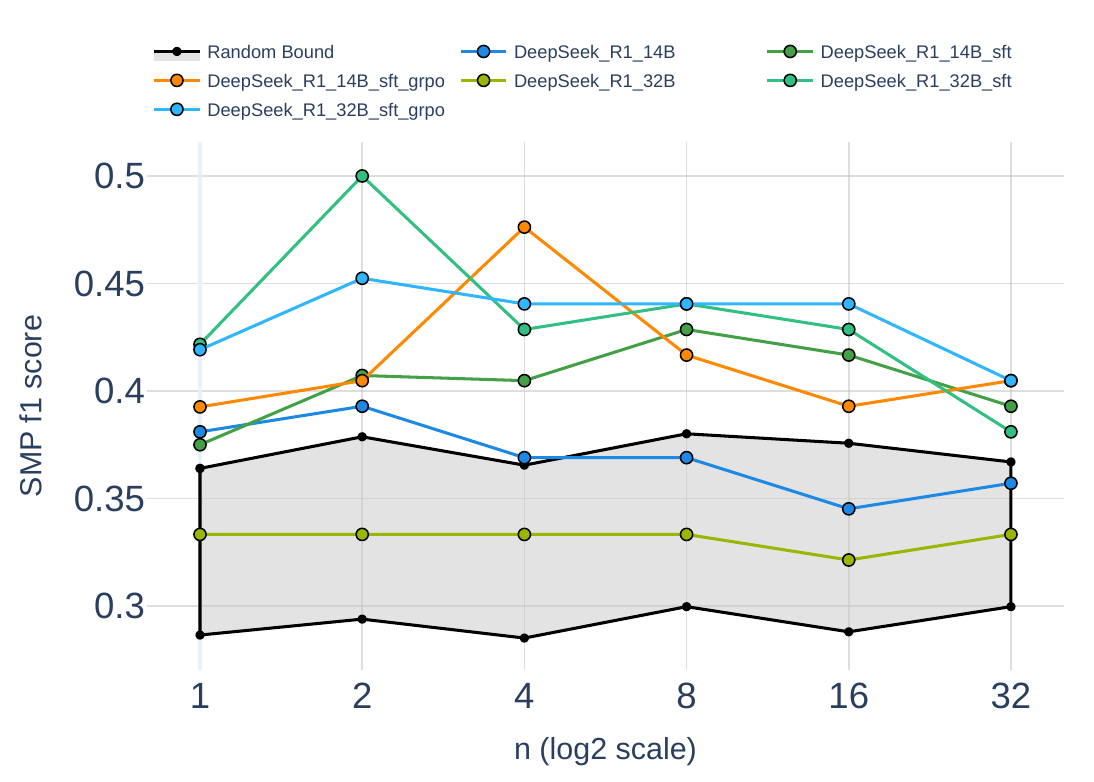}
    \captionof{figure}{\textbf{OOD\_Stock\&Date} results on \textbf{Fin-2024[December]}}
    \label{fig:ood_stock_date_f1_score}
\end{minipage}

We also explore how the model scales on different ground truth labels. The results are grouped by ground truth label and presented in Figure~\ref{fig:label_up_f1_score}\ref{fig:label_hold_f1_score}\ref{fig:label_down_f1_score}. The scaling is significant in \textbf{hold} cases, and the model performance exceeds the baseline on \textbf{up} and \textbf{down} cases. We claim that \textbf{up} and \textbf{down} cases are more challenging, and \textbf{RETuning} enhances the model's performance in these scenarios by enabling it to better leverage its reasoning capabilities, thus to identify factors influencing stock movements more effectively, thereby allowing the model to make more accurate predictions.

\begin{minipage}[b]{0.32\textwidth}
    \centering
    \includegraphics[width=\textwidth]{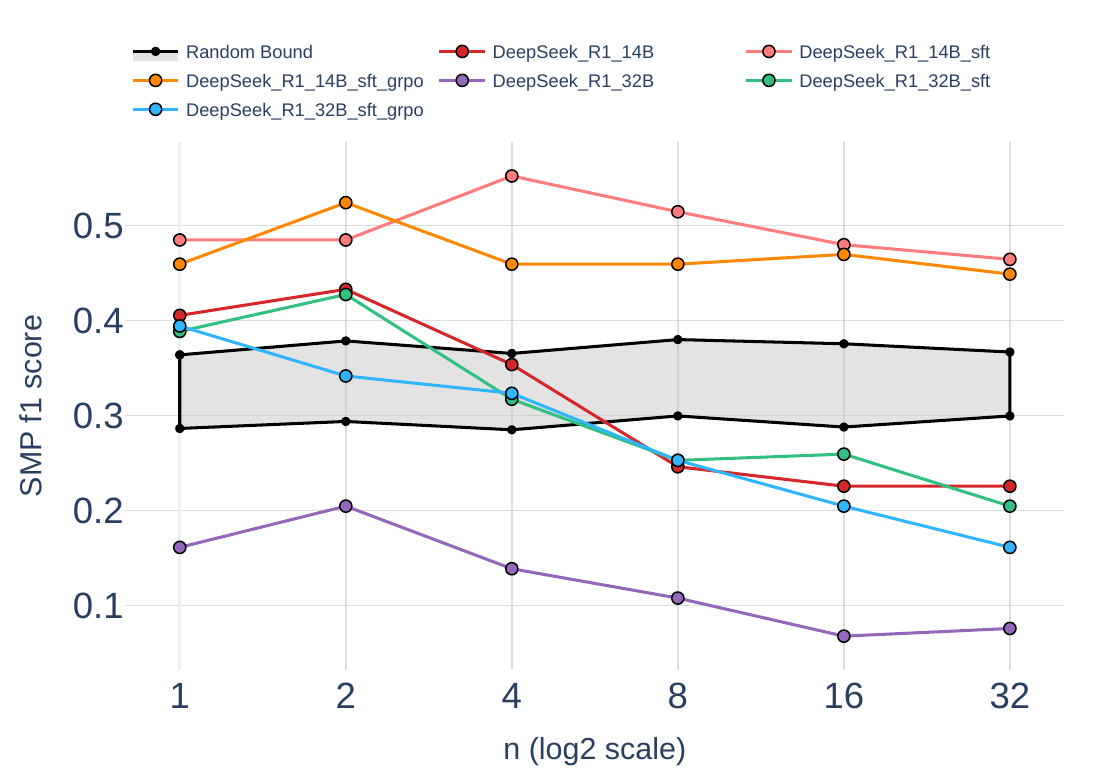}
    \captionof{figure}{Ground truth \textbf{up} results on \textbf{Fin-2024[December]}}
    \label{fig:label_up_f1_score}
\end{minipage}
\hfill
\begin{minipage}[b]{0.32\textwidth}
    \centering
    \includegraphics[width=\textwidth]{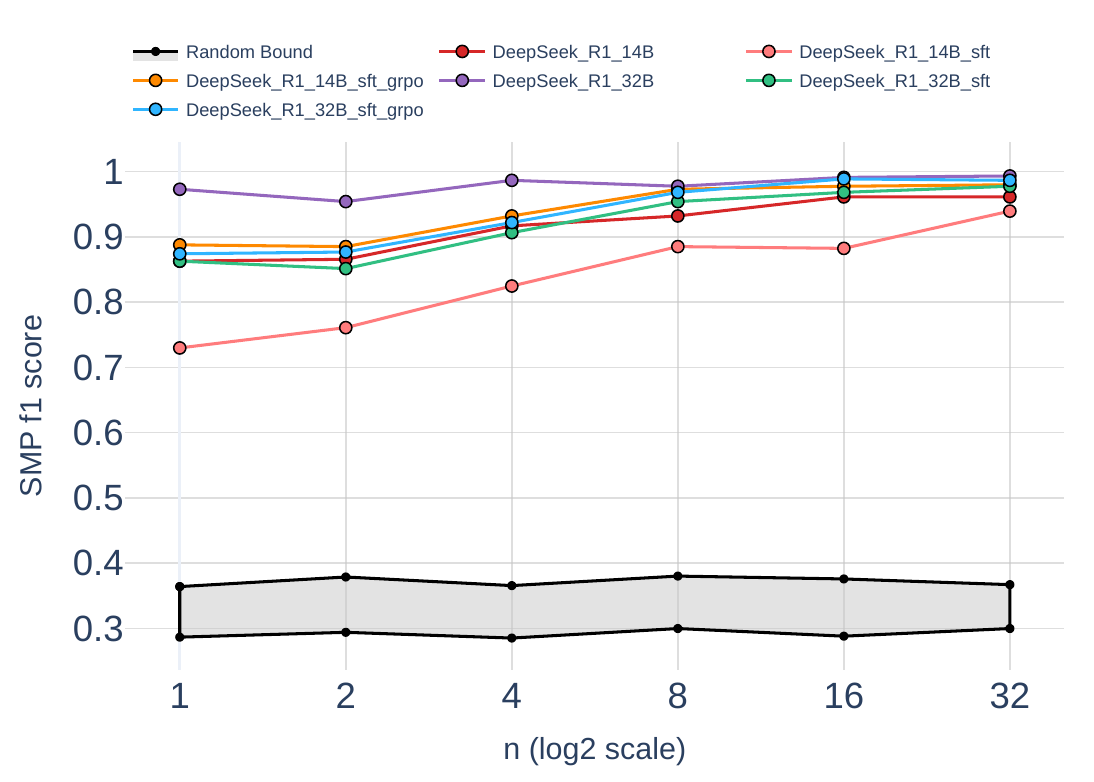}
    \captionof{figure}{Ground truth \textbf{hold} results on \textbf{Fin-2024[December]}}
    \label{fig:label_hold_f1_score}
\end{minipage}
\hfill
\begin{minipage}[b]{0.32\textwidth}
    \centering
    \includegraphics[width=\textwidth]{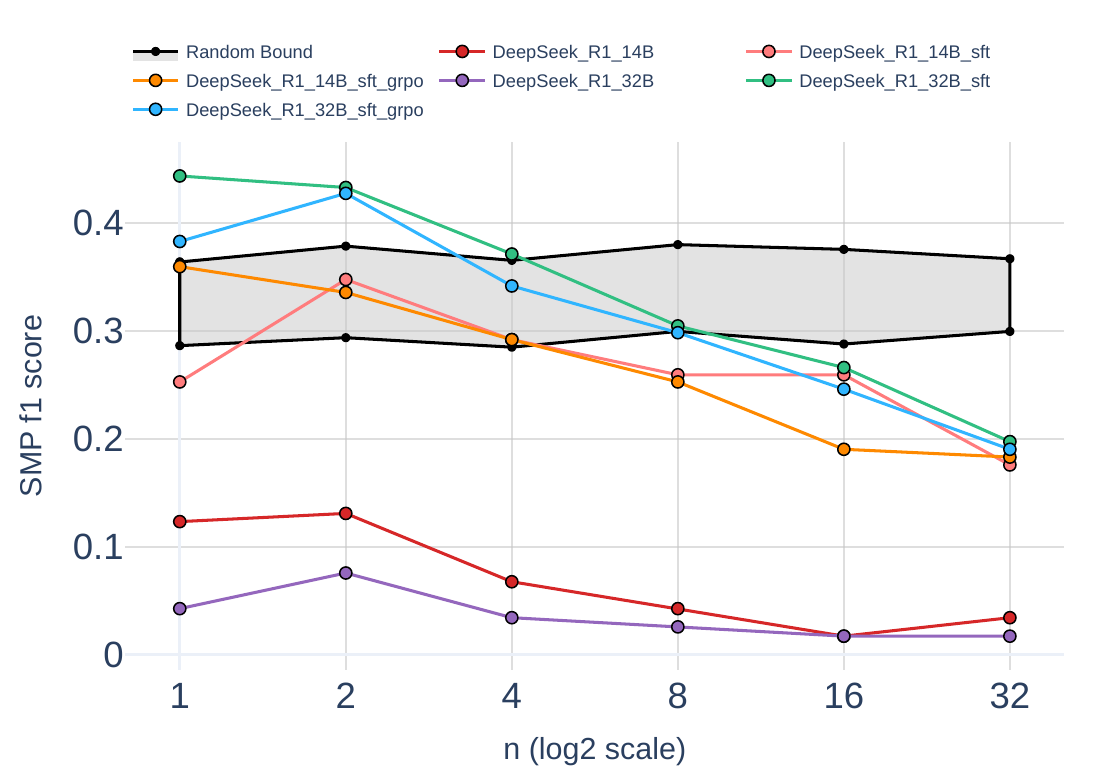}
    \captionof{figure}{Ground truth \textbf{down} results on \textbf{Fin-2024[December]}}
    \label{fig:label_down_f1_score}
\end{minipage}

\paragraph{Ablation on CoT prompting.}
We further examine the effect of Chain-of-Thought (CoT) prompting (Table~\ref{fig:compare_results}).
It causes slightly negative effects on smaller models (Qwen3-8B, Qwen3-14B), but significantly benefits larger models (Qwen3-32B, GPT-OSS-120B, DeepSeek-R1-671B, DeepSeek-V3-671B). This suggests that larger models have a greater capacity to leverage CoT prompting effectively.
For the 14B model, CoT improves the baseline by \textcolor{GREEN}{+3.71\%}, while the 32B model only gains a marginal \textcolor{GREEN}{+0.62\%}. This indicates that CoT alone yields limited benefits. In contrast, our SFT and SFT+GRPO variants consistently outperform CoT, suggesting that structured fine-tuning and reinforcement optimization are more effective than relying on prompting strategies alone.

\paragraph{Ablation on SFT stage.}
We compare the effect of applying GRPO directly versus combining it with an SFT stage (*\_SFT\_GRPO). For the 14B model, GRPO alone underperforms the baseline (0.3377 vs. 0.3475), while SFT followed by GRPO achieves a substantial gain (0.4196, \textcolor{GREEN}{+20.75\%}). A similar trend is observed in the 32B model: GRPO alone yields modest improvement (0.3683, \textcolor{GREEN}{+3.22\%}), whereas SFT+GRPO achieves the best performance (0.4071, \textcolor{GREEN}{+14.13\%}). These results highlight that the SFT stage provides essential initialization, enabling GRPO to realize its full benefit.

\begin{table*}[h]
  \centering
  \caption{Performance Comparison of LLMs on BizFinBench~\citep{lu2025bizfinbench0}. The colors represent the top three performers for each task: \colorbox{golden}{golden} indicates the top-performing model, \colorbox{lightblue}{silver} represents the second-best result, and \colorbox{lightgreen}{bronze} denotes the third-best performance.}
  \resizebox{\textwidth}{!}{%
  \begin{threeparttable}
    \begin{tabular}{lrrrrrrrrrr}
    \toprule
    \multicolumn{1}{c}{Model~\tnote{$\dagger$}} & \multicolumn{1}{c}{AEA} & \multicolumn{1}{c}{FNC} & \multicolumn{1}{c}{FTR} & \multicolumn{1}{c}{FTU} & \multicolumn{1}{c}{FQA} & \multicolumn{1}{c}{FDD} & \multicolumn{1}{c}{ER} & \multicolumn{1}{c}{SP} & \multicolumn{1}{c}{FNER}  & \multicolumn{1}{c}{Average}\\
    \midrule
    \multicolumn{11}{c}{Closed-Source LLMs} \\
    ChatGPT-o3 & \cellcolor{lightblue}86.23 & 61.30     & \cellcolor{lightblue}75.36      & \cellcolor{golden}89.15    & \cellcolor{golden}91.25    & \cellcolor{lightgreen}98.55     & 44.48     & 53.27     & 65.13    & \cellcolor{golden}73.86 \\
    ChatGPT-o4-mini & \cellcolor{lightgreen}85.62 & 60.10     & 71.23      & 74.40    & \cellcolor{lightgreen}90.27    & 95.73     & \cellcolor{golden}47.67     & 52.32     & 64.24    & \cellcolor{lightgreen}71.29 \\
    GPT-4o & 79.42 & 56.51     & \cellcolor{golden}76.20      & 82.37    & 87.79    & \cellcolor{golden}98.84     & 45.33     & 54.33     & \cellcolor{lightgreen}65.37    & \cellcolor{lightblue}71.80 \\
    Gemini-2.0-Flash & \cellcolor{golden}86.94 & 62.67    & \cellcolor{lightgreen}73.97    & 82.55   & \cellcolor{lightblue}90.29    & \cellcolor{lightblue}98.62   & 22.17     & \cellcolor{golden}56.14     & 54.43    & 69.75 \\
    Claude-3.5-Sonnet & 84.68 & \cellcolor{lightgreen}63.18    & 42.81     & \cellcolor{lightblue}88.05     & 87.35     & 96.85     & 16.67     & 47.60     & 63.09    & 65.59 \\
    \midrule
    \multicolumn{11}{c}{Open-Weight LLMs} \\
    Qwen3-14B & 84.20 & 58.20 & 65.80 & 82.19 & 84.12 & 92.91 & 33.00 & 52.31 & 50.70 & 67.05 \\
    Qwen3-32B & 83.80 & 59.60 & 64.60 & 85.12 & 85.43 & 95.37 & 39.00 & 52.26 & 49.19 & 68.26 \\
    DeepSeek\_R1\_14B\_Instruct~\tnote{1} & 71.33 & 44.35 & 50.45     & 81.96    & 85.52     & 92.81   & 39.50     & 50.20     & 52.76    & 59.49 \\
    DeepSeek\_R1\_32B\_Instruct~\tnote{1} & 73.68 & 51.20 & 50.86     & 83.27     & 87.54     & 97.81     & 41.50     & 53.92     & 56.80    & 66.29 \\
    \midrule
    \multicolumn{11}{c}{Our LLMs} \\
    \cellcolor{Background1}DeepSeek\_R1\_14B\_SFT~\tnote{2}       & \cellcolor{Background1}80.63 & \cellcolor{Background1}51.67 & \cellcolor{Background1}52.61 & \cellcolor{Background1}83.53 & \cellcolor{Background1}89.05 & \cellcolor{Background1}96.72 & \cellcolor{Background1}36.68 & \cellcolor{Background1}50.43 & \cellcolor{Background1}50.85 & \cellcolor{Background1}65.36 \\
    $\quad$ 14B $\Delta_{\text{Instruct}}(\text{SFT})$            & \textcolor{GREEN}{+9.25} & \textcolor{GREEN}{+7.28} & \textcolor{GREEN}{+2.19} & \textcolor{GREEN}{+1.53} & \textcolor{GREEN}{+3.42} & \textcolor{GREEN}{+3.85} & \textcolor{RED}{-2.93} & \textcolor{GREEN}{+0.24} & \textcolor{RED}{-1.92} & \textcolor{GREEN}{+5.83} \\
    \cellcolor{Background1}DeepSeek\_R1\_14B\_SFT\_GRPO~\tnote{2} & \cellcolor{Background1}81.46 & \cellcolor{Background1}52.41 & \cellcolor{Background1}53.47 & \cellcolor{Background1}83.57 & \cellcolor{Background1}89.02 & \cellcolor{Background1}95.58 & \cellcolor{Background1}36.83 & \cellcolor{Background1}54.06 & \cellcolor{Background1}51.24 & \cellcolor{Background1}66.92 \\
    $\quad$ 14B $\Delta_{\text{Instruct}}(\text{SFT\_GRPO})$      & \textcolor{GREEN}{+10.03} & \textcolor{GREEN}{+8.09} & \textcolor{GREEN}{+2.91} & \textcolor{GREEN}{+1.64} & \textcolor{GREEN}{+3.45} & \textcolor{GREEN}{+2.63} & \textcolor{RED}{-2.74} & \textcolor{GREEN}{+3.82} & \textcolor{RED}{-1.53} & \textcolor{GREEN}{+7.46} \\
    $\quad$ 14B $\Delta_{\text{SFT}}(\text{SFT\_GRPO})$           & \textcolor{GREEN}{+0.82} & \textcolor{GREEN}{+0.85} & \textcolor{GREEN}{+0.81} & \textcolor{GREEN}{+0.06} & 0.00 & \textcolor{RED}{-1.23} & \textcolor{GREEN}{+0.25} & \textcolor{GREEN}{+3.63} & \textcolor{GREEN}{+0.42} & \textcolor{GREEN}{+1.53} \\
    \cellcolor{Background1}DeepSeek\_R1\_32B\_SFT                 & \cellcolor{Background1}80.45 & \cellcolor{lightblue}66.42 & \cellcolor{Background1}63.28 & \cellcolor{lightgreen}86.88 & \cellcolor{Background1}88.43 & \cellcolor{Background1}93.76 & \cellcolor{lightblue}46.05 & \cellcolor{lightblue}55.27 & \cellcolor{golden}68.41 & \cellcolor{Background1}70.08 \\
    $\quad$ 32B $\Delta_{\text{Instruct}}(\text{SFT})$            & \textcolor{GREEN}{+6.75} & \textcolor{GREEN}{+15.23} & \textcolor{GREEN}{+12.37} & \textcolor{GREEN}{+3.64} & \textcolor{GREEN}{+0.83} & \textcolor{RED}{-4.14} & \textcolor{GREEN}{+4.52} & \textcolor{GREEN}{+1.25} & \textcolor{GREEN}{+11.63} & \textcolor{GREEN}{+3.75} \\
    \cellcolor{Background1}DeepSeek\_R1\_32B\_SFT\_GRPO           & \cellcolor{Background1}80.67 & \cellcolor{golden}66.83 & \cellcolor{Background1}64.45 & \cellcolor{Background1}86.79 & \cellcolor{Background1}88.52 & \cellcolor{Background1}91.26 & \cellcolor{lightgreen}45.68 & \cellcolor{lightgreen}54.83 & \cellcolor{lightblue}67.75 & \cellcolor{Background1}70.44 \\
    $\quad$ 32B $\Delta_{\text{Instruct}}(\text{SFT\_GRPO})$      & \textcolor{GREEN}{+6.95} & \textcolor{GREEN}{+15.62} & \textcolor{GREEN}{+13.57} & \textcolor{GREEN}{+3.55} & \textcolor{GREEN}{+0.93} & \textcolor{RED}{-6.64} & \textcolor{GREEN}{+4.13} & \textcolor{GREEN}{+0.85} & \textcolor{GREEN}{+10.93} & \textcolor{GREEN}{+4.12} \\
    $\quad$ 32B $\Delta_{\text{SFT}}(\text{SFT\_GRPO})$           & \textcolor{GREEN}{+0.23} & \textcolor{GREEN}{+0.42} & \textcolor{GREEN}{+1.23} & \textcolor{RED}{-0.06} & \textcolor{GREEN}{+0.13} & \textcolor{RED}{-2.53} & \textcolor{RED}{-0.42} & \textcolor{RED}{-0.43} & \textcolor{RED}{-0.72} & \textcolor{GREEN}{+0.33} \\
    \bottomrule
  \end{tabular}%
  \begin{tablenotes}
    \item[$\dagger$] Closed-source LLMs results are sourced from \cite{lu2025bizfinbench0}. Open-weight LLMs results are reproduced using temperature $0.6$.
    \item[1] DeepSeek\_R1\_14B\_Instruct (32B) here is the short of DeepSeek-R1-Distill-Qwen-14B (32B) in the original paper~\citep{deepseek-r1}.
    \item[2] DeepSeek\_R1\_14B\_SFT and DeepSeek\_R1\_14B\_SFT\_GRPO are DeepSeek\_R1\_14B\_Instruct model after RETuning SFT stage and after RETuning SFT + GRPO stages, respectively.
  \end{tablenotes}
  \end{threeparttable}
    }
  \label{tab:contribution_to_other_fin_tasks}%
\end{table*}



\paragraph{Does the enhanced predictive ability contribute to other financial tasks?}
Yes. On the financial benchmark BizFinBench~\citep{lu2025bizfinbench0} (Table~\ref{tab:contribution_to_other_fin_tasks}), \textbf{RETuning} generalizes beyond SMP: for 14B, the average score improves from 59.49 (Instruct) to 65.36 (SFT, $+5.83$) and 66.92 (SFT+GRPO, $+7.46$); for 32B, it improves from 66.29 to 70.08 ($+3.75$) and 70.44 ($+4.12$). The 32B models reach top-3 results on several tasks (e.g., FNC, FTU, ER, SP, FNER), while minor regressions appear on highly saturated dimensions (e.g., FDD after RL: $-1.23$ for 14B, $-2.53$ for 32B). Overall, \textbf{RETuning} yields broad, transferable gains with small trade-offs on a few tasks.

\paragraph{What insights does model learn to predict stock movement?}
Through analyzing the model's responses detailed in Appendix~\ref{sec:appendix:case_study:response}, we find that the model learns to:
1. Identifying key evidences that influence daily fluctuations in stock prices from multiple information sources.
2. Trending to correctly evaluate the short-term impact of gathered evidences, which benefits from the trade-off ability induced by adversarial scoring.
3. Gradually improving the consistency between the predicted fluctuation and the movement label during reinforcement learning, which we named "vibe prediction".







\section{Conclusion}\label{sec:conclusion}

In this work, we explored the underexamined problem of applying large language models (LLMs) to stock movement prediction. Our analysis revealed that vanilla LLMs tend to rely on contextual viewpoints rather than developing independent analytical reasoning, which limits their predictive reliability. To address this challenge, we introduced \textbf{RETuning}, a reflective evidence-based tuning method that encourages models to construct analytical frameworks, weigh adversarial evidence, and refine predictions through reflection. Experiments on our newly constructed large-scale financial dataset demonstrate that RETuning substantially improves predictive performance over strong baselines, enabling more systematic reasoning in the financial domain. Moreover, RETuning generalizes beyond stock movement prediction, yielding gains across diverse financial tasks, and exhibits robustness under inference-time scaling and out-of-distribution settings. Overall, this study highlights the importance of evidence-oriented reasoning in financial LLMs and establishes RETuning as a promising direction for enhancing their reliability and applicability in real-world financial decision-making.

\newpage

\section*{Ethics Statement} \label{sec:ethics_statement}
To address potential ethical considerations related to our research on large language models (LLMs) for stock movement prediction, we provide the following statement:

First, regarding data ethics: Our large-scale dataset (spanning 2024 for 5,123 A-share stocks) is constructed exclusively from \textcolor{Primary1}{publicly available sources}, including market price data, publicly disclosed news, analysts’ public opinions, company fundamental reports, official macroeconomic indicators, and publicly accessible information on peer stocks. We strictly comply with China’s Data Security Law, Securities Law, and relevant financial regulatory requirements, ensuring no collection or use of private, sensitive personal data, or non-public material information that could violate market fairness.

Second, on potential harm and application boundaries: This research is conducted for \textcolor{Primary1}{academic purposes only} to advance LLM reasoning capabilities in financial tasks. We explicitly emphasize that our model (RETuning) and findings do not constitute financial advice, nor do they endorse or promote real-world investment decisions. Stock market prediction inherently carries high uncertainty, and any practical application of such models for trading could lead to financial risks; we disclaim responsibility for any losses arising from non-academic use of our work.

Third, on bias and fairness: While we designed RETuning to enhance independent logical reasoning (reducing undue reliance on contextual viewpoints) and constructed a diverse dataset to cover a broad range of A-share stocks, we acknowledge potential residual biases (e.g., sector-specific skews in training data or sensitivity to market cycles). Future work will further validate and mitigate such biases to improve the model’s fairness across different market scenarios.

Finally, regarding research integrity: Our study involves no human subjects, so institutional review board (IRB) approval is not applicable. We commit to transparency in dataset construction details and methodology implementation (as detailed in the full paper) to enable reproducibility. We adhere to rigorous academic standards to avoid misrepresentation of results or misuse of technical insights.

\section*{Reproducibility Statement} \label{sec:reproducibility_statement}

We make every effort to ensure that the experiments in this paper are reproducible.  Specifically, anonymized source code (training and evaluation scripts), model checkpoints, and processed dataset splits will be released as supplementary material and at the repositories indicated in the Appendix.\footnote{\repourl, \huggingfaceurl, \dataseturl} The Appendix contains detailed descriptions of data collection and preprocessing (Appendix~\ref{sec:appendix:dataset_details}), the prompt templates and example inputs/outputs (Figures~\ref{fig:appendix:prompt_template}--\ref{fig:appendix:example_response_grading}), and the exact training and evaluation settings including compute and hardware details (Appendix~\ref{sec:appendix:implementation_details}, Tables~\ref{table:hyperparameters:sft} and~\ref{table:hyperparameters:grpo}).  Evaluation splits used for OOD and long-horizon tests (e.g., \textbf{Fin-2024[December]}, \textbf{Fin-2025[June]}) and the scripts to compute all reported metrics will also be provided.  Where full raw data cannot be released due to third-party licensing or privacy constraints, we describe the access procedure and provide processed, reproducible derivatives in the supplementary materials.

\section*{Broader Impact} \label{sec:broader_impact}
This work investigates the use of large language models (LLMs) for stock movement prediction, a domain with potentially high economic and societal implications. Our proposed method, RETuning, demonstrates how reflective evidence organization can enhance independent reasoning in financial tasks. On the positive side, this research contributes to the broader effort of making LLMs more reliable in high-stakes applications by encouraging systematic analysis rather than context-driven imitation. Such improvements may benefit both academic research in reasoning and practical applications in financial decision-support systems.

At the same time, we emphasize that financial forecasting is inherently uncertain and subject to market volatility, regulatory constraints, and ethical considerations. Automated prediction systems, if misused, could amplify risks, encourage speculative behavior, or contribute to unfair advantages for certain market participants. Our dataset and methods are designed for research purposes only, and we strongly discourage their direct use for live trading or investment without rigorous safeguards, human oversight, and compliance with financial regulations.

More broadly, this work highlights both the opportunities and limitations of applying LLMs in sensitive domains. We hope that our findings spur further research into building transparent, evidence-based reasoning frameworks that improve model reliability while also ensuring responsible deployment in practice.

\section*{Limitations} \label{sec:limitation}

\paragraph{Domain specificity.}
Our study focuses on the Chinese A-share market in 2024, which provides a rich testbed but may limit the generalizability of findings to other markets, such as U.S. equities or emerging markets with different structures, liquidity, and regulatory conditions.

\paragraph{Data coverage.}
Although our dataset integrates multiple information sources (prices, news, analyst opinions, fundamentals, and macroeconomic indicators), it remains incomplete. Certain high-frequency signals (e.g., intraday order flow, alternative data, or global macro shocks) are not incorporated, potentially constraining predictive accuracy.

\paragraph{Model assumptions.}
RETuning assumes that LLMs can benefit from explicitly structuring and reflecting on evidence. While this holds in our experiments, the approach may be less effective in domains like healthcare or cryptocurrency, where reliable evidence is scarce, noisy, or difficult to formalize.

\paragraph{Evaluation scope.}
Our evaluation mainly relies on F1 scores for three-class stock movement prediction and selected benchmarks (e.g., BizFinBench~\citep{lu2025bizfinbench0}). Broader metrics, such as profitability in trading simulations or risk-adjusted returns, are not considered. It takes time to validate real-world trading performance, which is beyond the scope of this paper.

\paragraph{Computation and scalability.}
Both training (SFT + GRPO) and inference-time scaling are computationally expensive. This may limit accessibility for smaller institutions or researchers without large-scale compute resources, raising questions about cost-efficiency in real-world deployment.

\section*{Future Work} \label{sec:future_work}


\paragraph{Extending to other markets.}
Future research could examine the effectiveness of RETuning across diverse markets such as U.S. equities, European exchanges, and emerging markets. This would validate whether the approach generalizes under different regulatory regimes, liquidity conditions, and investor behaviors.

\paragraph{Incorporating richer data sources.}
Enhancing the dataset with high-frequency trading signals, alternative data (e.g., satellite imagery, ESG reports), and global macroeconomic factors could provide a more comprehensive information environment and further strengthen predictive power.

\paragraph{Advancing reasoning frameworks.}
While RETuning encourages evidence-based reasoning, future work may integrate causal inference, probabilistic reasoning, or game-theoretic perspectives to capture deeper structures behind stock movements and reduce susceptibility to spurious correlations.

\paragraph{Evaluation beyond prediction accuracy.}
A natural next step is to link model predictions with financial outcomes, such as profitability, portfolio optimization, and risk-adjusted returns. This would bridge the gap between benchmark metrics and real-world decision-making.

\paragraph{Efficiency and accessibility.}
Research on lightweight RETuning variants, parameter-efficient fine-tuning, and inference-time acceleration could reduce computational costs, making the method more accessible to practitioners and researchers with limited resources.

\bibliographystyle{iclr2026_conference}
\bibliography{references/_paper, references/timeseries}

\begin{thebibliography}{69}
\providecommand{\natexlab}[1]{#1}
\providecommand{\url}[1]{\texttt{#1}}
\expandafter\ifx\csname urlstyle\endcsname\relax
  \providecommand{\doi}[1]{doi: #1}\else
  \providecommand{\doi}{doi: \begingroup \urlstyle{rm}\Url}\fi

\bibitem[Achiam et~al.(2023)Achiam, Adler, Agarwal, Ahmad, Akkaya, Aleman, Almeida, Altenschmidt, Altman, Anadkat, Avila, Babuschkin, Balaji, Balcom, Baltescu, ing Bao, Bavarian, Belgum, Bello, Berdine, Bernadett-Shapiro, Berner, Bogdonoff, Boiko, laine Boyd, Brakman, Brockman, Brooks, Brundage, Button, Cai, Campbell, Cann, Carey, Carlson, Carmichael, Chan, Chang, Chantzis, Chen, Chen, Chen, Chen, Chen, Chess, Cho, Chu, Chung, Cummings, Currier, Dai, Decareaux, Degry, Deutsch, Deville, Dhar, Dohan, Dowling, Dunning, Ecoffet, Eleti, Eloundou, Farhi, Fedus, Felix, Fishman, Forte, abella Fulford, Gao, Georges, Gibson, Goel, Gogineni, Goh, Gontijo-Lopes, Gordon, Grafstein, Gray, Greene, Gross, Gu, Guo, Hallacy, Han, Harris, He, Heaton, Heidecke, Hesse, Hickey, Hickey, Hoeschele, Houghton, Hsu, Hu, Hu, Huizinga, Jain, Jain, Jang, Jiang, Jiang, Jin, Jin, Jomoto, Jonn, Jun, Kaftan, Kaiser, Kamali, Kanitscheider, Keskar, Khan, Kilpatrick, Kim, Kim, Kim, Kirchner, Kiros, Knight, Kokotajlo, Kondraciuk, Kondrich, Konstantinidis, Kosic, Krueger, Kuo, Lampe, Lan, Lee, Leike, Leung, Levy, Li, Lim, Lin, Lin, teusz Litwin, Lopez, Lowe, Lue, Makanju, Malfacini, Manning, Markov, Markovski, Martin, Mayer, Mayne, McGrew, McKinney, McLeavey, McMillan, McNeil, Medina, Mehta, Menick, Metz, drey Mishchenko, Mishkin, Monaco, Morikawa, Mossing, Mu, Murati, Murk, M'ely, Nair, Nakano, Nayak, Neelakantan, Ngo, Noh, Long, O'Keefe, Pachocki, Paino, Palermo, Pantuliano, Parascandolo, Parish, Parparita, Passos, Pavlov, Peng, Perelman, de~Avila Belbute~Peres, Petrov, de~Oliveira~Pinto, Pokorny, Pokrass, Pong, Powell, Power, Power, Proehl, Puri, Radford, Rae, Ramesh, Raymond, Real, Rimbach, Ross, Rotsted, Roussez, Ryder, Saltarelli, Sanders, Santurkar, Sastry, Schmidt, Schnurr, Schulman, Selsam, Sheppard, Sherbakov, Shieh, Shoker, Shyam, Sidor, Sigler, Simens, Sitkin, Slama, Sohl, Sokolowsky, Song, Staudacher, Such, Summers, Sutskever, Tang, Tezak, Thompson, Tillet, Tootoonchian, Tseng, Tuggle, Turley, Tworek, Uribe, Vallone, Vijayvergiya, Voss, Wainwright, Wang, Wang, Wang, Ward, Wei, Weinmann, Welihinda, Welinder, Weng, Weng, Wiethoff, Willner, Winter, Wolrich, Wong, Workman, Wu, Wu, Wu, Xiao, Xu, Yoo, Yu, ing Yuan, Zaremba, Zellers, Zhang, Zhang, Zhao, Zheng, Zhuang, Zhuk, and Zoph]{Achiam2023GPT4TR}
OpenAI~Josh Achiam, Steven Adler, Sandhini Agarwal, Lama Ahmad, Ilge Akkaya, Florencia~Leoni Aleman, Diogo Almeida, Janko Altenschmidt, Sam Altman, Shyamal Anadkat, Red Avila, Igor Babuschkin, Suchir Balaji, Valerie Balcom, Paul Baltescu, Haim ing Bao, Mo~Bavarian, Jeff Belgum, Irwan Bello, Jake Berdine, Gabriel Bernadett-Shapiro, Christopher Berner, Lenny Bogdonoff, Oleg Boiko, Made laine Boyd, Anna-Luisa Brakman, Greg Brockman, Tim Brooks, Miles Brundage, Kevin Button, Trevor Cai, Rosie Campbell, Andrew Cann, Brittany Carey, Chelsea Carlson, Rory Carmichael, Brooke Chan, Che Chang, Fotis Chantzis, Derek Chen, Sully Chen, Ruby Chen, Jason Chen, Mark Chen, Benjamin Chess, Chester Cho, Casey Chu, Hyung~Won Chung, Dave Cummings, Jeremiah Currier, Yunxing Dai, Cory Decareaux, Thomas Degry, Noah Deutsch, Damien Deville, Arka Dhar, David Dohan, Steve Dowling, Sheila Dunning, Adrien Ecoffet, Atty Eleti, Tyna Eloundou, David Farhi, Liam Fedus, Niko Felix, Sim'on~Posada Fishman, Juston Forte, Is~abella Fulford, Leo Gao, Elie Georges, Christian Gibson, Vik Goel, Tarun Gogineni, Gabriel Goh, Raphael Gontijo-Lopes, Jonathan Gordon, Morgan Grafstein, Scott Gray, Ryan Greene, Joshua Gross, Shixiang~Shane Gu, Yufei Guo, Chris Hallacy, Jesse Han, Jeff Harris, Yuchen He, Mike Heaton, Johannes Heidecke, Chris Hesse, Alan Hickey, Wade Hickey, Peter Hoeschele, Brandon Houghton, Kenny Hsu, Shengli Hu, Xin Hu, Joost Huizinga, Shantanu Jain, Shawn Jain, Joanne Jang, Angela Jiang, Roger Jiang, Haozhun Jin, Denny Jin, Shino Jomoto, Billie Jonn, Heewoo Jun, Tomer Kaftan, Lukasz Kaiser, Ali Kamali, Ingmar Kanitscheider, Nitish~Shirish Keskar, Tabarak Khan, Logan Kilpatrick, Jong~Wook Kim, Christina Kim, Yongjik Kim, Hendrik Kirchner, Jamie~Ryan Kiros, Matthew Knight, Daniel Kokotajlo, Lukasz Kondraciuk, Andrew Kondrich, Aris Konstantinidis, Kyle Kosic, Gretchen Krueger, Vishal Kuo, Michael Lampe, Ikai Lan, Teddy Lee, Jan Leike, Jade Leung, Daniel Levy, Chak Li, Rachel Lim, Molly Lin, Stephanie Lin, Ma~teusz Litwin, Theresa Lopez, Ryan Lowe, Patricia Lue, Anna Makanju, Kim Malfacini, Sam Manning, Todor Markov, Yaniv Markovski, Bianca Martin, Katie Mayer, Andrew Mayne, Bob McGrew, Scott~Mayer McKinney, Christine McLeavey, Paul McMillan, Jake McNeil, David Medina, Aalok Mehta, Jacob Menick, Luke Metz, An~drey Mishchenko, Pamela Mishkin, Vinnie Monaco, Evan Morikawa, Daniel~P. Mossing, Tong Mu, Mira Murati, Oleg Murk, David M'ely, Ashvin Nair, Reiichiro Nakano, Rajeev Nayak, Arvind Neelakantan, Richard Ngo, Hyeonwoo Noh, Ouyang Long, Cullen O'Keefe, Jakub~W. Pachocki, Alex Paino, Joe Palermo, Ashley Pantuliano, Giambattista Parascandolo, Joel Parish, Emy Parparita, Alexandre Passos, Mikhail Pavlov, Andrew Peng, Adam Perelman, Filipe de~Avila Belbute~Peres, Michael Petrov, Henrique~Pond{\'e} de~Oliveira~Pinto, Michael Pokorny, Michelle Pokrass, Vitchyr~H. Pong, Tolly Powell, Alethea Power, Boris Power, Elizabeth Proehl, Raul Puri, Alec Radford, Jack~W. Rae, Aditya Ramesh, Cameron Raymond, Francis Real, Kendra Rimbach, Carl Ross, Bob Rotsted, Henri Roussez, Nick Ryder, Mario~D. Saltarelli, Ted Sanders, Shibani Santurkar, Girish Sastry, Heather Schmidt, David Schnurr, John Schulman, Daniel Selsam, Kyla Sheppard, Toki Sherbakov, Jessica Shieh, Sarah Shoker, Pranav Shyam, Szymon Sidor, Eric Sigler, Maddie Simens, Jordan Sitkin, Katarina Slama, Ian Sohl, Benjamin Sokolowsky, Yang Song, Natalie Staudacher, Felipe~Petroski Such, Natalie Summers, Ilya Sutskever, Jie Tang, Nikolas~A. Tezak, Madeleine Thompson, Phil Tillet, Amin Tootoonchian, Elizabeth Tseng, Preston Tuggle, Nick Turley, Jerry Tworek, Juan Felipe~Cer'on Uribe, Andrea Vallone, Arun Vijayvergiya, Chelsea Voss, Carroll~L. Wainwright, Justin~Jay Wang, Alvin Wang, Ben Wang, Jonathan Ward, Jason Wei, CJ~Weinmann, Akila Welihinda, Peter Welinder, Jiayi Weng, Lilian Weng, Matt Wiethoff, Dave Willner, Clemens Winter, Samuel Wolrich, Hannah Wong, Lauren Workman, Sherwin Wu, Jeff Wu, Michael Wu, Kai Xiao, Tao Xu, Sarah Yoo, Kevin Yu, Qim ing Yuan, Wojciech Zaremba, Rowan Zellers, Chong Zhang, Marvin Zhang, Shengjia Zhao, Tianhao Zheng, Juntang Zhuang, William Zhuk, and Barret Zoph.
\newblock Gpt-4 technical report.
\newblock 2023.
\newblock URL \url{https://api.semanticscholar.org/CorpusID:257532815}.

\bibitem[Adyatma \& Alamsyah(2022)Adyatma and Alamsyah]{Adyatma2022TheIS}
Farhan Adyatma and Andry Alamsyah.
\newblock The indonesia stock exchange composite prediction based on macroeconomic indicators using arima, lstm, and ann.
\newblock \emph{2022 8th International Conference on Science and Technology (ICST)}, 1:\penalty0 1--5, 2022.
\newblock URL \url{https://api.semanticscholar.org/CorpusID:258992891}.

\bibitem[Bai et~al.(2023)Bai, Bai, Chu, Cui, Dang, Deng, Fan, Ge, Han, Huang, Hui, Ji, Li, Lin, Lin, Liu, Liu, Lu, Lu, Ma, Men, Ren, Ren, Tan, Tan, Tu, Wang, Wang, Wang, Wu, Xu, Xu, Yang, Yang, Yang, Yang, Yang, Yao, Yu, Bowen, Yuan, Yuan, Zhang, Zhang, Zhang, Zhang, Zhou, Zhou, Zhou, and Zhu]{Qwen}
Jinze Bai, Shuai Bai, Yunfei Chu, Zeyu Cui, Kai Dang, Xiaodong Deng, Yang Fan, Wenhang Ge, Yu~Han, Fei Huang, Binyuan Hui, Luo Ji, Mei Li, Junyang Lin, Runji Lin, Dayiheng Liu, Gao Liu, Chengqiang Lu, K.~Lu, Jianxin Ma, Rui Men, Xingzhang Ren, Xuancheng Ren, Chuanqi Tan, Sinan Tan, Jianhong Tu, Peng Wang, Shijie Wang, Wei Wang, Shengguang Wu, Benfeng Xu, Jin Xu, An~Yang, Hao Yang, Jian Yang, Jian Yang, Shusheng Yang, Yang Yao, Bowen Yu, Yu~Bowen, Hongyi Yuan, Zheng Yuan, Jianwei Zhang, Xing Zhang, Yichang Zhang, Zhenru Zhang, Chang Zhou, Jingren Zhou, Xiaohuan Zhou, and Tianhang Zhu.
\newblock Qwen technical report.
\newblock \emph{ArXiv}, abs/2309.16609, 2023.
\newblock URL \url{https://api.semanticscholar.org/CorpusID:263134555}.

\bibitem[Brown et~al.(2020)Brown, Mann, Ryder, Subbiah, Kaplan, Dhariwal, Neelakantan, Shyam, Sastry, Askell, et~al.]{brown2020language}
Tom Brown, Benjamin Mann, Nick Ryder, Melanie Subbiah, Jared~D Kaplan, Prafulla Dhariwal, Arvind Neelakantan, Pranav Shyam, Girish Sastry, Amanda Askell, et~al.
\newblock Language models are few-shot learners.
\newblock \emph{Advances in neural information processing systems}, 33:\penalty0 1877--1901, 2020.

\bibitem[Chari et~al.(2025)Chari, Tiwari, Lian, Reddy, and Zhou]{chari2025pheromonebasedlearningoptimalreasoning}
Anirudh Chari, Aditya Tiwari, Richard Lian, Suraj Reddy, and Brian Zhou.
\newblock Pheromone-based learning of optimal reasoning paths, 2025.
\newblock URL \url{https://arxiv.org/abs/2501.19278}.

\bibitem[Chen et~al.(2024)Chen, Liao, Li, and Fan]{chen2024alphamathzeroprocesssupervision}
Guoxin Chen, Minpeng Liao, Chengxi Li, and Kai Fan.
\newblock Alphamath almost zero: Process supervision without process, 2024.
\newblock URL \url{https://arxiv.org/abs/2405.03553}.

\bibitem[Cobbe et~al.(2021)Cobbe, Kosaraju, Bavarian, Chen, Jun, Kaiser, Plappert, Tworek, Hilton, Nakano, Hesse, and Schulman]{cobbe2021trainingverifierssolvemath}
Karl Cobbe, Vineet Kosaraju, Mohammad Bavarian, Mark Chen, Heewoo Jun, Lukasz Kaiser, Matthias Plappert, Jerry Tworek, Jacob Hilton, Reiichiro Nakano, Christopher Hesse, and John Schulman.
\newblock Training verifiers to solve math word problems, 2021.
\newblock URL \url{https://arxiv.org/abs/2110.14168}.

\bibitem[Contributors(2023)]{2023xtuner}
XTuner Contributors.
\newblock Xtuner: A toolkit for efficiently fine-tuning llm.
\newblock \url{https://github.com/InternLM/xtuner}, 2023.

\bibitem[DeepSeek-AI et~al.(2025)DeepSeek-AI, Guo, Yang, Zhang, Song, Zhang, Xu, Zhu, Ma, Wang, Bi, Zhang, Yu, Wu, Wu, Gou, Shao, Li, Gao, Liu, Xue, Wang, Wu, Feng, Lu, Zhao, Deng, Zhang, Ruan, Dai, Chen, Ji, Li, Lin, Dai, Luo, Hao, Chen, Li, Zhang, Bao, Xu, Wang, Ding, Xin, Gao, Qu, Li, Guo, Li, Wang, Chen, Yuan, Qiu, Li, Cai, Ni, Liang, Chen, Dong, Hu, Gao, Guan, Huang, Yu, Wang, Zhang, Zhao, Wang, Zhang, Xu, Xia, Zhang, Zhang, Tang, Li, Wang, Li, Tian, Huang, Zhang, Wang, Chen, Du, Ge, Zhang, Pan, Wang, Chen, Jin, Chen, Lu, Zhou, Chen, Ye, Wang, Yu, Zhou, Pan, Li, Zhou, Wu, Ye, Yun, Pei, Sun, Wang, Zeng, Zhao, Liu, Liang, Gao, Yu, Zhang, Xiao, An, Liu, Wang, Chen, Nie, Cheng, Liu, Xie, Liu, Yang, Li, Su, Lin, Li, Jin, Shen, Chen, Sun, Wang, Song, Zhou, Wang, Shan, Li, Wang, Wei, Zhang, Xu, Li, Zhao, Sun, Wang, Yu, Zhang, Shi, Xiong, He, Piao, Wang, Tan, Ma, Liu, Guo, Ou, Wang, Gong, Zou, He, Xiong, Luo, You, Liu, Zhou, Zhu, Xu, Huang, Li, Zheng, Zhu, Ma, Tang, Zha, Yan, Ren, Ren, Sha, Fu, Xu, Xie, Zhang, Hao, Ma, Yan, Wu, Gu, Zhu, Liu, Li, Xie, Song, Pan, Huang, Xu, Zhang, and Zhang]{deepseek-r1}
DeepSeek-AI, Daya Guo, Dejian Yang, Haowei Zhang, Junxiao Song, Ruoyu Zhang, Runxin Xu, Qihao Zhu, Shirong Ma, Peiyi Wang, Xiao Bi, Xiaokang Zhang, Xingkai Yu, Yu~Wu, Z.~F. Wu, Zhibin Gou, Zhihong Shao, Zhuoshu Li, Ziyi Gao, Aixin Liu, Bing Xue, Bingxuan Wang, Bochao Wu, Bei Feng, Chengda Lu, Chenggang Zhao, Chengqi Deng, Chenyu Zhang, Chong Ruan, Damai Dai, Deli Chen, Dongjie Ji, Erhang Li, Fangyun Lin, Fucong Dai, Fuli Luo, Guangbo Hao, Guanting Chen, Guowei Li, H.~Zhang, Han Bao, Hanwei Xu, Haocheng Wang, Honghui Ding, Huajian Xin, Huazuo Gao, Hui Qu, Hui Li, Jianzhong Guo, Jiashi Li, Jiawei Wang, Jingchang Chen, Jingyang Yuan, Junjie Qiu, Junlong Li, J.~L. Cai, Jiaqi Ni, Jian Liang, Jin Chen, Kai Dong, Kai Hu, Kaige Gao, Kang Guan, Kexin Huang, Kuai Yu, Lean Wang, Lecong Zhang, Liang Zhao, Litong Wang, Liyue Zhang, Lei Xu, Leyi Xia, Mingchuan Zhang, Minghua Zhang, Minghui Tang, Meng Li, Miaojun Wang, Mingming Li, Ning Tian, Panpan Huang, Peng Zhang, Qiancheng Wang, Qinyu Chen, Qiushi Du, Ruiqi Ge, Ruisong Zhang, Ruizhe Pan, Runji Wang, R.~J. Chen, R.~L. Jin, Ruyi Chen, Shanghao Lu, Shangyan Zhou, Shanhuang Chen, Shengfeng Ye, Shiyu Wang, Shuiping Yu, Shunfeng Zhou, Shuting Pan, S.~S. Li, Shuang Zhou, Shaoqing Wu, Shengfeng Ye, Tao Yun, Tian Pei, Tianyu Sun, T.~Wang, Wangding Zeng, Wanjia Zhao, Wen Liu, Wenfeng Liang, Wenjun Gao, Wenqin Yu, Wentao Zhang, W.~L. Xiao, Wei An, Xiaodong Liu, Xiaohan Wang, Xiaokang Chen, Xiaotao Nie, Xin Cheng, Xin Liu, Xin Xie, Xingchao Liu, Xinyu Yang, Xinyuan Li, Xuecheng Su, Xuheng Lin, X.~Q. Li, Xiangyue Jin, Xiaojin Shen, Xiaosha Chen, Xiaowen Sun, Xiaoxiang Wang, Xinnan Song, Xinyi Zhou, Xianzu Wang, Xinxia Shan, Y.~K. Li, Y.~Q. Wang, Y.~X. Wei, Yang Zhang, Yanhong Xu, Yao Li, Yao Zhao, Yaofeng Sun, Yaohui Wang, Yi~Yu, Yichao Zhang, Yifan Shi, Yiliang Xiong, Ying He, Yishi Piao, Yisong Wang, Yixuan Tan, Yiyang Ma, Yiyuan Liu, Yongqiang Guo, Yuan Ou, Yuduan Wang, Yue Gong, Yuheng Zou, Yujia He, Yunfan Xiong, Yuxiang Luo, Yuxiang You, Yuxuan Liu, Yuyang Zhou, Y.~X. Zhu, Yanhong Xu, Yanping Huang, Yaohui Li, Yi~Zheng, Yuchen Zhu, Yunxian Ma, Ying Tang, Yukun Zha, Yuting Yan, Z.~Z. Ren, Zehui Ren, Zhangli Sha, Zhe Fu, Zhean Xu, Zhenda Xie, Zhengyan Zhang, Zhewen Hao, Zhicheng Ma, Zhigang Yan, Zhiyu Wu, Zihui Gu, Zijia Zhu, Zijun Liu, Zilin Li, Ziwei Xie, Ziyang Song, Zizheng Pan, Zhen Huang, Zhipeng Xu, Zhongyu Zhang, and Zhen Zhang.
\newblock Deepseek-r1: Incentivizing reasoning capability in llms via reinforcement learning.
\newblock \emph{arXiv preprint arXiv: 2501.12948}, 2025.

\bibitem[Feng et~al.(2019)Feng, He, Wang, Luo, Liu, and Chua]{feng2019temporal}
Fuli Feng, Xiangnan He, Xiang Wang, Cheng Luo, Yiqun Liu, and Tat-Seng Chua.
\newblock Temporal relational ranking for stock prediction.
\newblock \emph{ACM Transactions on Information Systems (TOIS)}, 37\penalty0 (2):\penalty0 1--30, 2019.
\newblock URL \url{https://doi.org/10.1145/3309547}.

\bibitem[Gao et~al.(2023)Gao, Schulman, and Hilton]{gao2023scaling}
Leo Gao, John Schulman, and Jacob Hilton.
\newblock Scaling laws for reward model overoptimization.
\newblock In \emph{International Conference on Machine Learning}, pp.\  10835--10866. PMLR, 2023.

\bibitem[Gou et~al.(2024)Gou, Shao, Gong, yelong shen, Yang, Duan, and Chen]{gou2024critic}
Zhibin Gou, Zhihong Shao, Yeyun Gong, yelong shen, Yujiu Yang, Nan Duan, and Weizhu Chen.
\newblock {CRITIC}: Large language models can self-correct with tool-interactive critiquing.
\newblock In \emph{The Twelfth International Conference on Learning Representations}, 2024.
\newblock URL \url{https://openreview.net/forum?id=Sx038qxjek}.

\bibitem[Gulcehre et~al.(2023)Gulcehre, Paine, Srinivasan, Konyushkova, Weerts, Sharma, Siddhant, Ahern, Wang, Gu, Macherey, Doucet, Firat, and de~Freitas]{gulcehre2023reinforcedselftrainingrestlanguage}
Caglar Gulcehre, Tom~Le Paine, Srivatsan Srinivasan, Ksenia Konyushkova, Lotte Weerts, Abhishek Sharma, Aditya Siddhant, Alex Ahern, Miaosen Wang, Chenjie Gu, Wolfgang Macherey, Arnaud Doucet, Orhan Firat, and Nando de~Freitas.
\newblock Reinforced self-training (rest) for language modeling, 2023.
\newblock URL \url{https://arxiv.org/abs/2308.08998}.

\bibitem[Hao et~al.(2021)Hao, Kung, Chang, and Ou]{hao2021predicting}
Pei-Yi Hao, Chien-Feng Kung, Chun-Yang Chang, and Jen-Bing Ou.
\newblock Predicting stock price trends based on financial news articles and using a novel twin support vector machine with fuzzy hyperplane.
\newblock \emph{Applied Soft Computing}, 98:\penalty0 106806, 2021.
\newblock URL \url{https://www.sciencedirect.com/science/article/pii/S1568494620307444}.

\bibitem[Hao et~al.(2023)Hao, Gu, Ma, Hong, Wang, Wang, and Hu]{hao-etal-2023-reasoning}
Shibo Hao, Yi~Gu, Haodi Ma, Joshua Hong, Zhen Wang, Daisy Wang, and Zhiting Hu.
\newblock Reasoning with language model is planning with world model.
\newblock In Houda Bouamor, Juan Pino, and Kalika Bali (eds.), \emph{Proceedings of the 2023 Conference on Empirical Methods in Natural Language Processing}, pp.\  8154--8173, Singapore, December 2023. Association for Computational Linguistics.
\newblock \doi{10.18653/v1/2023.emnlp-main.507}.
\newblock URL \url{https://aclanthology.org/2023.emnlp-main.507}.

\bibitem[Hou et~al.(2024)Hou, Du, Niu, Du, Zeng, Liu, Huang, Wang, Tang, and Dong]{hou2024doesrlhfscaleexploring}
Zhenyu Hou, Pengfan Du, Yilin Niu, Zhengxiao Du, Aohan Zeng, Xiao Liu, Minlie Huang, Hongning Wang, Jie Tang, and Yuxiao Dong.
\newblock Does rlhf scale? exploring the impacts from data, model, and method, 2024.
\newblock URL \url{https://arxiv.org/abs/2412.06000}.

\bibitem[Hsu et~al.(2021)Hsu, Tsai, and Li]{hsu2021fingat}
Yi-Ling Hsu, Yu-Che Tsai, and Cheng-Te Li.
\newblock Fingat: Financial graph attention networks for recommending top-$ k $ k profitable stocks.
\newblock \emph{IEEE Transactions on Knowledge and Data Engineering}, 35\penalty0 (1):\penalty0 469--481, 2021.
\newblock \doi{10.1109/TKDE.2021.3079496}.

\bibitem[Huang et~al.(2024)Huang, Chen, Mishra, Zheng, Yu, Song, and Zhou]{huang2024large}
Jie Huang, Xinyun Chen, Swaroop Mishra, Huaixiu~Steven Zheng, Adams~Wei Yu, Xinying Song, and Denny Zhou.
\newblock Large language models cannot self-correct reasoning yet.
\newblock In \emph{The Twelfth International Conference on Learning Representations}, 2024.
\newblock URL \url{https://openreview.net/forum?id=IkmD3fKBPQ}.

\bibitem[Kirkpatrick et~al.(2017)Kirkpatrick, Pascanu, Rabinowitz, Veness, Desjardins, Rusu, Milan, Quan, Ramalho, Grabska-Barwinska, Hassabis, Clopath, Kumaran, and Hadsell]{catastrophic_forgetting}
James Kirkpatrick, Razvan Pascanu, Neil Rabinowitz, Joel Veness, Guillaume Desjardins, Andrei~A. Rusu, Kieran Milan, John Quan, Tiago Ramalho, Agnieszka Grabska-Barwinska, Demis Hassabis, Claudia Clopath, Dharshan Kumaran, and Raia Hadsell.
\newblock Overcoming catastrophic forgetting in neural networks.
\newblock \emph{Proceedings of the National Academy of Sciences}, 114\penalty0 (13):\penalty0 3521--3526, 2017.
\newblock \doi{10.1073/pnas.1611835114}.
\newblock URL \url{https://www.pnas.org/doi/abs/10.1073/pnas.1611835114}.

\bibitem[Li et~al.(2025)Li, Cao, Cao, Li, Tan, Keutzer, Xing, Gonzalez, and Stoica]{S*}
Dacheng Li, Shiyi Cao, Chengkun Cao, Xiuyu Li, Shangyin Tan, Kurt Keutzer, Jiarong Xing, Joseph~E Gonzalez, and Ion Stoica.
\newblock S*: Test time scaling for code generation.
\newblock \emph{arXiv preprint arXiv:2502.14382}, 2025.

\bibitem[Li et~al.(2024{\natexlab{a}})Li, Zhang, Yu, Fu, and Ye]{li2024agentsneed}
Junyou Li, Qin Zhang, Yangbin Yu, Qiang Fu, and Deheng Ye.
\newblock More agents is all you need, 2024{\natexlab{a}}.
\newblock URL \url{https://arxiv.org/abs/2402.05120}.

\bibitem[Li et~al.(2024{\natexlab{b}})Li, Chen, Chen, Zhang, Su, Xing, and Zhang]{li2024confidencemattersrevisitingintrinsic}
Loka Li, Zhenhao Chen, Guangyi Chen, Yixuan Zhang, Yusheng Su, Eric Xing, and Kun Zhang.
\newblock Confidence matters: Revisiting intrinsic self-correction capabilities of large language models, 2024{\natexlab{b}}.
\newblock URL \url{https://arxiv.org/abs/2402.12563}.

\bibitem[Li et~al.(2021)Li, Bao, Harimoto, Chen, Xu, and Su]{li2021modeling}
Wei Li, Ruihan Bao, Keiko Harimoto, Deli Chen, Jingjing Xu, and Qi~Su.
\newblock Modeling the stock relation with graph network for overnight stock movement prediction.
\newblock In \emph{Proceedings of the twenty-ninth international conference on international joint conferences on artificial intelligence}, pp.\  4541--4547, 2021.
\newblock URL \url{https://doi.org/10.24963/ijcai.2020/626}.

\bibitem[Li et~al.(2023)Li, Lin, Zhang, Fu, Chen, Lou, and Chen]{li-etal-2023-making}
Yifei Li, Zeqi Lin, Shizhuo Zhang, Qiang Fu, Bei Chen, Jian-Guang Lou, and Weizhu Chen.
\newblock Making language models better reasoners with step-aware verifier.
\newblock In Anna Rogers, Jordan Boyd-Graber, and Naoaki Okazaki (eds.), \emph{Proceedings of the 61st Annual Meeting of the Association for Computational Linguistics (Volume 1: Long Papers)}, pp.\  5315--5333, Toronto, Canada, July 2023. Association for Computational Linguistics.
\newblock \doi{10.18653/v1/2023.acl-long.291}.
\newblock URL \url{https://aclanthology.org/2023.acl-long.291}.

\bibitem[Lin et~al.(2024)Lin, Fu, Liu, Li, Gong, Wan, Zhang, Wang, Zhang, and Gai]{lin-etal-2024-just}
Lei Lin, Jiayi Fu, Pengli Liu, Qingyang Li, Yan Gong, Junchen Wan, Fuzheng Zhang, Zhongyuan Wang, Di~Zhang, and Kun Gai.
\newblock Just ask one more time! self-agreement improves reasoning of language models in (almost) all scenarios.
\newblock In Lun-Wei Ku, Andre Martins, and Vivek Srikumar (eds.), \emph{Findings of the Association for Computational Linguistics: ACL 2024}, pp.\  3829--3852, Bangkok, Thailand, August 2024. Association for Computational Linguistics.
\newblock \doi{10.18653/v1/2024.findings-acl.230}.
\newblock URL \url{https://aclanthology.org/2024.findings-acl.230}.

\bibitem[Liu et~al.(2025{\natexlab{a}})Liu, Yao, Min, Cao, Hou, and Li]{liu2025pairwise}
Yantao Liu, Zijun Yao, Rui Min, Yixin Cao, Lei Hou, and Juanzi Li.
\newblock Pairwise rm: Perform best-of-n sampling with knockout tournament.
\newblock \emph{arXiv preprint arXiv:2501.13007}, 2025{\natexlab{a}}.

\bibitem[Liu et~al.(2025{\natexlab{b}})Liu, Guo, Lou, Zeng, Niu, Wang, Xu, Cai, Yang, Zhao, Li, Xu, Chen, Chen, Bai, and Zhang]{Fin-R1}
Zhaowei Liu, Xin Guo, Fangqi Lou, Lingfeng Zeng, Jinyi Niu, Zixuan Wang, Jiajie Xu, Weige Cai, Ziwei Yang, Xueqian Zhao, Chao Li, Sheng Xu, Dezhi Chen, Yun Chen, Zuo Bai, and Liwen Zhang.
\newblock Fin-r1: A large language model for financial reasoning through reinforcement learning.
\newblock \emph{arXiv preprint arXiv: 2503.16252}, 2025{\natexlab{b}}.

\bibitem[Long(2023)]{long2023largelanguagemodelguided}
Jieyi Long.
\newblock Large language model guided tree-of-thought, 2023.
\newblock URL \url{https://arxiv.org/abs/2305.08291}.

\bibitem[Lu et~al.(2025)Lu, Guo, Zhang, Zhu, and Liu]{lu2025bizfinbench0}
Guilong Lu, Xuntao Guo, Rongjunchen Zhang, Wenqiao Zhu, and Ji~Liu.
\newblock Bizfinbench: A business-driven real-world financial benchmark for evaluating llms.
\newblock \emph{arXiv preprint arXiv: 2505.19457}, 2025.

\bibitem[Luo et~al.(2023)Luo, Liao, Li, Cheng, and Yan]{CMIN}
Di~Luo, Weiheng Liao, Shuqi Li, Xin Cheng, and Rui Yan.
\newblock Causality-guided multi-memory interaction network for multivariate stock price movement prediction.
\newblock In \emph{Proceedings of the 61st Annual Meeting of the Association for Computational Linguistics (Volume 1: Long Papers)}, pp.\  12164--12176, Toronto, Canada, July 2023. Association for Computational Linguistics.
\newblock \doi{10.18653/v1/2023.acl-long.679}.
\newblock URL \url{https://aclanthology.org/2023.acl-long.679}.

\bibitem[Mahfooz et~al.(2022)Mahfooz, Ali, and Khan]{Mahfooz2022ImprovingST}
S.~Z. Mahfooz, Iftikhar Ali, and Muhammad~Navaid Khan.
\newblock Improving stock trend prediction using lstm neural network trained on a complex trading strategy.
\newblock \emph{International Journal for Research in Applied Science and Engineering Technology}, 2022.
\newblock URL \url{https://api.semanticscholar.org/CorpusID:251176332}.

\bibitem[Manvi et~al.(2024)Manvi, Singh, and Ermon]{manvi2024adaptiveinferencetimecomputellms}
Rohin Manvi, Anikait Singh, and Stefano Ermon.
\newblock Adaptive inference-time compute: Llms can predict if they can do better, even mid-generation, 2024.
\newblock URL \url{https://arxiv.org/abs/2410.02725}.

\bibitem[Muennighoff et~al.(2025)Muennighoff, Yang, Shi, Li, Fei-Fei, Hajishirzi, Zettlemoyer, Liang, Candès, and Hashimoto]{s1}
Niklas Muennighoff, Zitong Yang, Weijia Shi, Xiang~Lisa Li, Li~Fei-Fei, Hannaneh Hajishirzi, Luke Zettlemoyer, Percy Liang, Emmanuel Candès, and Tatsunori Hashimoto.
\newblock s1: Simple test-time scaling.
\newblock \emph{arXiv preprint arXiv: 2501.19393}, 2025.

\bibitem[Nakano et~al.(2022)Nakano, Hilton, Balaji, Wu, Ouyang, Kim, Hesse, Jain, Kosaraju, Saunders, Jiang, Cobbe, Eloundou, Krueger, Button, Knight, Chess, and Schulman]{nakano2022webgptbrowserassistedquestionansweringhuman}
Reiichiro Nakano, Jacob Hilton, Suchir Balaji, Jeff Wu, Long Ouyang, Christina Kim, Christopher Hesse, Shantanu Jain, Vineet Kosaraju, William Saunders, Xu~Jiang, Karl Cobbe, Tyna Eloundou, Gretchen Krueger, Kevin Button, Matthew Knight, Benjamin Chess, and John Schulman.
\newblock Webgpt: Browser-assisted question-answering with human feedback, 2022.
\newblock URL \url{https://arxiv.org/abs/2112.09332}.

\bibitem[Nguyen et~al.(2015)Nguyen, Shirai, and Velcin]{nguyen2015sentiment}
Thien~Hai Nguyen, Kiyoaki Shirai, and Julien Velcin.
\newblock Sentiment analysis on social media for stock movement prediction.
\newblock \emph{Expert Systems with Applications}, 42\penalty0 (24):\penalty0 9603--9611, 2015.
\newblock URL \url{https://www.sciencedirect.com/science/article/pii/S0957417415005126}.

\bibitem[Olausson et~al.(2024)Olausson, Inala, Wang, Gao, and Solar-Lezama]{olausson2024is}
Theo~X. Olausson, Jeevana~Priya Inala, Chenglong Wang, Jianfeng Gao, and Armando Solar-Lezama.
\newblock Is self-repair a silver bullet for code generation?
\newblock In \emph{The Twelfth International Conference on Learning Representations}, 2024.
\newblock URL \url{https://openreview.net/forum?id=y0GJXRungR}.

\bibitem[OpenAI et~al.(2024)OpenAI, :, Jaech, Kalai, Lerer, Richardson, El-Kishky, Low, Helyar, Madry, Beutel, Carney, Iftimie, Karpenko, Passos, Neitz, Prokofiev, Wei, Tam, Bennett, Kumar, Saraiva, Vallone, Duberstein, Kondrich, Mishchenko, Applebaum, Jiang, Nair, Zoph, Ghorbani, Rossen, Sokolowsky, Barak, McGrew, Minaiev, Hao, Baker, Houghton, McKinzie, Eastman, Lugaresi, Bassin, Hudson, Li, de~Bourcy, Voss, Shen, Zhang, Koch, Orsinger, Hesse, Fischer, Chan, Roberts, Kappler, Levy, Selsam, Dohan, Farhi, Mely, Robinson, Tsipras, Li, Oprica, Freeman, Zhang, Wong, Proehl, Cheung, Mitchell, Wallace, Ritter, Mays, Wang, Such, Raso, Leoni, Tsimpourlas, Song, von Lohmann, Sulit, Salmon, Parascandolo, Chabot, Zhao, Brockman, Leclerc, Salman, Bao, Sheng, Andrin, Bagherinezhad, Ren, Lightman, Chung, Kivlichan, O'Connell, Osband, Gilaberte, Akkaya, Kostrikov, Sutskever, Kofman, Pachocki, Lennon, Wei, Harb, Twore, Feng, Yu, Weng, Tang, Yu, Candela, Palermo, Parish, Heidecke, Hallman, Rizzo, Gordon, Uesato, Ward, Huizinga, Wang, Chen, Xiao, Singhal, Nguyen, Cobbe, Shi, Wood, Rimbach, Gu-Lemberg, Liu, Lu, Stone, Yu, Ahmad, Yang, Liu, Maksin, Ho, Fedus, Weng, Li, McCallum, Held, Kuhn, Kondraciuk, Kaiser, Metz, Boyd, Trebacz, Joglekar, Chen, Tintor, Meyer, Jones, Kaufer, Schwarzer, Shah, Yatbaz, Guan, Xu, Yan, Glaese, Chen, Lampe, Malek, Wang, Fradin, McClay, Pavlov, Wang, Wang, Murati, Bavarian, Rohaninejad, McAleese, Chowdhury, Chowdhury, Ryder, Tezak, Brown, Nachum, Boiko, Murk, Watkins, Chao, Ashbourne, Izmailov, Zhokhov, Dias, Arora, Lin, Lopes, Gaon, Miyara, Leike, Hwang, Garg, Brown, James, Shu, Cheu, Greene, Jain, Altman, Toizer, Toyer, Miserendino, Agarwal, Hernandez, Baker, McKinney, Yan, Zhao, Hu, Santurkar, Chaudhuri, Zhang, Fu, Papay, Lin, Balaji, Sanjeev, Sidor, Broda, Clark, Wang, Gordon, Sanders, Patwardhan, Sottiaux, Degry, Dimson, Zheng, Garipov, Stasi, Bansal, Creech, Peterson, Eloundou, Qi, Kosaraju, Monaco, Pong, Fomenko, Zheng, Zhou, McCabe, Zaremba, Dubois, Lu, Chen, Cha, Bai, He, Zhang, Wang, Shao, and Li]{o1}
OpenAI, :, Aaron Jaech, Adam Kalai, Adam Lerer, Adam Richardson, Ahmed El-Kishky, Aiden Low, Alec Helyar, Aleksander Madry, Alex Beutel, Alex Carney, Alex Iftimie, Alex Karpenko, Alex~Tachard Passos, Alexander Neitz, Alexander Prokofiev, Alexander Wei, Allison Tam, Ally Bennett, Ananya Kumar, Andre Saraiva, Andrea Vallone, Andrew Duberstein, Andrew Kondrich, Andrey Mishchenko, Andy Applebaum, Angela Jiang, Ashvin Nair, Barret Zoph, Behrooz Ghorbani, Ben Rossen, Benjamin Sokolowsky, Boaz Barak, Bob McGrew, Borys Minaiev, Botao Hao, Bowen Baker, Brandon Houghton, Brandon McKinzie, Brydon Eastman, Camillo Lugaresi, Cary Bassin, Cary Hudson, Chak~Ming Li, Charles de~Bourcy, Chelsea Voss, Chen Shen, Chong Zhang, Chris Koch, Chris Orsinger, Christopher Hesse, Claudia Fischer, Clive Chan, Dan Roberts, Daniel Kappler, Daniel Levy, Daniel Selsam, David Dohan, David Farhi, David Mely, David Robinson, Dimitris Tsipras, Doug Li, Dragos Oprica, Eben Freeman, Eddie Zhang, Edmund Wong, Elizabeth Proehl, Enoch Cheung, Eric Mitchell, Eric Wallace, Erik Ritter, Evan Mays, Fan Wang, Felipe~Petroski Such, Filippo Raso, Florencia Leoni, Foivos Tsimpourlas, Francis Song, Fred von Lohmann, Freddie Sulit, Geoff Salmon, Giambattista Parascandolo, Gildas Chabot, Grace Zhao, Greg Brockman, Guillaume Leclerc, Hadi Salman, Haiming Bao, Hao Sheng, Hart Andrin, Hessam Bagherinezhad, Hongyu Ren, Hunter Lightman, Hyung~Won Chung, Ian Kivlichan, Ian O'Connell, Ian Osband, Ignasi~Clavera Gilaberte, Ilge Akkaya, Ilya Kostrikov, Ilya Sutskever, Irina Kofman, Jakub Pachocki, James Lennon, Jason Wei, Jean Harb, Jerry Twore, Jiacheng Feng, Jiahui Yu, Jiayi Weng, Jie Tang, Jieqi Yu, Joaquin~Quiñonero Candela, Joe Palermo, Joel Parish, Johannes Heidecke, John Hallman, John Rizzo, Jonathan Gordon, Jonathan Uesato, Jonathan Ward, Joost Huizinga, Julie Wang, Kai Chen, Kai Xiao, Karan Singhal, Karina Nguyen, Karl Cobbe, Katy Shi, Kayla Wood, Kendra Rimbach, Keren Gu-Lemberg, Kevin Liu, Kevin Lu, Kevin Stone, Kevin Yu, Lama Ahmad, Lauren Yang, Leo Liu, Leon Maksin, Leyton Ho, Liam Fedus, Lilian Weng, Linden Li, Lindsay McCallum, Lindsey Held, Lorenz Kuhn, Lukas Kondraciuk, Lukasz Kaiser, Luke Metz, Madelaine Boyd, Maja Trebacz, Manas Joglekar, Mark Chen, Marko Tintor, Mason Meyer, Matt Jones, Matt Kaufer, Max Schwarzer, Meghan Shah, Mehmet Yatbaz, Melody~Y. Guan, Mengyuan Xu, Mengyuan Yan, Mia Glaese, Mianna Chen, Michael Lampe, Michael Malek, Michele Wang, Michelle Fradin, Mike McClay, Mikhail Pavlov, Miles Wang, Mingxuan Wang, Mira Murati, Mo~Bavarian, Mostafa Rohaninejad, Nat McAleese, Neil Chowdhury, Neil Chowdhury, Nick Ryder, Nikolas Tezak, Noam Brown, Ofir Nachum, Oleg Boiko, Oleg Murk, Olivia Watkins, Patrick Chao, Paul Ashbourne, Pavel Izmailov, Peter Zhokhov, Rachel Dias, Rahul Arora, Randall Lin, Rapha~Gontijo Lopes, Raz Gaon, Reah Miyara, Reimar Leike, Renny Hwang, Rhythm Garg, Robin Brown, Roshan James, Rui Shu, Ryan Cheu, Ryan Greene, Saachi Jain, Sam Altman, Sam Toizer, Sam Toyer, Samuel Miserendino, Sandhini Agarwal, Santiago Hernandez, Sasha Baker, Scott McKinney, Scottie Yan, Shengjia Zhao, Shengli Hu, Shibani Santurkar, Shraman~Ray Chaudhuri, Shuyuan Zhang, Siyuan Fu, Spencer Papay, Steph Lin, Suchir Balaji, Suvansh Sanjeev, Szymon Sidor, Tal Broda, Aidan Clark, Tao Wang, Taylor Gordon, Ted Sanders, Tejal Patwardhan, Thibault Sottiaux, Thomas Degry, Thomas Dimson, Tianhao Zheng, Timur Garipov, Tom Stasi, Trapit Bansal, Trevor Creech, Troy Peterson, Tyna Eloundou, Valerie Qi, Vineet Kosaraju, Vinnie Monaco, Vitchyr Pong, Vlad Fomenko, Weiyi Zheng, Wenda Zhou, Wes McCabe, Wojciech Zaremba, Yann Dubois, Yinghai Lu, Yining Chen, Young Cha, Yu~Bai, Yuchen He, Yuchen Zhang, Yunyun Wang, Zheng Shao, and Zhuohan Li.
\newblock Openai o1 system card.
\newblock \emph{arXiv preprint arXiv: 2412.16720}, 2024.

\bibitem[OpenAI et~al.(2025)OpenAI, :, Agarwal, Ahmad, Ai, Altman, Applebaum, Arbus, Arora, Bai, Baker, Bao, Barak, Bennett, Bertao, Brett, Brevdo, Brockman, Bubeck, Chang, Chen, Chen, Cheung, Clark, Cook, Dukhan, Dvorak, Fives, Fomenko, Garipov, Georgiev, Glaese, Gogineni, Goucher, Gross, Guzman, Hallman, Hehir, Heidecke, Helyar, Hu, Huet, Huh, Jain, Johnson, Koch, Kofman, Kundel, Kwon, Kyrylov, Le, Leclerc, Lennon, Lessans, Lezcano-Casado, Li, Li, Lin, Liss, Lily, Liu, Liu, Lu, Lu, Martinovic, McCallum, McGrath, McKinney, McLaughlin, Mei, Mostovoy, Mu, Myles, Neitz, Nichol, Pachocki, Paino, Palmie, Pantuliano, Parascandolo, Park, Pathak, Paz, Peran, Pimenov, Pokrass, Proehl, Qiu, Raila, Raso, Ren, Richardson, Robinson, Rotsted, Salman, Sanjeev, Schwarzer, Sculley, Sikchi, Simon, Singhal, Song, Stuckey, Sun, Tillet, Toizer, Tsimpourlas, Vyas, Wallace, Wang, Wang, Watkins, Weil, Wendling, Whinnery, Whitney, Wong, Yang, Yang, Yasunaga, Ying, Zaremba, Zhan, Zhang, Zhang, Zhang, and Zhao]{openai2025gpt0oss0120b}
OpenAI, :, Sandhini Agarwal, Lama Ahmad, Jason Ai, Sam Altman, Andy Applebaum, Edwin Arbus, Rahul~K. Arora, Yu~Bai, Bowen Baker, Haiming Bao, Boaz Barak, Ally Bennett, Tyler Bertao, Nivedita Brett, Eugene Brevdo, Greg Brockman, Sebastien Bubeck, Che Chang, Kai Chen, Mark Chen, Enoch Cheung, Aidan Clark, Dan Cook, Marat Dukhan, Casey Dvorak, Kevin Fives, Vlad Fomenko, Timur Garipov, Kristian Georgiev, Mia Glaese, Tarun Gogineni, Adam Goucher, Lukas Gross, Katia~Gil Guzman, John Hallman, Jackie Hehir, Johannes Heidecke, Alec Helyar, Haitang Hu, Romain Huet, Jacob Huh, Saachi Jain, Zach Johnson, Chris Koch, Irina Kofman, Dominik Kundel, Jason Kwon, Volodymyr Kyrylov, Elaine~Ya Le, Guillaume Leclerc, James~Park Lennon, Scott Lessans, Mario Lezcano-Casado, Yuanzhi Li, Zhuohan Li, Ji~Lin, Jordan Liss, Lily, Liu, Jiancheng Liu, Kevin Lu, Chris Lu, Zoran Martinovic, Lindsay McCallum, Josh McGrath, Scott McKinney, Aidan McLaughlin, Song Mei, Steve Mostovoy, Tong Mu, Gideon Myles, Alexander Neitz, Alex Nichol, Jakub Pachocki, Alex Paino, Dana Palmie, Ashley Pantuliano, Giambattista Parascandolo, Jongsoo Park, Leher Pathak, Carolina Paz, Ludovic Peran, Dmitry Pimenov, Michelle Pokrass, Elizabeth Proehl, Huida Qiu, Gaby Raila, Filippo Raso, Hongyu Ren, Kimmy Richardson, David Robinson, Bob Rotsted, Hadi Salman, Suvansh Sanjeev, Max Schwarzer, D.~Sculley, Harshit Sikchi, Kendal Simon, Karan Singhal, Yang Song, Dane Stuckey, Zhiqing Sun, Philippe Tillet, Sam Toizer, Foivos Tsimpourlas, Nikhil Vyas, Eric Wallace, Xin Wang, Miles Wang, Olivia Watkins, Kevin Weil, Amy Wendling, Kevin Whinnery, Cedric Whitney, Hannah Wong, Lin Yang, Yu~Yang, Michihiro Yasunaga, Kristen Ying, Wojciech Zaremba, Wenting Zhan, Cyril Zhang, Brian Zhang, Eddie Zhang, and Shengjia Zhao.
\newblock gpt-oss-120b \& gpt-oss-20b model card.
\newblock \emph{arXiv preprint arXiv: 2508.10925}, 2025.

\bibitem[Qian et~al.(2025)Qian, Zhou, Wang, Peng, Huang, and Xie]{Fino1}
Lingfei Qian, Weipeng Zhou, Yan Wang, Xueqing Peng, Jimin Huang, and Qianqian Xie.
\newblock Fino1: On the transferability of reasoning enhanced llms to finance.
\newblock \emph{ArXiv}, abs/2502.08127, 2025.
\newblock URL \url{https://api.semanticscholar.org/CorpusID:276287200}.

\bibitem[Qiu et~al.(2024)Qiu, Lu, Zeng, Guo, Geng, Wang, Huang, Wu, and Wang]{qiu2024treebonenhancinginferencetimealignment}
Jiahao Qiu, Yifu Lu, Yifan Zeng, Jiacheng Guo, Jiayi Geng, Huazheng Wang, Kaixuan Huang, Yue Wu, and Mengdi Wang.
\newblock Treebon: Enhancing inference-time alignment with speculative tree-search and best-of-n sampling, 2024.
\newblock URL \url{https://arxiv.org/abs/2410.16033}.

\bibitem[Rasley et~al.(2020)Rasley, Rajbhandari, Ruwase, and He]{deepspeed}
Jeff Rasley, Samyam Rajbhandari, Olatunji Ruwase, and Yuxiong He.
\newblock Deepspeed: System optimizations enable training deep learning models with over 100 billion parameters.
\newblock In \emph{Proceedings of the 26th ACM SIGKDD International Conference on Knowledge Discovery \& Data Mining}, KDD '20, pp.\  3505–3506, New York, NY, USA, 2020. Association for Computing Machinery.
\newblock ISBN 9781450379984.
\newblock \doi{10.1145/3394486.3406703}.
\newblock URL \url{https://doi.org/10.1145/3394486.3406703}.

\bibitem[Shao et~al.(2024)Shao, Wang, Zhu, Xu, Song, Zhang, Li, Wu, and Guo]{deepseek-math}
Zhihong Shao, Peiyi Wang, Qihao Zhu, Runxin Xu, Junxiao Song, Mingchuan Zhang, Y.K. Li, Y.~Wu, and Daya Guo.
\newblock Deepseekmath: Pushing the limits of mathematical reasoning in open language models.
\newblock \emph{CoRR}, abs/2402.03300, 2024.
\newblock URL \url{https://arxiv.org/abs/2402.03300}.

\bibitem[Sheng et~al.(2025)Sheng, Zhang, Ye, Wu, Zhang, Zhang, Peng, Lin, and Wu]{verl}
Guangming Sheng, Chi Zhang, Zilingfeng Ye, Xibin Wu, Wang Zhang, Ru~Zhang, Yanghua Peng, Haibin Lin, and Chuan Wu.
\newblock Hybridflow: A flexible and efficient rlhf framework.
\newblock In \emph{Proceedings of the Twentieth European Conference on Computer Systems}, EuroSys '25, pp.\  1279–1297, New York, NY, USA, 2025. Association for Computing Machinery.
\newblock ISBN 9798400711961.
\newblock \doi{10.1145/3689031.3696075}.
\newblock URL \url{https://doi.org/10.1145/3689031.3696075}.

\bibitem[Shi et~al.(2020)Shi, Zheng, Guo, and Ren]{Shi2020StockMP}
Yong Shi, Yuanchun Zheng, Kun Guo, and Xinyue Ren.
\newblock Stock movement prediction with sentiment analysis based on deep learning networks.
\newblock \emph{Concurrency and Computation: Practice and Experience}, 33, 2020.
\newblock URL \url{https://api.semanticscholar.org/CorpusID:228848908}.

\bibitem[Shinn et~al.(2023)Shinn, Cassano, Berman, Gopinath, Narasimhan, and Yao]{shinn2023reflexionlanguageagentsverbal}
Noah Shinn, Federico Cassano, Edward Berman, Ashwin Gopinath, Karthik Narasimhan, and Shunyu Yao.
\newblock Reflexion: Language agents with verbal reinforcement learning, 2023.
\newblock URL \url{https://arxiv.org/abs/2303.11366}.

\bibitem[Snell et~al.(2024)Snell, Lee, Xu, and Kumar]{snell2024scaling}
Charlie Snell, Jaehoon Lee, Kelvin Xu, and Aviral Kumar.
\newblock Scaling llm test-time compute optimally can be more effective than scaling model parameters.
\newblock \emph{arXiv preprint arXiv: 2408.03314}, 2024.

\bibitem[Song et~al.(2025)Song, Wu, Wang, Liu, Su, and Zheng]{song2025progcoprogramhelpsselfcorrection}
Xiaoshuai Song, Yanan Wu, Weixun Wang, Jiaheng Liu, Wenbo Su, and Bo~Zheng.
\newblock Progco: Program helps self-correction of large language models, 2025.
\newblock URL \url{https://arxiv.org/abs/2501.01264}.

\bibitem[Stiennon et~al.(2020)Stiennon, Ouyang, Wu, Ziegler, Lowe, Voss, Radford, Amodei, and Christiano]{stiennon2020learning}
Nisan Stiennon, Long Ouyang, Jeffrey Wu, Daniel Ziegler, Ryan Lowe, Chelsea Voss, Alec Radford, Dario Amodei, and Paul~F Christiano.
\newblock Learning to summarize with human feedback.
\newblock In H.~Larochelle, M.~Ranzato, R.~Hadsell, M.F. Balcan, and H.~Lin (eds.), \emph{Advances in Neural Information Processing Systems}, volume~33, pp.\  3008--3021. Curran Associates, Inc., 2020.
\newblock URL \url{https://proceedings.neurips.cc/paper_files/paper/2020/file/1f89885d556929e98d3ef9b86448f951-Paper.pdf}.

\bibitem[Sun et~al.(2024)Sun, Haider, Zhang, Yang, Qiu, Yin, Wang, Bartlett, and Zanette]{sun2024fast}
Hanshi Sun, Momin Haider, Ruiqi Zhang, Huitao Yang, Jiahao Qiu, Ming Yin, Mengdi Wang, Peter Bartlett, and Andrea Zanette.
\newblock Fast best-of-n decoding via speculative rejection.
\newblock In \emph{The Thirty-eighth Annual Conference on Neural Information Processing Systems}, 2024.
\newblock URL \url{https://openreview.net/forum?id=348hfcprUs}.

\bibitem[Ta et~al.(2018)Ta, Liu, and Addis]{Ta2018PredictionAP}
Van-Dai Ta, Chuan-Ming Liu, and Direselign Addis.
\newblock Prediction and portfolio optimization in quantitative trading using machine learning techniques.
\newblock \emph{Proceedings of the 9th International Symposium on Information and Communication Technology}, 2018.
\newblock URL \url{https://api.semanticscholar.org/CorpusID:56450817}.

\bibitem[Toh et~al.(2024)Toh, Ghosal, and Poria]{toh2024votescountprogramsverifiers}
Vernon Y.~H. Toh, Deepanway Ghosal, and Soujanya Poria.
\newblock Not all votes count! programs as verifiers improve self-consistency of language models for math reasoning, 2024.
\newblock URL \url{https://arxiv.org/abs/2410.12608}.

\bibitem[Touvron et~al.(2023)Touvron, Lavril, Izacard, Martinet, Lachaux, Lacroix, Rozière, Goyal, Hambro, Azhar, Rodriguez, Joulin, Grave, and Lample]{LLaMA}
Hugo Touvron, Thibaut Lavril, Gautier Izacard, Xavier Martinet, Marie-Anne Lachaux, Timothée Lacroix, Baptiste Rozière, Naman Goyal, Eric Hambro, Faisal Azhar, Aurelien Rodriguez, Armand Joulin, Edouard Grave, and Guillaume Lample.
\newblock Llama: Open and efficient foundation language models.
\newblock \emph{Arxiv}, 2023.

\bibitem[Uesato et~al.(2022)Uesato, Kushman, Kumar, Song, Siegel, Wang, Creswell, Irving, and Higgins]{uesato2022solving}
Jonathan Uesato, Nate Kushman, Ramana Kumar, Francis Song, Noah Siegel, Lisa Wang, Antonia Creswell, Geoffrey Irving, and Irina Higgins.
\newblock Solving math word problems with process- and outcome-based feedback.
\newblock \emph{arXiv preprint arXiv: 2211.14275}, 2022.

\bibitem[Vargas et~al.(2018)Vargas, Dos~Anjos, Bichara, and Evsukoff]{vargas2018deep}
Manuel~R Vargas, Carlos~EM Dos~Anjos, Gustavo~LG Bichara, and Alexandre~G Evsukoff.
\newblock Deep learning for stock market prediction using technical indicators and financial news articles.
\newblock In \emph{2018 international joint conference on neural networks (IJCNN)}, pp.\  1--8. IEEE, 2018.
\newblock \doi{10.1109/IJCNN.2018.8489208}.

\bibitem[Vui et~al.(2013)Vui, Gan, On, Alfred, and Anthony]{Vui2013ARO}
Chang~Sim Vui, Kim~Soon Gan, Chin~Kim On, R.~Alfred, and Patricia Anthony.
\newblock A review of stock market prediction with artificial neural network (ann).
\newblock \emph{2013 IEEE International Conference on Control System, Computing and Engineering}, pp.\  477--482, 2013.
\newblock URL \url{https://api.semanticscholar.org/CorpusID:9567658}.

\bibitem[Wang et~al.(2024{\natexlab{a}})Wang, Jain, Zhang, Ray, Kumar, and Athiwaratkun]{wang2024reasoningtokeneconomiesbudgetaware}
Junlin Wang, Siddhartha Jain, Dejiao Zhang, Baishakhi Ray, Varun Kumar, and Ben Athiwaratkun.
\newblock Reasoning in token economies: Budget-aware evaluation of llm reasoning strategies, 2024{\natexlab{a}}.
\newblock URL \url{https://arxiv.org/abs/2406.06461}.

\bibitem[Wang et~al.(2024{\natexlab{b}})Wang, Izumi, and Sakaji]{wang-etal-2024-llmfactor}
Meiyun Wang, Kiyoshi Izumi, and Hiroki Sakaji.
\newblock {LLMF}actor: Extracting profitable factors through prompts for explainable stock movement prediction.
\newblock In Lun-Wei Ku, Andre Martins, and Vivek Srikumar (eds.), \emph{Findings of the Association for Computational Linguistics: ACL 2024}, pp.\  3120--3131, Bangkok, Thailand, August 2024{\natexlab{b}}. Association for Computational Linguistics.
\newblock \doi{10.18653/v1/2024.findings-acl.185}.
\newblock URL \url{https://aclanthology.org/2024.findings-acl.185/}.

\bibitem[Wang et~al.(2023)Wang, Wei, Schuurmans, Le, Chi, Narang, Chowdhery, and Zhou]{wang2023selfconsistency}
Xuezhi Wang, Jason Wei, Dale Schuurmans, Quoc~V Le, Ed~H. Chi, Sharan Narang, Aakanksha Chowdhery, and Denny Zhou.
\newblock Self-consistency improves chain of thought reasoning in language models.
\newblock In \emph{The Eleventh International Conference on Learning Representations}, 2023.
\newblock URL \url{https://openreview.net/forum?id=1PL1NIMMrw}.

\bibitem[Xie et~al.(2023)Xie, Kawaguchi, Zhao, Zhao, Kan, He, and Xie]{xie2023selfevaluation}
Yuxi Xie, Kenji Kawaguchi, Yiran Zhao, Xu~Zhao, Min-Yen Kan, Junxian He, and Qizhe Xie.
\newblock Self-evaluation guided beam search for reasoning.
\newblock In \emph{Thirty-seventh Conference on Neural Information Processing Systems}, 2023.
\newblock URL \url{https://openreview.net/forum?id=Bw82hwg5Q3}.

\bibitem[Xu(2023)]{xu2023traingainunleashmathematical}
Haotian Xu.
\newblock No train still gain. unleash mathematical reasoning of large language models with monte carlo tree search guided by energy function, 2023.
\newblock URL \url{https://arxiv.org/abs/2309.03224}.

\bibitem[Xu \& Cohen(2018)Xu and Cohen]{StockNet}
Yumo Xu and Shay~B. Cohen.
\newblock Stock movement prediction from tweets and historical prices.
\newblock In Iryna Gurevych and Yusuke Miyao (eds.), \emph{Proceedings of the 56th Annual Meeting of the Association for Computational Linguistics (Volume 1: Long Papers)}, pp.\  1970--1979, Melbourne, Australia, July 2018. Association for Computational Linguistics.
\newblock \doi{10.18653/v1/P18-1183}.
\newblock URL \url{https://aclanthology.org/P18-1183/}.

\bibitem[Yang et~al.(2025)Yang, Li, Yang, Zhang, Hui, Zheng, Yu, Gao, Huang, Lv, Zheng, Liu, Zhou, Huang, Hu, Ge, Wei, Lin, Tang, Yang, Tu, Zhang, Yang, Yang, Zhou, Zhou, Lin, Dang, Bao, Yang, Yu, Deng, Li, Xue, Li, Zhang, Wang, Zhu, Men, Gao, Liu, Luo, Li, Tang, Yin, Ren, Wang, Zhang, Ren, Fan, Su, Zhang, Zhang, Wan, Liu, Wang, Cui, Zhang, Zhou, and Qiu]{yang2025qwen3technicalreport}
An~Yang, Anfeng Li, Baosong Yang, Beichen Zhang, Binyuan Hui, Bo~Zheng, Bowen Yu, Chang Gao, Chengen Huang, Chenxu Lv, Chujie Zheng, Dayiheng Liu, Fan Zhou, Fei Huang, Feng Hu, Hao Ge, Haoran Wei, Huan Lin, Jialong Tang, Jian Yang, Jianhong Tu, Jianwei Zhang, Jianxin Yang, Jiaxi Yang, Jing Zhou, Jingren Zhou, Junyang Lin, Kai Dang, Keqin Bao, Kexin Yang, Le~Yu, Lianghao Deng, Mei Li, Mingfeng Xue, Mingze Li, Pei Zhang, Peng Wang, Qin Zhu, Rui Men, Ruize Gao, Shixuan Liu, Shuang Luo, Tianhao Li, Tianyi Tang, Wenbiao Yin, Xingzhang Ren, Xinyu Wang, Xinyu Zhang, Xuancheng Ren, Yang Fan, Yang Su, Yichang Zhang, Yinger Zhang, Yu~Wan, Yuqiong Liu, Zekun Wang, Zeyu Cui, Zhenru Zhang, Zhipeng Zhou, and Zihan Qiu.
\newblock Qwen3 technical report, 2025.
\newblock URL \url{https://arxiv.org/abs/2505.09388}.

\bibitem[Yang et~al.(2024)Yang, Zhang, Wang, Xu, Lin, and Sui]{yang2024confidencevscritiquedecomposition}
Zhe Yang, Yichang Zhang, Yudong Wang, Ziyao Xu, Junyang Lin, and Zhifang Sui.
\newblock Confidence v.s. critique: A decomposition of self-correction capability for llms, 2024.
\newblock URL \url{https://arxiv.org/abs/2412.19513}.

\bibitem[Yao et~al.(2023)Yao, Yu, Zhao, Shafran, Griffiths, Cao, and Narasimhan]{NEURIPS2023_271db992}
Shunyu Yao, Dian Yu, Jeffrey Zhao, Izhak Shafran, Tom Griffiths, Yuan Cao, and Karthik Narasimhan.
\newblock Tree of thoughts: Deliberate problem solving with large language models.
\newblock In A.~Oh, T.~Naumann, A.~Globerson, K.~Saenko, M.~Hardt, and S.~Levine (eds.), \emph{Advances in Neural Information Processing Systems}, volume~36, pp.\  11809--11822. Curran Associates, Inc., 2023.
\newblock URL \url{https://proceedings.neurips.cc/paper_files/paper/2023/file/271db9922b8d1f4dd7aaef84ed5ac703-Paper-Conference.pdf}.

\bibitem[Ye \& Ng(2024)Ye and Ng]{ye-ng-2024-preference}
Hai Ye and Hwee~Tou Ng.
\newblock Preference-guided reflective sampling for aligning language models.
\newblock In Yaser Al-Onaizan, Mohit Bansal, and Yun-Nung Chen (eds.), \emph{Proceedings of the 2024 Conference on Empirical Methods in Natural Language Processing}, pp.\  21646--21668, Miami, Florida, USA, November 2024. Association for Computational Linguistics.
\newblock \doi{10.18653/v1/2024.emnlp-main.1206}.
\newblock URL \url{https://aclanthology.org/2024.emnlp-main.1206}.

\bibitem[Zhang et~al.(2024{\natexlab{a}})Zhang, Wu, Lei, Che, Li, Xie, Huang, Zhang, Pavone, Li, Ouyang, and Zhou]{zhang2024llamaberrypairwiseoptimizationo1like}
Di~Zhang, Jianbo Wu, Jingdi Lei, Tong Che, Jiatong Li, Tong Xie, Xiaoshui Huang, Shufei Zhang, Marco Pavone, Yuqiang Li, Wanli Ouyang, and Dongzhan Zhou.
\newblock Llama-berry: Pairwise optimization for o1-like olympiad-level mathematical reasoning, 2024{\natexlab{a}}.
\newblock URL \url{https://arxiv.org/abs/2410.02884}.

\bibitem[Zhang et~al.(2024{\natexlab{b}})Zhang, Haider, Yin, Qiu, Wang, Bartlett, and Zanette]{zhang2024accelerating}
Ruiqi Zhang, Momin Haider, Ming Yin, Jiahao Qiu, Mengdi Wang, Peter Bartlett, and Andrea Zanette.
\newblock Accelerating best-of-n via speculative rejection.
\newblock In \emph{2nd Workshop on Advancing Neural Network Training: Computational Efficiency, Scalability, and Resource Optimization (WANT@ICML 2024)}, 2024{\natexlab{b}}.
\newblock URL \url{https://openreview.net/forum?id=dRp8tAIPhj}.

\bibitem[Zhang et~al.(2024{\natexlab{c}})Zhang, Wu, Yang, Shu, Xiao, Kong, and Sang]{zhang2024o1codero1replicationcoding}
Yuxiang Zhang, Shangxi Wu, Yuqi Yang, Jiangming Shu, Jinlin Xiao, Chao Kong, and Jitao Sang.
\newblock o1-coder: an o1 replication for coding, 2024{\natexlab{c}}.
\newblock URL \url{https://arxiv.org/abs/2412.00154}.

\bibitem[Zhou et~al.(2021)Zhou, Ma, and Liu]{EDT}
Zhihan Zhou, Liqian Ma, and Han Liu.
\newblock Trade the event: Corporate events detection for news-based event-driven trading.
\newblock In \emph{Findings of the Association for Computational Linguistics: ACL-IJCNLP 2021}, pp.\  2114--2124, Online, August 2021. Association for Computational Linguistics.
\newblock \doi{10.18653/v1/2021.findings-acl.186}.
\newblock URL \url{https://aclanthology.org/2021.findings-acl.186}.

\end{thebibliography}

\newpage
\appendix

\begin{center}
\begin{huge}
\textbf{Appendix}
\end{huge}
\end{center}

\section{The Usage of Large Language Models (LLMs)} \label{sec:llm_usage}

This section details the specific role of Large Language Models (LLMs) in this paper.

We employ LLMs \textit{GPT-5} (developed by OpenAI) and \textit{Doubao} (developed by ByteDance) to enhance the clarity, coherence, and overall quality of our manuscript.
The LLM assistance primarily focus on language polishing (refining structure, terminology consistency, grammar) and formatting adjustments, ensuring that the paper meets high standards of academic writing.
No other LLMs were used for research ideation or image generation.

All reviewed/approved by authors.
All authors bear \textit{full responsibility} for the final paper. All content (including LLM-generated/polished text) was verified: factual claims cross-checked against datasets/literature, and the manuscript screened to avoid unintended plagiarism.

\section{Implementation Details} \label{sec:appendix:implementation_details}

\subsection{Open Source, Open Weights, and Open Data}
The source code is available at GitHub\footnote{\repourl}.
The model weights are available at HuggingFace\footnote{\huggingfaceurl}. The training and evaluation datasets are available at HuggingFace\footnote{\dataseturl}.

\subsection{Model Training}
The models are trained on up to 4*8 H100 GPUs. Rollout $n$ is set to $8$.
The training epochs are set to $3$ for SFT and $1$ for RL.
The details of the data synthesis workflow for building SFT dataset are shown in Appendix~\ref{sec:appendix:sft_data_synthesis}. With the help of the SFT model to determine the prediction difficulty, we further apply GRPO~\citep{deepseek-math} with curriculum learning and reward shaping. The objective is
\begin{equation}
  \footnotesize
  \begin{split}
      & \mathcal{J}_{GRPO}(\theta) = \mathbb{E}{[q \sim P(Q), \{o_i\}_{i=1}^G \sim \pi_{\theta_{old}}(O|q)]} \frac{1}{G}\sum_{i=1}^G\frac{1}{|o_i|} \sum_{t=1}^{|o_i|} \\
      &  \left\{ \min \left[ \frac{\pi_\theta(o_{i,t} | q, o_{i,<t})}{\pi_{\theta_{old}}(o_{i,t} | q, o_{i,<t})} \hat{A}_{i,t}, \text{clip} \left( \frac{\pi_\theta(o_{i,t} | q, o_{i,<t})}{\pi_{\theta_{old}}(o_{i,t} | q, o_{i,<t})}, 1 - \epsilon, 1 + \epsilon \right)  \hat{A}_{i,t} \right] - \beta \mathbb{D}_{KL}\left[\pi_{\theta} || \pi_{ref}\right]\right\} ,
  \end{split}
  \label{eq:GRPO-obj}
\end{equation}
where advantage $\hat{A}_{i, t} = \frac{r_i- {\rm mean}(\mathbf{r})}{{\rm std}(\mathbf{r})}$, $G$ is the group size, $\beta$ is the coefficient of KL penalty, and $q = (x, \{y_i\}_{i=1}^n)$ with prompt $x$ and $n$ generations $y$.


\subsection{Hyperparameters}
We report the detailed hyperparameters for Supervised Fine-Tuning (SFT) in Table~\ref{table:hyperparameters:sft} and for GRPO in Table~\ref{table:hyperparameters:grpo}. We use up to 4*8 H100 80GB GPUs for experiments.

\subsection{Training Frameworks}
We use Xtuner~\citep{2023xtuner} to SFT with DeepSpeed~\citep{deepspeed} to accelerate training and ZeRO-3 to reduce memory usage, and verl~\citep{verl} to implement GRPO.

\begin{table*}[htbp]
  \centering
  \caption{The training hyperparameters for Supervised Fine-Tuning (SFT). 32B, 14B, and 7B denote models based on DeepSeek-R1-Distill-Qwen with 32B, 14B, and 7B parameters respectively.}
  \begin{tabularx}{\linewidth}{l XXX} 
    \toprule
    \cellcolor{Background1}\textbf{Hyperparameter Category} & \cellcolor{Background1}\textbf{32B} & \cellcolor{Background1}\textbf{14B} & \cellcolor{Background1}\textbf{7B} \\
    \hline
    \multicolumn{4}{c}{\cellcolor{Background1}\textcolor{Font1}{\textbf{1. Data Configuration}}} \\
    Train Micro-Batch Size per GPU                  & 1 & 1 & 1 \\
    Gradient Accumulation Steps                     & 128 & 128 & 128 \\
    Total Effective Batch Size                      & 1024  & 1024 & 1024 \\
                          & ($1\times128\times8$) & ($1\times128\times8$) & ($1\times128\times8$) \\
    Pack Sequences to Max Length                    & False & False & False \\
    Data Shuffling Before Packing                   & True & True & True \\
    \hline
    \multicolumn{4}{c}{\cellcolor{Background1}\textcolor{Font1}{\textbf{2. Model \& LoRA Configuration}}} \\
    LLM Torch Dtype                                 & torch.float16 & torch.float16 & torch.float16 \\
    LoRA Rank ($r$)                                   & 32 & 32 & 32 \\
    LoRA Alpha ($\alpha$)                                  & 64 & 64 & 64 \\
    LoRA Dropout                                    & 0.1 & 0.1 & 0.1 \\
    LoRA Bias Type                                  & none & none & none \\
    LoRA Task Type                                  & CAUSAL\_LM & CAUSAL\_LM & CAUSAL\_LM \\
    Variable-Length Attention                       & False & False & False \\
    \hline
    \multicolumn{4}{c}{\cellcolor{Background1}\textcolor{Font1}{\textbf{3. Optimizer \& LR Scheduler}}} \\
    Optimizer Type                                  & \multicolumn{3}{c}{torch.optim.AdamW} \\
    Learning Rate (LR)                              & $2\times 10^{-4}$ & $2\times 10^{-4}$ & $2\times 10^{-4}$ \\
    AdamW Betas ($\beta_1$, $\beta_2$)              & (0.9, 0.999) & (0.9, 0.999) & (0.9, 0.999) \\
    Weight Decay                                    & 0 & 0 & 0 \\
    Gradient Clipping Max Norm                      & 1 & 1 & 1 \\
    LR Scheduler Type                               & \multicolumn{3}{c}{Linear Warmup + Cosine Annealing} \\
    Warmup Ratio (Warmup Epochs / Total Epochs)     & 0.03 (0.09/3) & 0.03 (0.09/3) & 0.03 (0.09/3) \\
    Warmup Start Factor                             & $1\times 10^{-5}$ & $1\times 10^{-5}$ & $1\times 10^{-5}$ \\
    Cosine Annealing Final LR ($\eta_{min}$)                & 0.0 & 0.0 & 0.0 \\
    \hline
    \multicolumn{4}{c}{\cellcolor{Background1}\textcolor{Font1}{\textbf{4. Training Strategy \& Distributed Config}}} \\
    Training Strategy Type                          & \multicolumn{3}{c}{DeepSpeedStrategy} \\
    DeepSpeed Zero Optimization Stage               & 3 & 3 & 3 \\
    BF16 Precision Enabled                          & True & True & True \\
    FP16 Precision Enabled                          & False & False & False \\
    Sequence Parallel Size                          & 8 & 8 & 8 \\
    Sampler Shuffling                               & True & True & True \\
    \hline
    \multicolumn{4}{c}{\cellcolor{Background1}\textcolor{Font1}{\textbf{6. Environment \& Misc Config}}} \\
    Launcher Type                                   & \multicolumn{3}{c}{pytorch} \\
    Distributed Backend                             & nccl & nccl & nccl \\
    Multiprocessing Start Method                    & fork & fork & fork \\
    Deterministic Training                          & False & False & False \\
    \bottomrule
  \end{tabularx}
  \label{table:hyperparameters:sft}
\end{table*}

\begin{table*}[htbp]
  \centering
  \caption{The training hyperparameters for GRPO. 32B SFT, 14B SFT, and 7B SFT denote models based on DeepSeek R1 with 32B, 14B, and 7B parameters after SFT stage respectively.}
  \begin{tabularx}{\linewidth}{l XXX} 
    \toprule
    \cellcolor{Background1}\textbf{Base Model} & \cellcolor{Background1}\textbf{32B SFT} & \cellcolor{Background1}\textbf{14B SFT} & \cellcolor{Background1}\textbf{7B SFT} \\
    \hline
    \multicolumn{4}{c}{\cellcolor{Background1}\textcolor{Font1}{\textbf{1. Data Configuration}}} \\
    Training Batch Size                            & 256 & 256 & 256 \\
    Validation Batch Size                          & 256 & 256 & 256 \\
    Max Prompt Length                              & 32768 & 32768 & 32768 \\
    Max Response Length                            & 4096 & 4096 & 4096 \\
    Data Shuffling                                 & True & True & True \\
    \hline
    \multicolumn{4}{c}{\cellcolor{Background1}\textcolor{Font1}{\textbf{2. Algorithm Configuration}}} \\
    Advantage Estimator                            & GRPO & GRPO & GRPO \\
    Gamma ($\gamma$, Discount Factor)              & 1.0 & 1.0 & 1.0 \\
    Lambda ($\lambda$, Advantage Smoothing)        & 1.0 & 1.0 & 1.0 \\
    KL Coefficient                                 & 0.001 & 0.001 & 0.001 \\
    Target KL Divergence                           & 0.1 & 0.1 & 0.1 \\
    Normalize Advantage by Std                     & True & True & True \\
    \hline
    \multicolumn{4}{c}{\cellcolor{Background1}\textcolor{Font1}{\textbf{3. Actor \& Ref Configuration}}} \\
    Learning Rate                                  & $3 \times 10^{-7}$ & $3 \times 10^{-7}$ & $3 \times 10^{-7}$ \\
    Weight Decay                                   & 0.01 & 0.01 & 0.01 \\
    Clip Ratio                                     & 0.2 & 0.2 & 0.2 \\
    Entropy Coefficient                            & 0.0 & 0.0 & 0.0 \\
    PPO Epochs                                     & 1 & 1 & 1 \\
    Log Prob Micro Batch Size                      & 8 & 4 & 4 \\
    Tensor Parallel Size                           & 4 & 4 & 4 \\
    \hline
    \multicolumn{4}{c}{\cellcolor{Background1}\textcolor{Font1}{\textbf{4. Rollout Configuration}}} \\
    Rollout Count (n)                              & 8 & 8 & 8 \\
    Rollout Mode                                   & sync & sync & sync \\
    Engine Name                                    & vllm & vllm & vllm \\
    Data Type (dtype)                              & bfloat16 & bfloat16 & bfloat16 \\
    Temperature                                    & 0.6 & 0.6 & 0.6 \\
    Max Num Batched Tokens                         & 36864 & 36864 & 36864 \\
    Tensor Parallel Size                           & 4 & 4 & 4 \\
    Enable Chunked Prefill                         & True & True & True \\
    \hline
    \multicolumn{4}{c}{\cellcolor{Background1}\textcolor{Font1}{\textbf{5. Trainer Configuration}}} \\
    Number of Nodes                                & 4 & 2 & 1 \\
    GPUs per Node                                  & 8 & 8 & 8 \\
    Total Epochs                                   & 1 & 1 & 1 \\
    Checkpoint Save Frequency (steps)              & 10 & 10 & 10 \\
    Validation Frequency (steps)                   & 10 & 10 & 10 \\
    Validate Before Training                       & True & True & True \\
    \bottomrule
  \end{tabularx}
  \label{table:hyperparameters:grpo}
\end{table*}

\section{Dataset Details} \label{sec:appendix:dataset_details}

\subsection{Fin-2024 and Fin-2025}
We consider studying the stock movement prediction task based on data from the Chinese A-share market. Naturally, the collected data are in Chinese, and consequently, the associated prompts and synthetic data are also in Chinese. This consistency within a single language allows LLMs to achieve better understanding and more coherent reasoning over the data. Thus, the language model can fully leverage its pre-trained knowledge in Chinese to analyze the stock market. Therefore, in this paper, we choose Qwen~\citep{Qwen} and DeepSeek~\citep{deepseek-r1} as the backbone models, both of which are strong Chinese LLMs. We apologize for any inconvenience the language gap may cause to readers who are not native Chinese speakers, and we hope that this gap will not hinder the understanding of our work.

The reason why we decide to construct a new dataset to study the stock movement prediction task is twofold. First, existing datasets (StockNet~\citep{StockNet}, CMIN-US~\citep{CMIN}, CMIN-CN~\citep{CMIN}, EDT~\citep{EDT}) are outdated and do not reflect the current market conditions. Financial markets are dynamic and constantly evolving, with new trends, regulations, and events shaping the landscape. Using outdated datasets may lead to models that are not well-suited for current market scenarios. Second, existing datasets lack diversity in data sources. Relying solely on price and news data may not capture the full complexity of stock movements. Incorporating additional data sources such as analyst reports, macroeconomic indicators, and quantitative reports can provide a more comprehensive view of the market and improve prediction accuracy. Rich enough data sources are crucial for reliable forecasting in the high–signal-noise ratio of financial markets.

Then, we build a new dataset \textbf{Fin-2024} covering January to December 2024, with a test split set \textbf{Fin-2024[December]} and an additional long-horizon evaluation set \textbf{Fin-2025[June]}. We collect data from multiple sources, including stock prices, financial news, analyst reports, macroeconomic indicators, and quantitative reports. We process the raw data into a structured format suitable for LLMs, including entity recognition, sentiment analysis, event extraction, and traditional time-series analysis (for generating quantitative reports). The dataset contains 209,063 data points across 5,123 A-share stocks from various sectors. Each data point includes a timestamp, stock identifier, historical prices (open, close, high, low, volume), relevant news articles, analyst reports, macroeconomic indicators, quantitative reports, and the corresponding stock movement label (up/down/hold), which is based on the change\_pct between the open price of the next trading day and the close price of the current trading day.

\begin{figure*}[htbp]
	\centering
	\includegraphics[width=1.\textwidth]{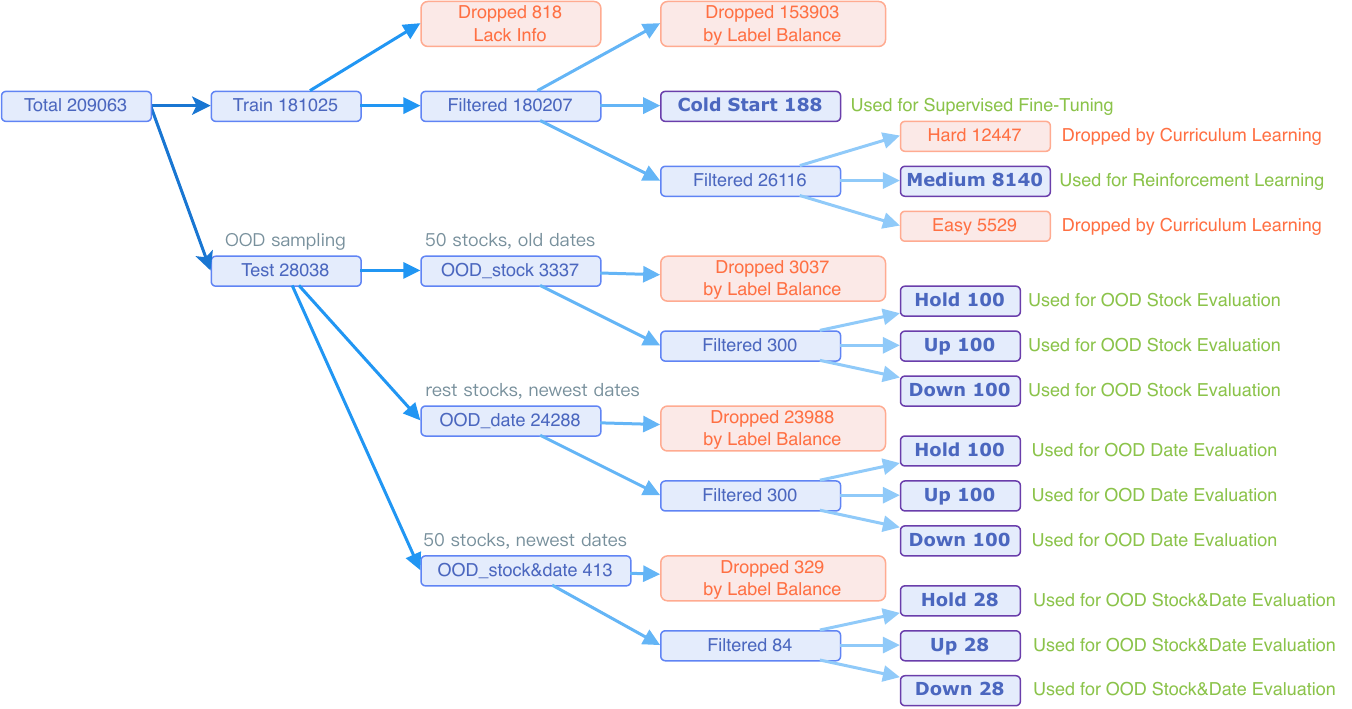}
	\caption{Data processing pipeline showing total 209063 points split into train/test, filtered via multiple steps (lack info, label balance, curriculum learning), and categorized for SFT, RL, OOD evaluations.}
	\label{fig:dataset_pipeline}
\end{figure*}

\begin{figure*}[htbp]
	\centering
	\includegraphics[width=0.9\textwidth]{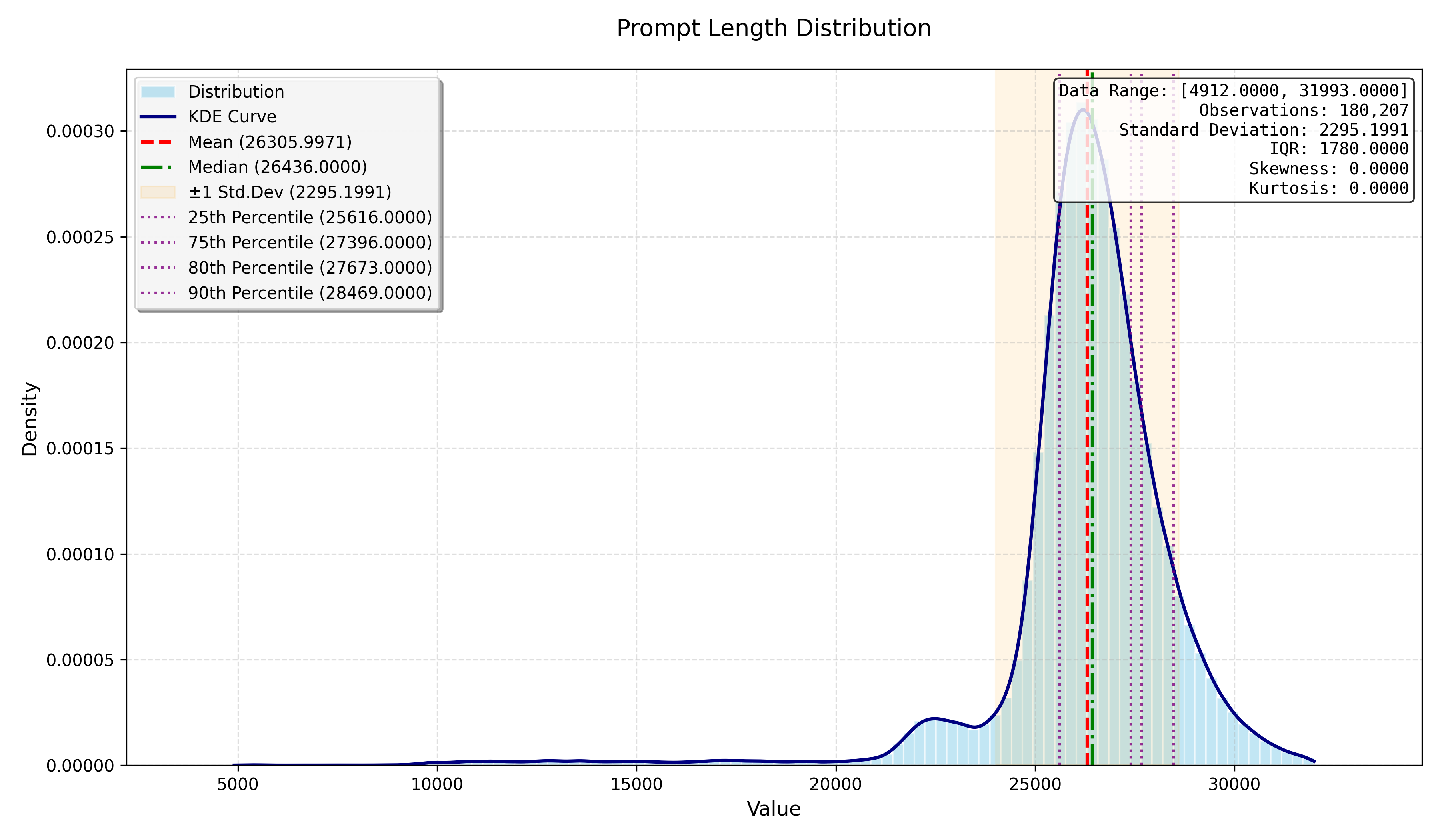}
	\caption{Prompt length distribution across the dataset, illustrating the varying lengths of input prompts used for training and evaluation.}
	\label{fig:prompt_length_distribution}
\end{figure*}

In order to ensure data quality, we apply several filtering steps. We remove data points with missing or incomplete information, filter out stocks with low trading volume or insufficient historical data, and balance the label distribution in the dataset to avoid bias towards any particular class. The data processing pipeline is shown in Figure~\ref{fig:dataset_pipeline}, and the prompt length distribution is shown in Figure~\ref{fig:prompt_length_distribution}.

The final dataset is split into training (90\%) and out-of-distribution (OOD) (10\%) sets. The training set is used for model fine-tuning and reinforcement learning, and the OOD set for final evaluation. The long-horizon evaluation set \textbf{Fin-2025[June]} contains data from June 2025 to assess model performance in a future market scenario.
Please refer to Figure~\ref{fig:dataset_pipeline} for detailed numbers and splits.

\begin{figure*}[htbp]
    \centering
    \includegraphics[width=\textwidth]{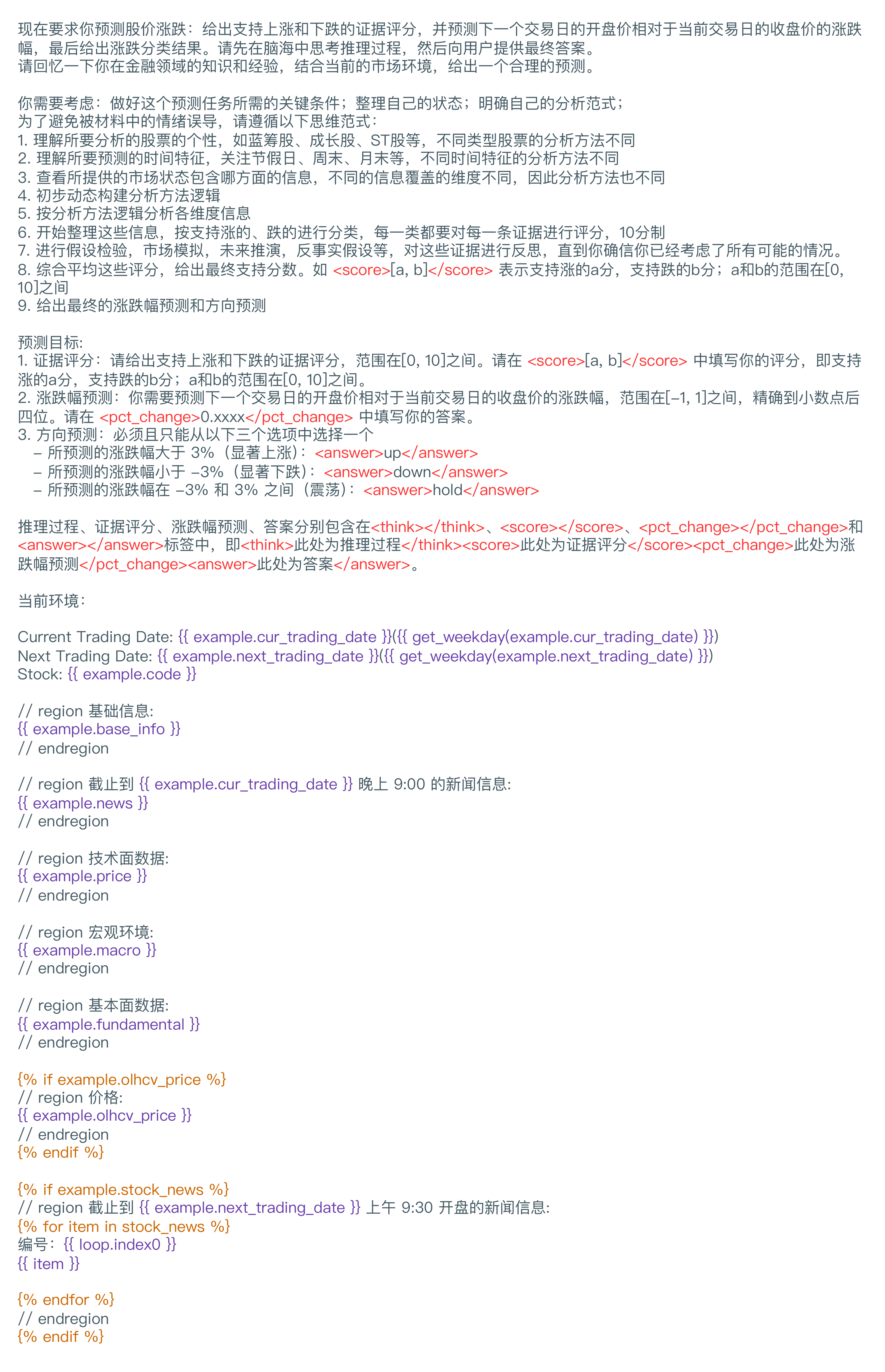}
    \caption{The prompt template for stock movement prediction.}
    \label{fig:appendix:prompt_template}
\end{figure*}

\begin{figure*}[htbp]
    \centering
    \includegraphics[width=\textwidth]{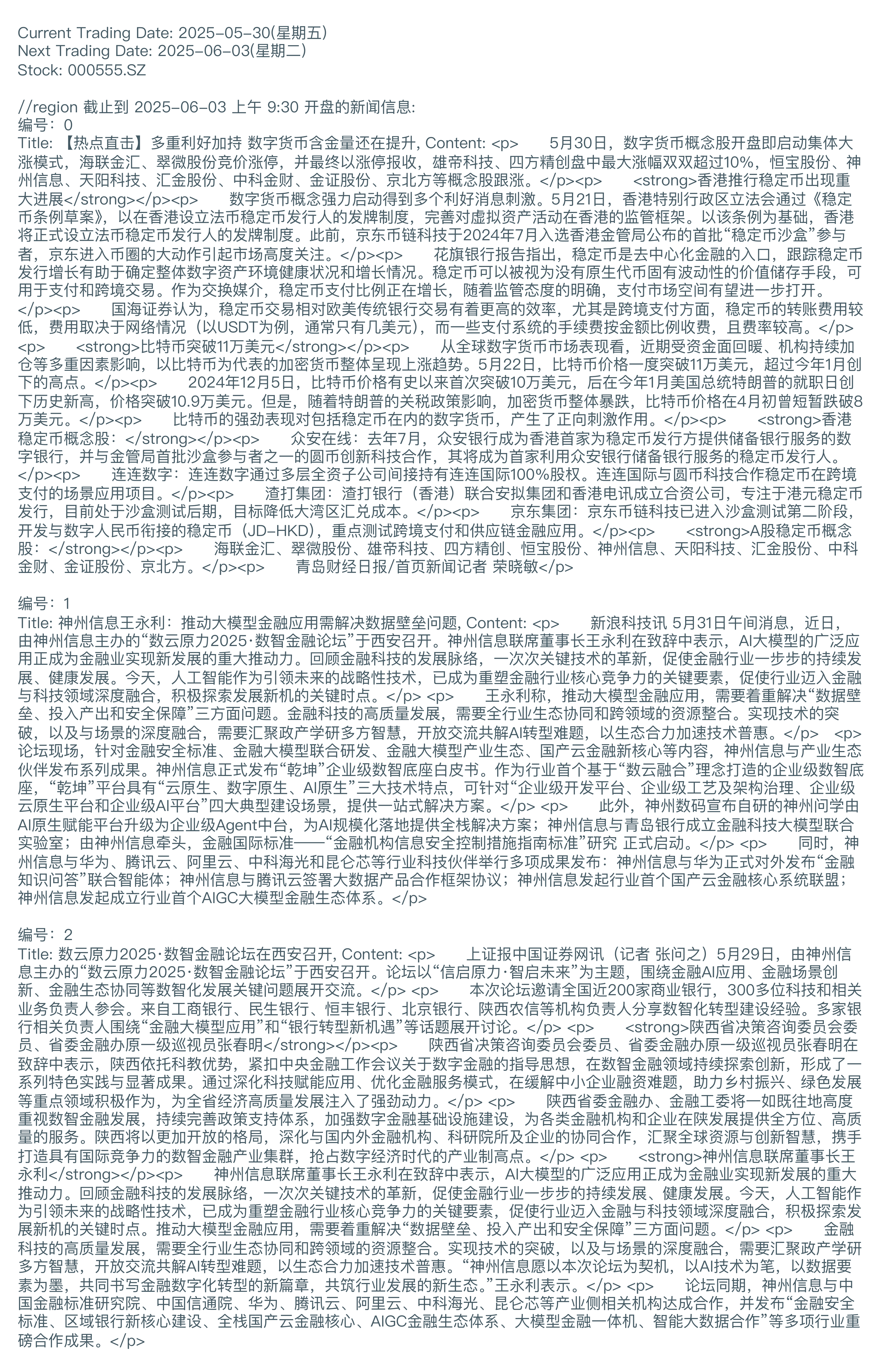}
    \caption{Example. Part 1.1: Stock News.}
    \label{fig:appendix:example_prompt_part_news}
\end{figure*}

\begin{figure*}[htbp]
    \centering
    \includegraphics[width=\textwidth]{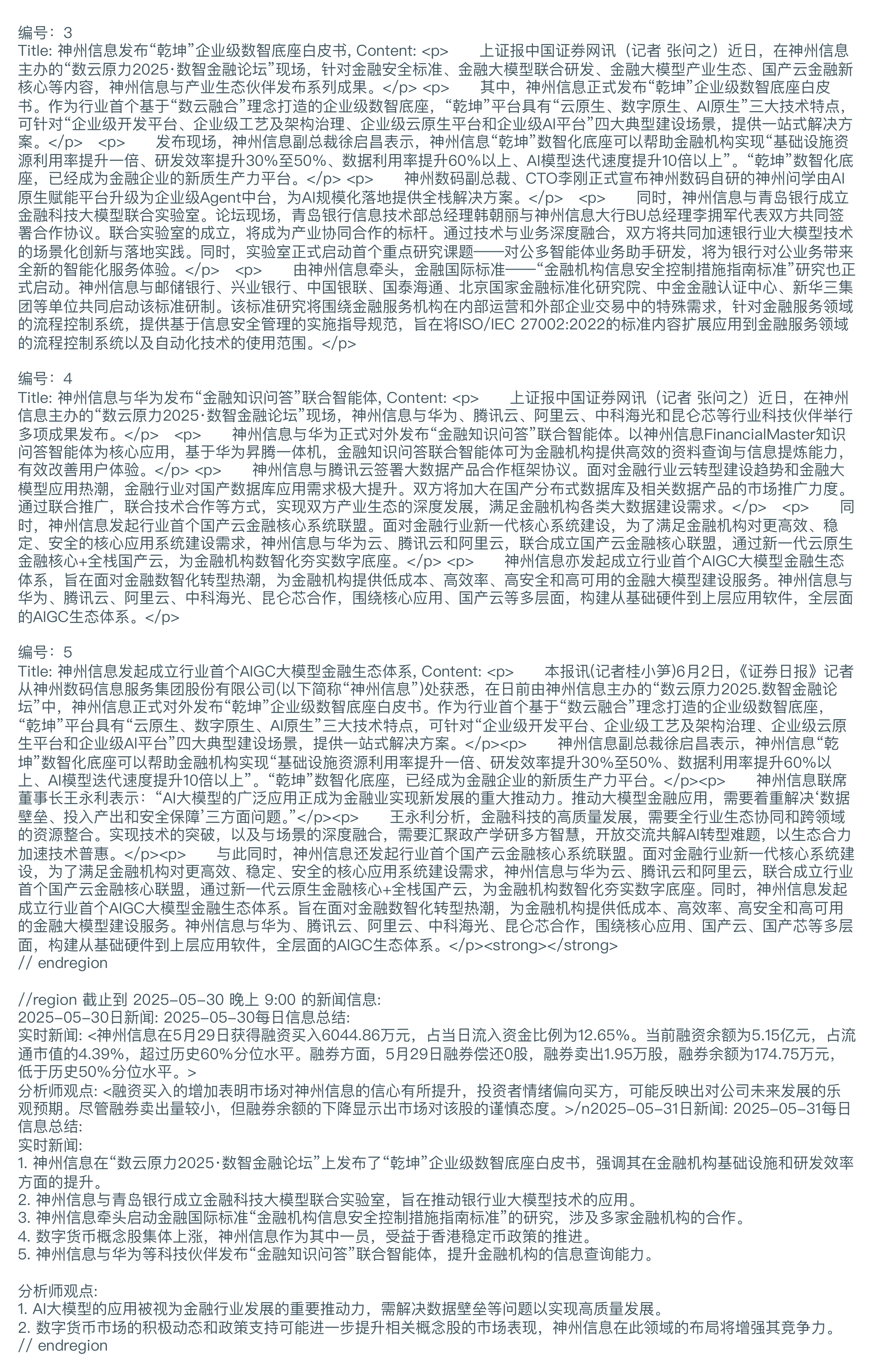}
    \caption{Example. Part 1.2: Stock News.}
    \label{fig:appendix:example_prompt_part_news2}
\end{figure*}

\begin{figure*}[htbp]
    \centering
    \includegraphics[width=0.75\textwidth]{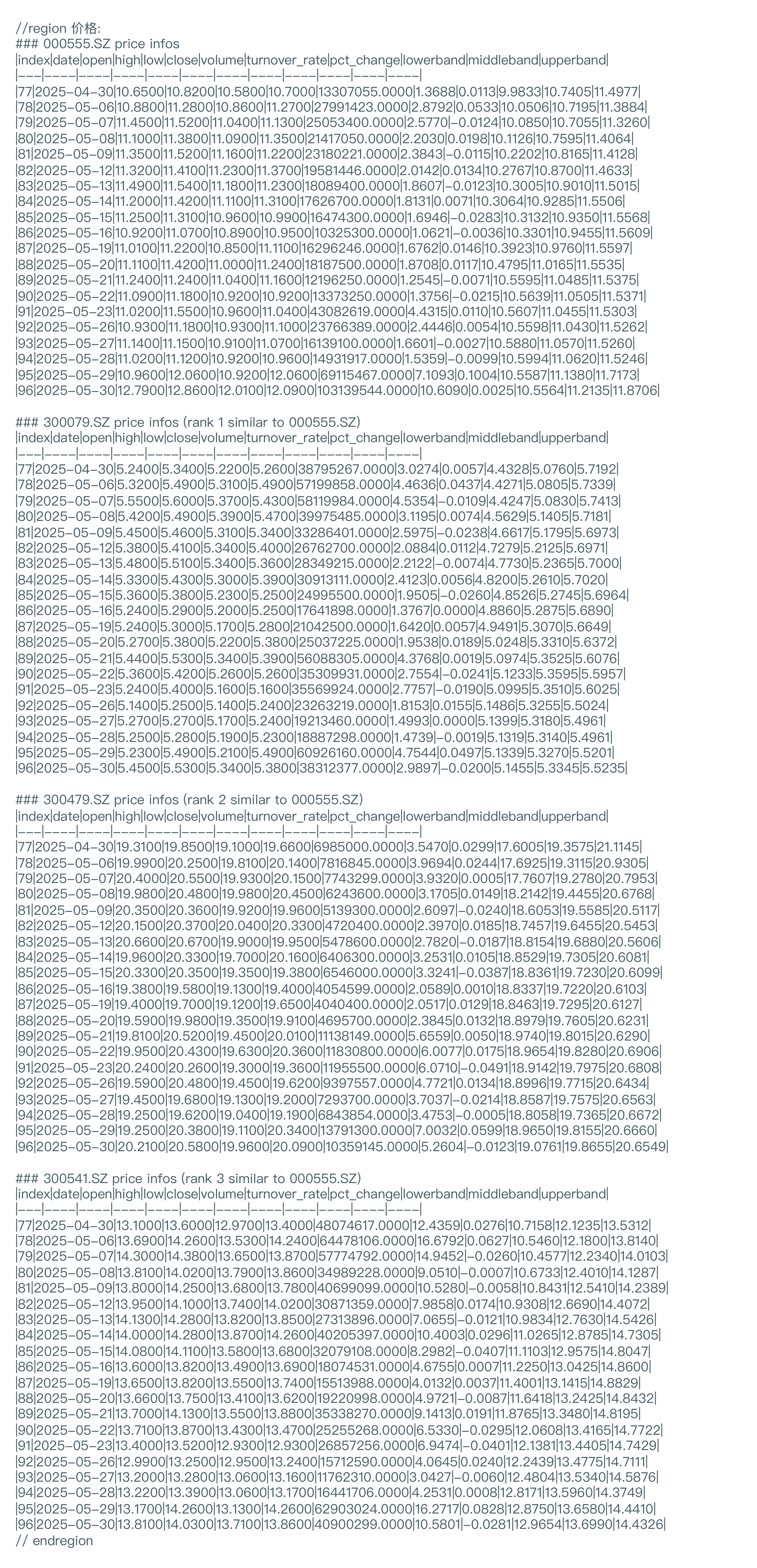}
    \caption{Example. Part 2: Stock Price Info of Current Stock and Top-3 Similar Stocks.}
    \label{fig:appendix:example_prompt_part_price}
\end{figure*}

\begin{figure*}[htbp]
    \centering
    \includegraphics[width=\textwidth]{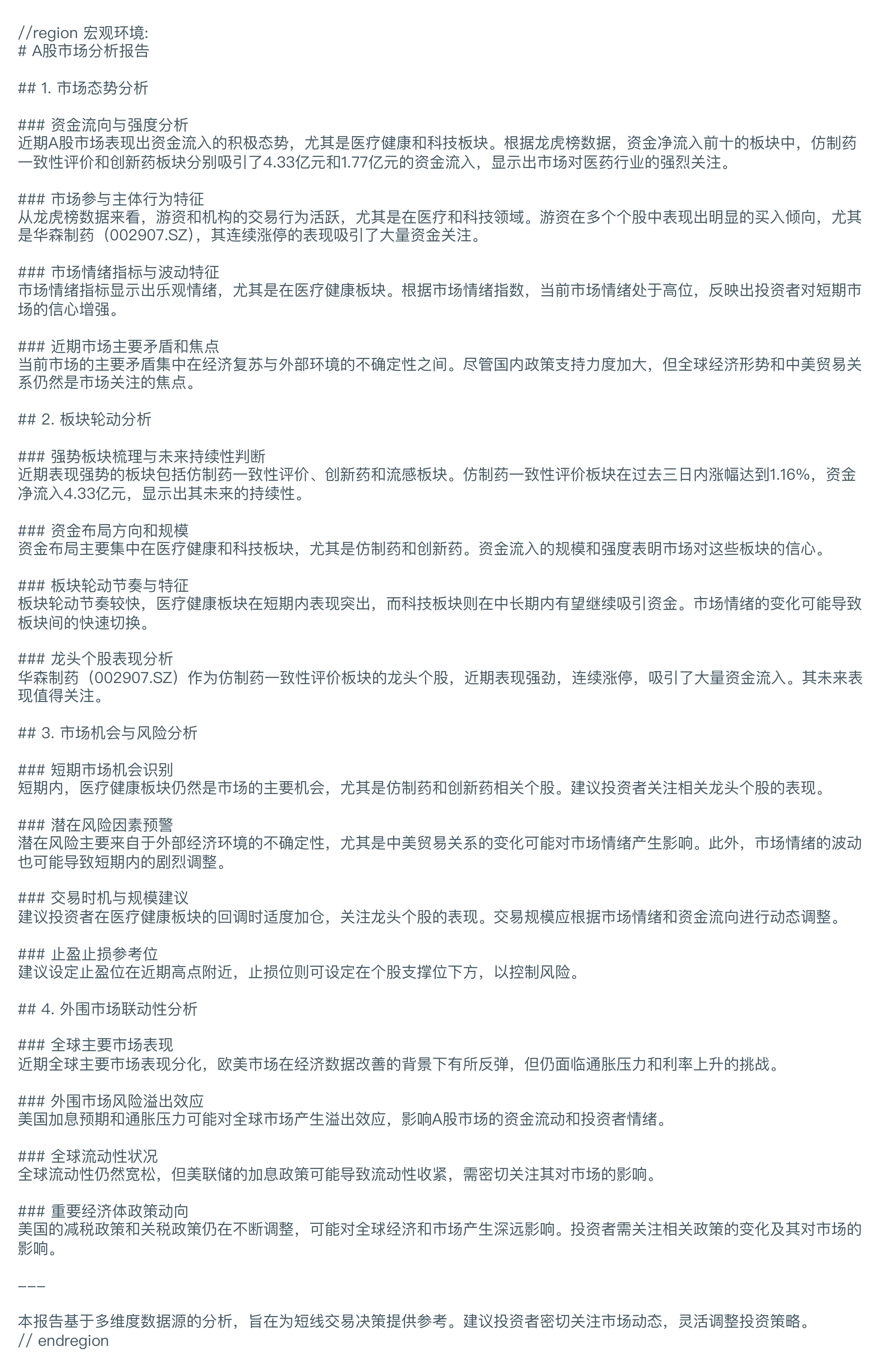}
    \caption{Example. Part 3: Macroeconomic Indicators Report.}
    \label{fig:appendix:example_prompt_part_marco}
\end{figure*}

\begin{figure*}[htbp]
    \centering
    \includegraphics[width=\textwidth]{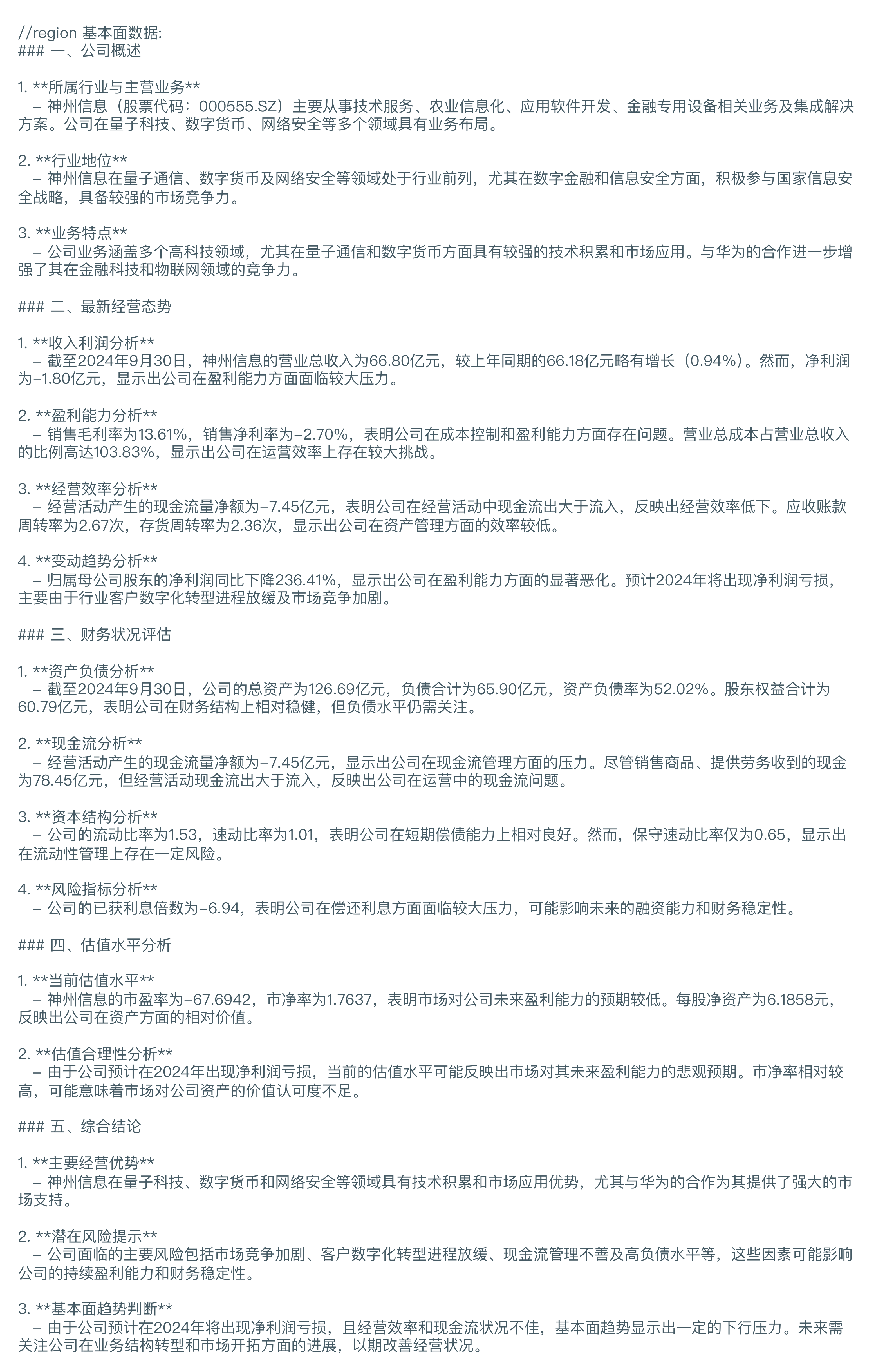}
    \caption{Example. Part 4.1: Stock Fundamentals Report.}
    \label{fig:appendix:example_prompt_part_fundamentals}
\end{figure*}

\begin{figure*}[htbp]
    \centering
    \includegraphics[width=0.99\textwidth]{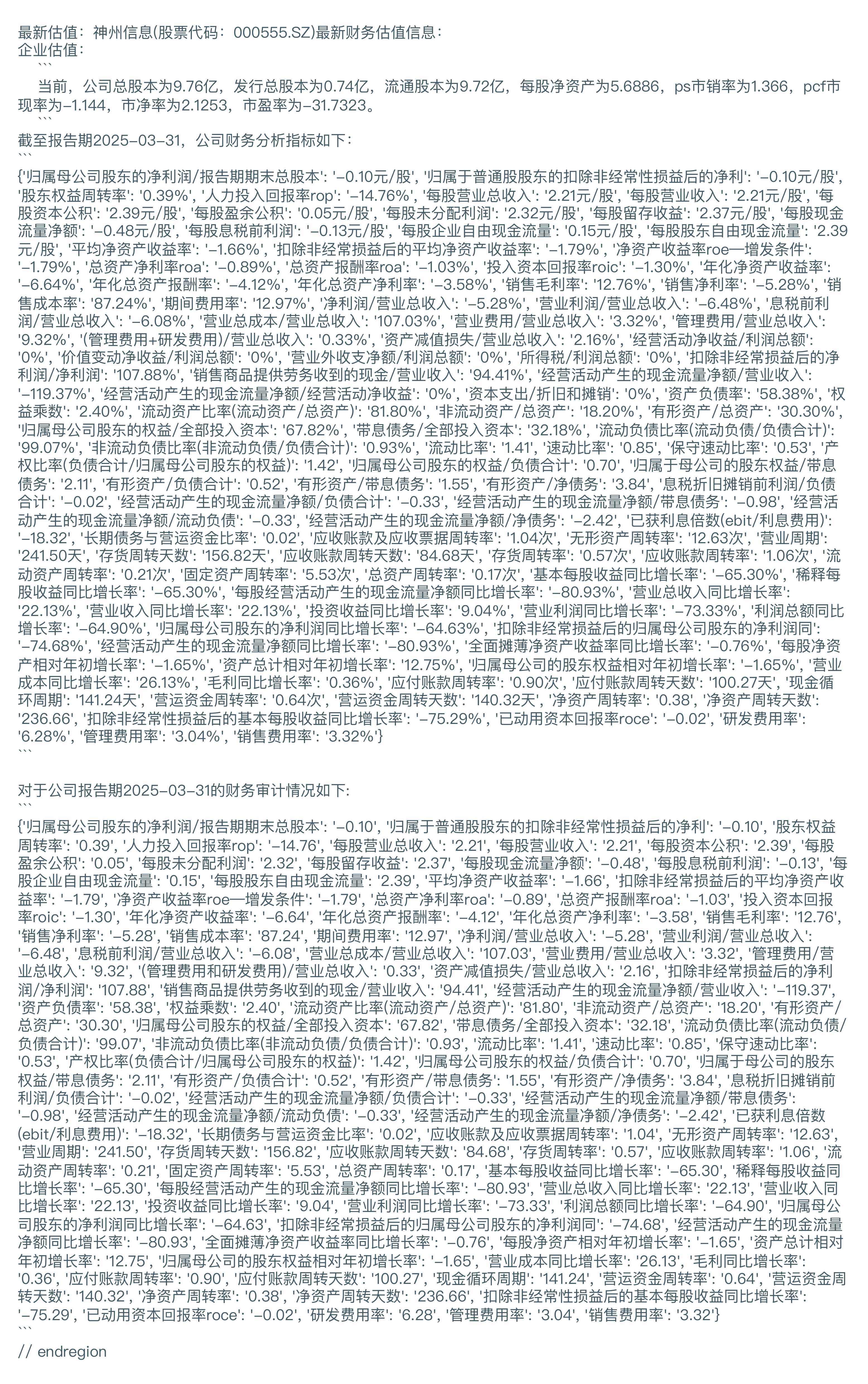}
    \caption{Example. Part 4.2: Stock Fundamentals Report.}
    \label{fig:appendix:example_prompt_part_fundamentals2}
\end{figure*}

\begin{figure*}[htbp]
    \centering
    \includegraphics[width=\textwidth]{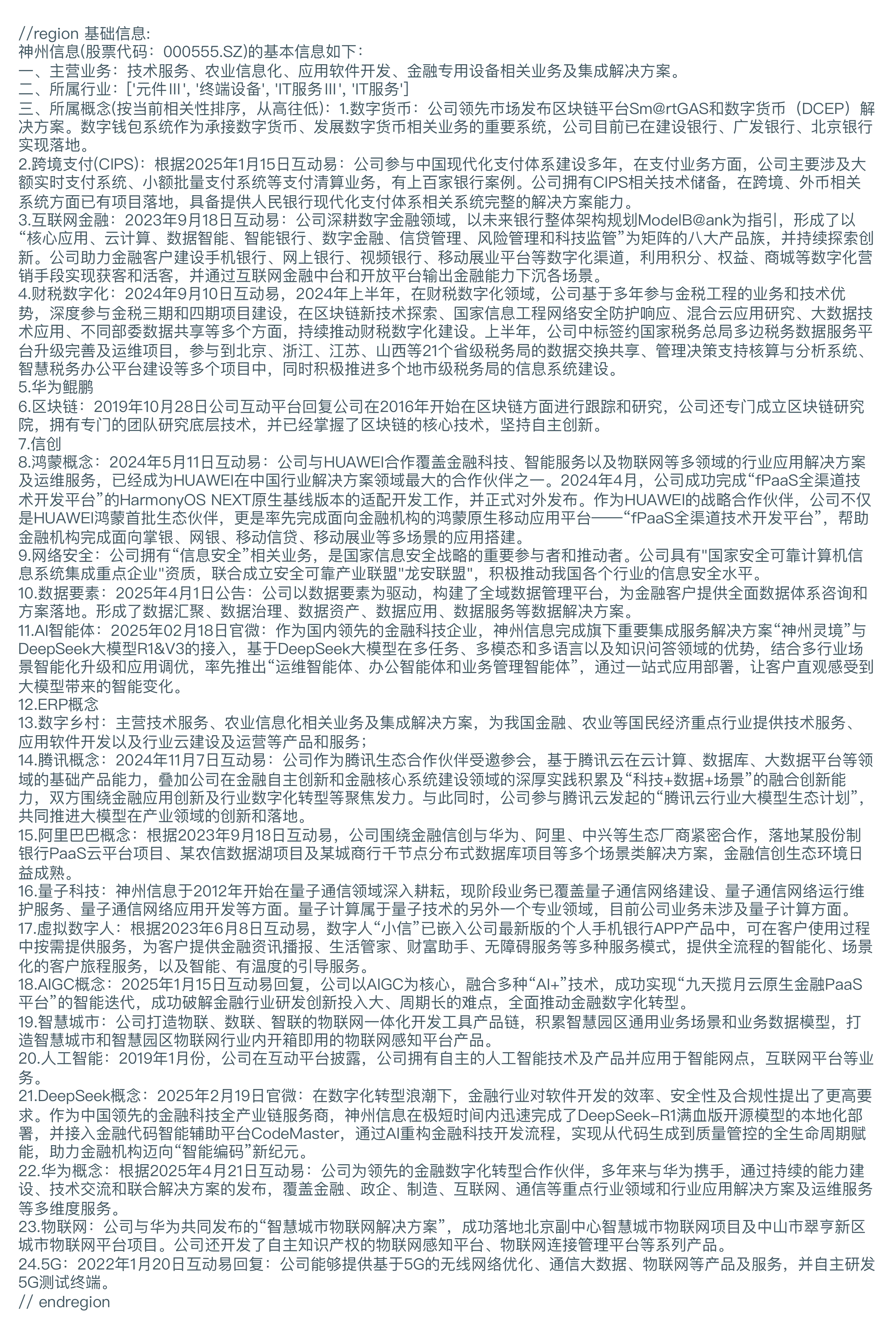}
    \caption{Example. Part 5: Stock Basic Information.}
    \label{fig:appendix:example_prompt_part_baseinfo}
\end{figure*}

\begin{figure*}[htbp]
    \centering
    \includegraphics[width=0.85\textwidth]{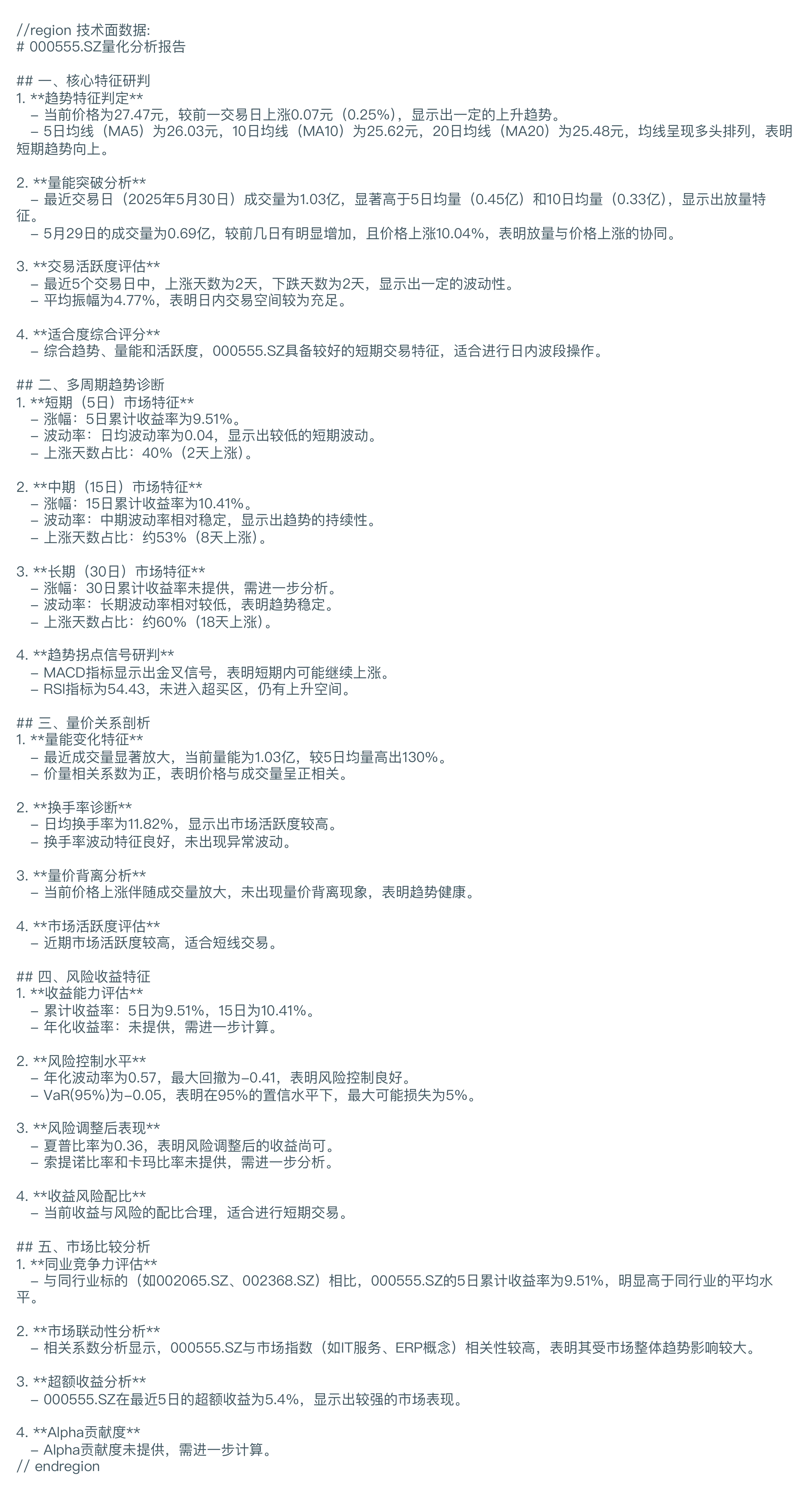}
    \caption{Example. Part 6: Stock Quantitative Reports.}
    \label{fig:appendix:example_prompt_part_quantitative_reports}
\end{figure*}

\begin{figure*}[htbp]
    \centering
    \includegraphics[width=\textwidth]{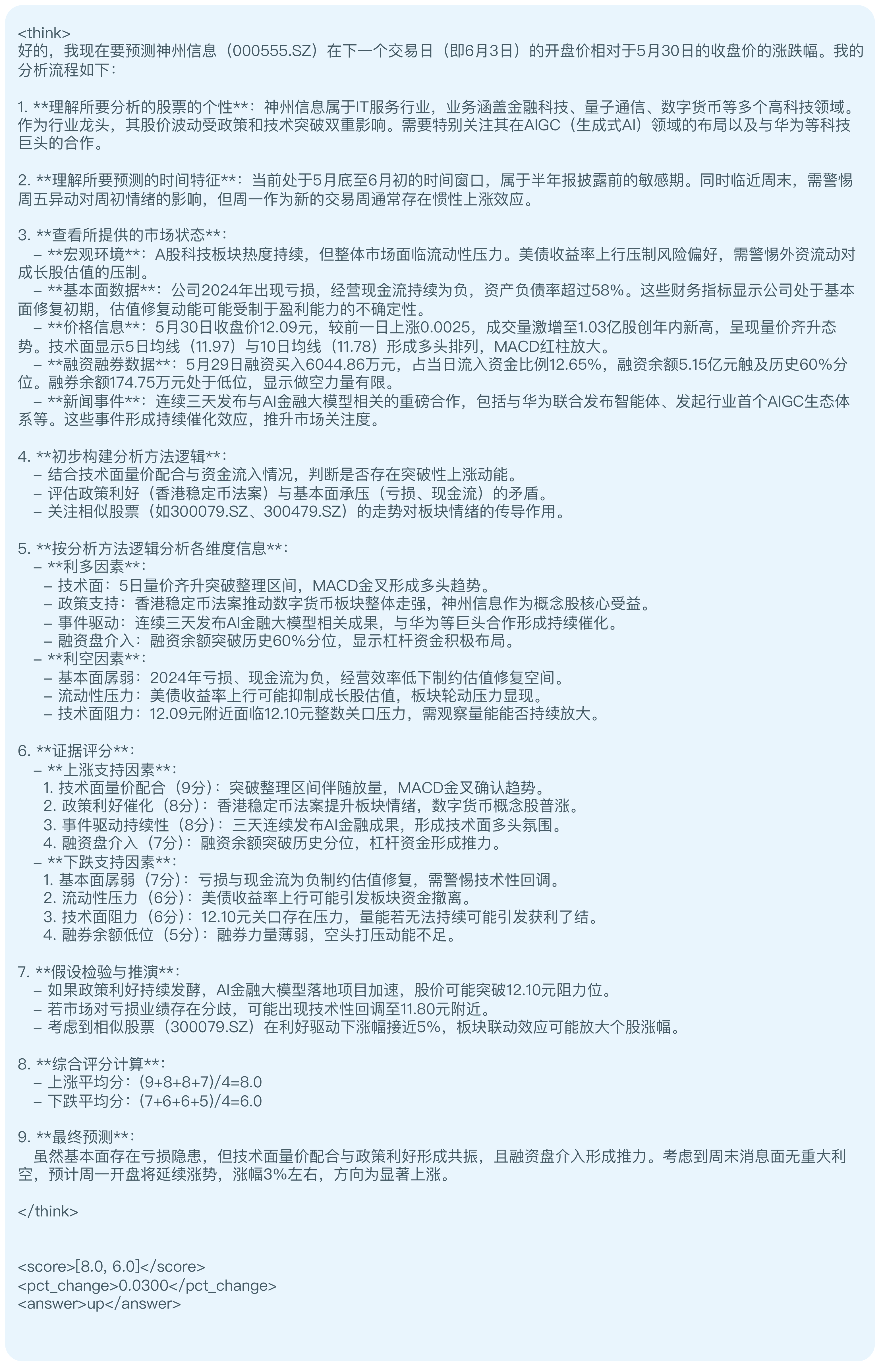}
    \caption{Example. Part 7: Model Response.}
    \label{fig:appendix:example_response}
\end{figure*}

\begin{figure*}[htbp]
    \centering
    \includegraphics[width=\textwidth]{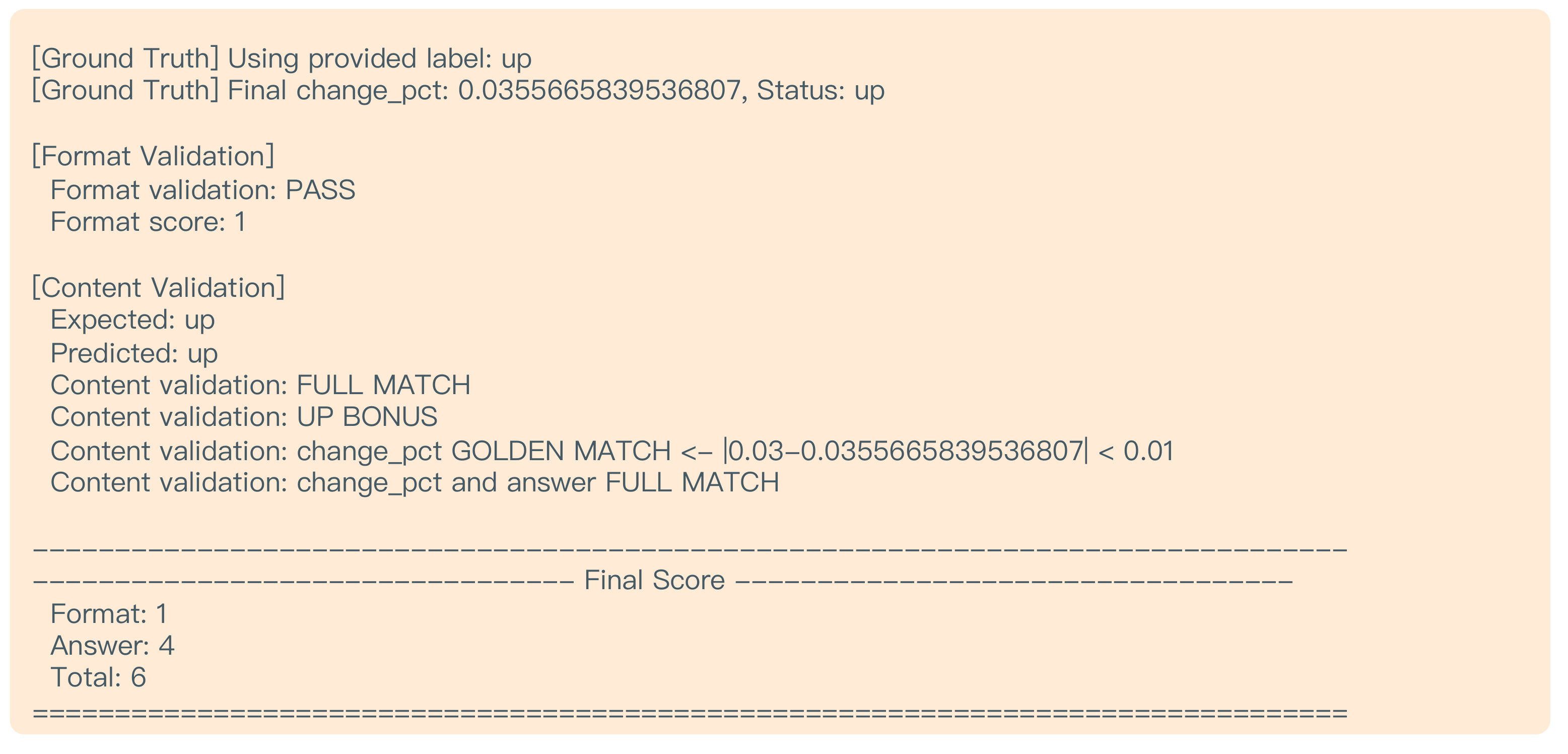}
    \caption{Example. Part 8: Model Response Grading.}
    \label{fig:appendix:example_response_grading}
\end{figure*}

We present an example of the prompt template in Figure~\ref{fig:appendix:prompt_template}, which consists of multiple parts: stock news (Figure~\ref{fig:appendix:example_prompt_part_news} and Figure~\ref{fig:appendix:example_prompt_part_news2}), stock price information of the current stock and top-3 similar stocks (Figure~\ref{fig:appendix:example_prompt_part_price}), macroeconomic indicators report (Figure~\ref{fig:appendix:example_prompt_part_marco}), stock fundamentals report (Figure~\ref{fig:appendix:example_prompt_part_fundamentals} and Figure~\ref{fig:appendix:example_prompt_part_fundamentals2}), stock basic information (Figure~\ref{fig:appendix:example_prompt_part_baseinfo}), stock quantitative reports (Figure~\ref{fig:appendix:example_prompt_part_quantitative_reports}), model response (Figure~\ref{fig:appendix:example_response}), and model response grading (Figure~\ref{fig:appendix:example_response_grading}).

\subsection{Fin-2024-SFT}\label{sec:appendix:sft_data_synthesis}

The dataset \textbf{Fin-2024-SFT} contains 188 cold-start items and 10K general reasoning samples. Firstly, we selected 10K items from \url{https://huggingface.co/datasets/GeneralReasoning/GeneralThought-323K}, which aims to avoid catastrophic forgetting~\citep{catastrophic_forgetting} and even strengthen reasoning ability when fine-tuning the model. These data points are related to math, code, common sense, chatting, role play, writing, etc. However, they are not allowed to be related to finance to avoid bias, so that we can determine the model's performance in a more controlled setting.

\begin{figure*}[htbp]
    \centering
    \includegraphics[width=\textwidth]{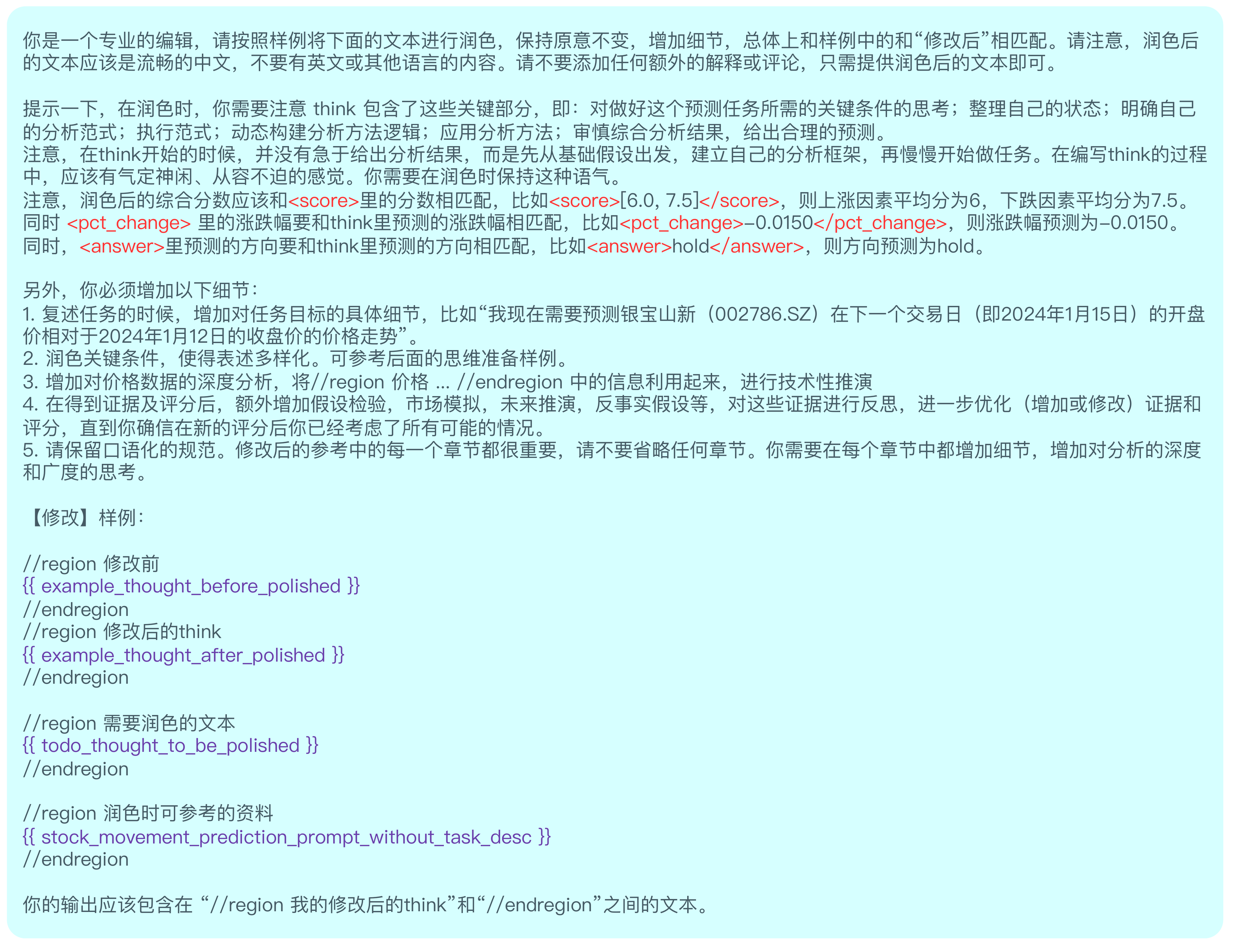}
    \caption{The polish prompt allows the backbone model to generate refined responses that follow the proposed thinking schema in Section~\ref{sec:method:1}.}
    \label{fig:appendix:polish_prompt}
\end{figure*}

Then, the samples for cold-starting are constructed using a workflow with DeepSeek-R1 (671B) as the backbone model and the polish prompt (Figure~\ref{fig:appendix:polish_prompt}) to generate high-quality, diverse training samples that follow the proposed thinking schema presented in Section~\ref{sec:method:1}. These synthesized samples are further filtered by a reward function, which checks format, validates prediction (\texttt{<score>}, \texttt{<change\_pct>}, \texttt{<answer>}) against the ground truth, and ensures adherence to the desired output style. At this stage, 300 samples passed the validation. From these, we cherry-picked 188 samples in total, which required approximately six hours of expert annotation (around 100 USD) for final curation.

Notably, the financial data we collected spans the entire year of 2024 and June 2025. Among these, the data from January to November 2024 were used for training. From this subset of training, we extracted 300 instances (100 labeled as \emph{up}, 100 as \emph{hold}, and 100 as \emph{down}) to construct the cold-start dataset for synthesis.

\textbf{RETuning} guides the model through a multistep process:

\begin{enumerate}
  \item \textbf{\underline{Understand the Task.}} The model first internalizes the task—predicting the direction of price movement from the previous close to the next open, with a $\pm3\%$ threshold—improving alignment and reducing hallucination.
  \item \textbf{\underline{Establish Analytical Principles.}} It constructs a dynamic analytical framework (e.g., fundamentals, news trends, macro signals), independent of analyst commentary.
  \item \textbf{\underline{Extract Evidence.}} The model collects multiple context-based signals and assesses their support for each directional hypothesis.
  \item \textbf{\underline{Group and Score.}} Extracted evidence is grouped and scored by directional leaning, forming a soft-evidence pool.
  \item \textbf{\underline{Reflect and Reconcile.}} Averaging directional scores, the model enters a reflection phase to resolve conflicting evidence.
  \item \textbf{\underline{Produce Structured Output.}} The model generates a final decision in a consistent, interpretable format, specifying both direction and percentage change.
\end{enumerate}

In detail, we employed the following workflow:

\begin{itemize}
    \item \textbf{Step 1: Multi-Source Data Aggregation.} We systematically collected a diverse range of financial data covering the period from January 2024 to June 2025. This included quantitative data, such as historical stock prices (open, high, low, close) and trading volumes, alongside qualitative data from relevant financial news articles. For training purposes, we focused on the January--November 2024 subset, from which 300 balanced instances (up, hold, down) were selected for cold-start data synthesis. To ensure the dataset's breadth and representativeness, we selected a varied portfolio of stocks from multiple market sectors.

    \item \textbf{Step 2: Structured Prompt Formulation.} We engineered a series of structured prompts designed to elicit detailed, context-aware analytical reasoning from the language model. These prompts incorporated explicit instructions, illustrative examples (few-shot learning), and comprehensive contextual information (e.g., market conditions, company background) to precisely guide the model's generation process according to our proposed thinking schema.

    \item \textbf{Step 3: Controlled Response Generation.} Leveraging the 671B-parameter DeepSeek-R1 model, we executed inference on the engineered prompts. To foster response diversity and prevent deterministic outputs, we employed stochastic sampling techniques, specifically temperature sampling and top-k sampling. This allowed the model to explore a wider range of analytical paths and linguistic styles.

    \item \textbf{Step 4: Hybrid Quality Assurance.} The generated responses underwent a rigorous two-stage filtering process. Initially, automated metrics were used to assess fundamental quality attributes such as textual coherence, logical consistency, and relevance to the prompt. Subsequently, all machine-vetted samples were subjected to manual review by a financial expert to discard any outputs that were factually incorrect, nonsensical, or failed to meet the required analytical depth.

    \item \textbf{Step 5: Semantic Data Augmentation.} To enrich the dataset and enhance model robustness, we applied several data augmentation techniques to the high-quality samples. Methods such as paraphrasing, synonym replacement, and back-translation were utilized to create semantically equivalent but syntactically diverse training instances, thereby reducing the risk of overfitting to specific phrasings.

    \item \textbf{Step 6: Final Dataset Curation.} The fully processed and augmented data was compiled into the final training set. During this stage, we ensured a balanced class distribution among the target labels (up, down, hold) and a representative allocation of stocks across different sectors and market conditions to form a well-rounded and unbiased dataset for fine-tuning.
\end{itemize}

\subsection{BizFinBench}

BizFinBench~\citep{lu2025bizfinbench0} is a comprehensive financial benchmark covering 10 tasks, including financial analysis, financial news classification, financial text summarization, financial question answering, and financial named entity recognition. We evaluate the generalization ability of \textbf{RETuning} on BizFinBench. The details of each dataset type are as follows.

\begin{itemize}
    \item \textbf{Anomalous Event Attribution (AEA):} This dataset evaluates the model's ability to trace financial anomalies based on given information such as timestamps, news articles, financial reports, and stock movements. The model must identify the cause-and-effect relationships behind sudden market fluctuations and distinguish relevant factors from noise.
    \item \textbf{Financial Numerical Computation (FNC):} This dataset assesses the model's ability to perform accurate numerical calculations in financial scenarios, including interest rate calculations, return on investment (ROI), and financial ratios.
    \item \textbf{Financial Time Reasoning (FTR):} This dataset tests the model’s ability to understand and reason about time-based financial events, such as predicting interest accruals, identifying the impact of quarterly reports, and assessing financial trends over different periods.
    \item \textbf{Financial Tool Usage (FTU):} This dataset evaluates the model's ability to comprehend user queries and effectively use financial tools to solve real-world problems. It covers scenarios like investment analysis, market research, and information retrieval, requiring the model to select appropriate tools, input parameters accurately, and coordinate multiple tools when needed.
    \item \textbf{Financial Knowledge QA (FQA):} This dataset evaluates the model's understanding and response capabilities regarding core knowledge in the financial domain. It spans a wide range of financial topics, encompassing key areas such as fundamental financial concepts, financial markets, investment theory, macroeconomics, and finance.
    \item \textbf{Financial Data Description (FDD):} This dataset measures the model's ability to analyze and describe structured and unstructured financial data, such as balance sheets, stock reports, and financial statements.
    \item \textbf{Emotion Recognition (ER):} This dataset evaluates the model's capability to recognize nuanced user emotions in complex financial market environments. The input data encompasses multiple dimensions, including market conditions, news articles, research reports, user portfolio information, and queries. The dataset covers six distinct emotional categories: optimism, anxiety, negativity, excitement, calmness, and regret.
    \item \textbf{Stock Price Prediction (SP):} This dataset evaluates the model’s ability to predict future stock prices based on historical trends, financial indicators, and market news.
    \item \textbf{Financial Named Entity Recognition (FNER):} This dataset focuses on evaluating the model’s ability to identify and classify financial entities such as company names, stock symbols, financial instruments, regulatory agencies, and economic indicators.
\end{itemize}

Table~\ref{datasetall} presents a detailed breakdown of the dataset, covering the evaluation dimensions, corresponding metrics, the number of instances per task, and the average token length per entry. Notably, the dataset shows considerable variability in input length, spanning from a minimum of 22 tokens to a maximum of 4,556 tokens. This wide range not only mirrors the complexity and heterogeneity of real-world financial scenarios but also poses a meaningful challenge for models—specifically, in demonstrating their capability to process both short and long financial texts effectively.
Table~\ref{query_token_stats} presents their maximum token length, minimum token length, and average length.

\begin{table}[h]
  \centering
  \caption{Overview of BizFinBench~\citep{lu2025bizfinbench0} Datasets}
  \resizebox{\textwidth}{!}{%
    \begin{tabular}{cp{17.8em}p{14.445em}p{5.39em}ll}
    \toprule
    Category & \multicolumn{1}{l}{Data} & \multicolumn{1}{l}{Evaluation Dimensions} & \multicolumn{1}{l}{Metrics} & Numbers & Avg Len. \\
    \midrule
    Reasoning & Anomalous Event Attribution (AEA) & Causal consistency\newline{}Information relevance\newline{}Noise resistance & Accuracy & 1064  & 939 \\
          & Financial Time Reasoning (FTR) & Temporal reasoning correctness & Accuracy & 514   & 1162 \\
          & Financial Tool Usage (FTU) & Tool selection appropriateness\newline{}Parameter input accuracy\newline{}Multi-tool coordination & Judge Score & 641   & 4556 \\
    \midrule
    Numerical calculation & Financial Numerical Computation (FNC) & Computational accuracy\newline{}Unit consistency & Accuracy & 581   & 651 \\
    \midrule
    Q\&A  & Financial Knowledge QA (FQA) & Question comprehension\newline{}Knowledge coverage\newline{}Answer accuracy & Judge Score & 990   & 22 \\
          & Financial Data Description (FDD) & Trend accuracy\newline{}Data consistency & Judge Score & 1461  & 311 \\
    \midrule
    Prediction recognition & Emotion Recognition (ER) & Emotion classification accuracy\newline{}Implicit information extraction & Accuracy & 600   & 2179 \\
          & Stock Price Prediction (SP) & Trend judgment, Causal reasoning & Accuracy & 497   & 4498 \\
    \midrule
    Information extraction & Financial Named Entity Recognition (FNER) & Recognition accuracy\newline{}Entity classification correctness & Accuracy & 435   & 533 \\
    \bottomrule
    \end{tabular}%
    }
  \label{datasetall}%
\end{table}%

\begin{table}[htbp]
    \centering
    \caption{Token Length Statistics of BizFinBench~\citep{lu2025bizfinbench0}.}
    \label{query_token_stats}
    \begin{tabular}{lrrrr}
        \toprule
        \textbf{Dataset} & \textbf{Min} & \textbf{Max} & \textbf{Avg} & \textbf{Count} \\
        \midrule
        NER & 415 & 1,194 & 533.1 & 433 \\
        FTU & 4,169 & 6,289 & 4,555.5 & 641 \\
        AEA & 680 & 1,396 & 938.7 & 1,064 \\
        ER & 1,919 & 2,569 & 2,178.5 & 600 \\
        FNC & 287 & 2,698 & 650.5 & 581 \\
        FDD & 26 & 645 & 310.9 & 1,461 \\
        FTR & 203 & 8,265 & 1,162.0 & 514 \\
        FQA & 5 & 45 & 21.7 & 990 \\
        SP & 1,254 & 5,532 & 4,498.1 & 497 \\
        \bottomrule
    \end{tabular}
\end{table}

\section{Evaluation Details} \label{sec:appendix:evaluation_details}
We evaluate our model using several metrics, including accuracy, precision, recall, and F1-score. These metrics provide a comprehensive view of the model's performance across different aspects.

\subsection{Detailed Results on Fin-2024[December]} \label{sec:appendix:detailed_results:fin2024december}

\begin{figure*}[htbp]
	\centering
	\includegraphics[width=0.85\textwidth]{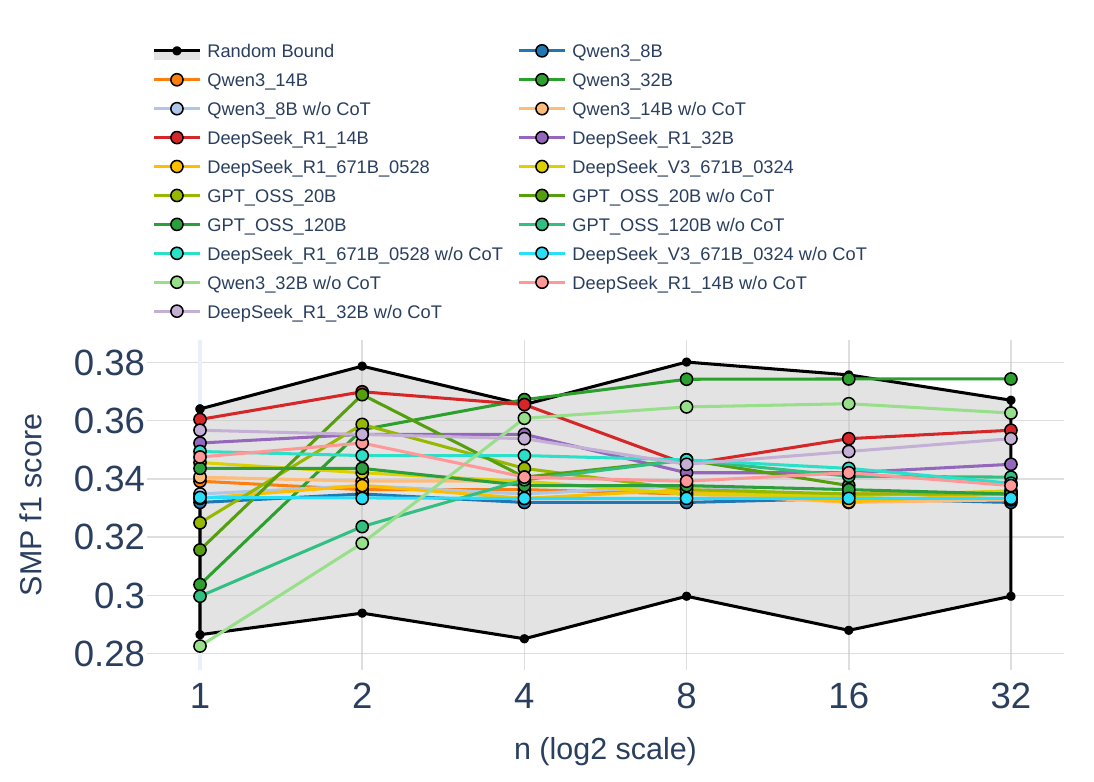}
	\caption{Detailed results on \textbf{Fin-2024[December]} for different baselines.}
	\label{fig:detailed_results:fin2024december}
\end{figure*}

We present detailed results on \textbf{Fin-2024[December]} for different baselines in Figure~\ref{fig:detailed_results:fin2024december}. The results are grouped by the number of repeated sampling counts \( n \) (log\(_2\) scale: 1, 2, 4, 8, 16, 32). Performance is measured using the stock movement prediction (SMP) F1 score. It can be observed that current LLMs struggle to achieve satisfactory performance on this challenging task. Most of them are not able to scale up their performance with increasing \( n \). In contrast, our proposed \textbf{RETuning} method demonstrates significant improvements, especially when combined with larger models and reinforcement learning techniques. Notably, the \textbf{DeepSeek\_R1\_32B\_SFT\_GRPO} model achieves the highest F1 score of approximately 0.44 at \( n=32 \), showcasing the effectiveness of our approach in enhancing model capabilities for stock movement prediction.

\subsection{Results Grouped by OOD split on Fin-2024[December] and Fin-2025[June]} \label{sec:appendix:detailed_results:group_by_ood}

This section analyzes the performance of different models on the \textbf{stock movement prediction (SMP) task}, a 3-category classification problem based on price changes, under four out-of-distribution (OOD) scenarios. The evaluation focuses on the models’ ability to \emph{scale up} by increasing the repeated sampling count \( n \) (log\(_2\) scale: 1, 2, 4, 8, 16, 32). Performance is measured using the SMP F1 score, and the comparison includes a baseline (Random Bound), the 14B/32B variants of DeepSeek-R1, and their optimized versions (SFT, SFT+GRPO).

\begin{minipage}[b]{0.49\textwidth}
    \centering
    \includegraphics[width=\textwidth]{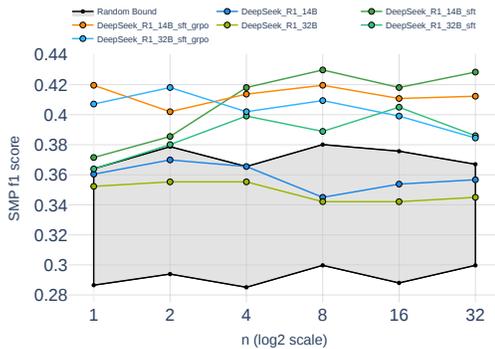}
    \captionof{figure}{\textbf{Overall} f1 score results on \textbf{Fin-2024[December]}.}
    \label{fig:appendix:overall_f1_score}
\end{minipage}
\hfill
\begin{minipage}[b]{0.49\textwidth}
    \centering
    \includegraphics[width=\textwidth]{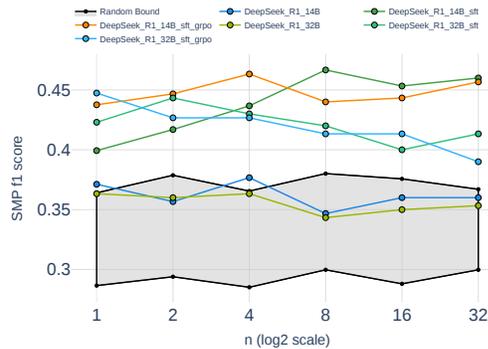}
    \captionof{figure}{\textbf{OOD\_Stock} f1 score results on \textbf{Fin-2024[December]}.}
    \label{fig:appendix:ood_stock_f1_score}
\end{minipage}

\begin{minipage}[b]{0.49\textwidth}
    \centering
    \includegraphics[width=\textwidth]{OOD_date_main_results.pdf}
    \captionof{figure}{\textbf{OOD\_Date} f1 score results on \textbf{Fin-2024[December]}.}
    \label{fig:appendix:ood_date_f1_score}
\end{minipage}
\hfill
\begin{minipage}[b]{0.49\textwidth}
    \centering
    \includegraphics[width=\textwidth]{OOD_date_stock_main_results.pdf}
    \captionof{figure}{\textbf{OOD\_Stock\&Date} f1 score results on \textbf{Fin-2024[December]}.}
    \label{fig:appendix:ood_stock_date_f1_score}
\end{minipage}

\begin{minipage}[b]{0.49\textwidth}
    \centering
    \includegraphics[width=\textwidth]{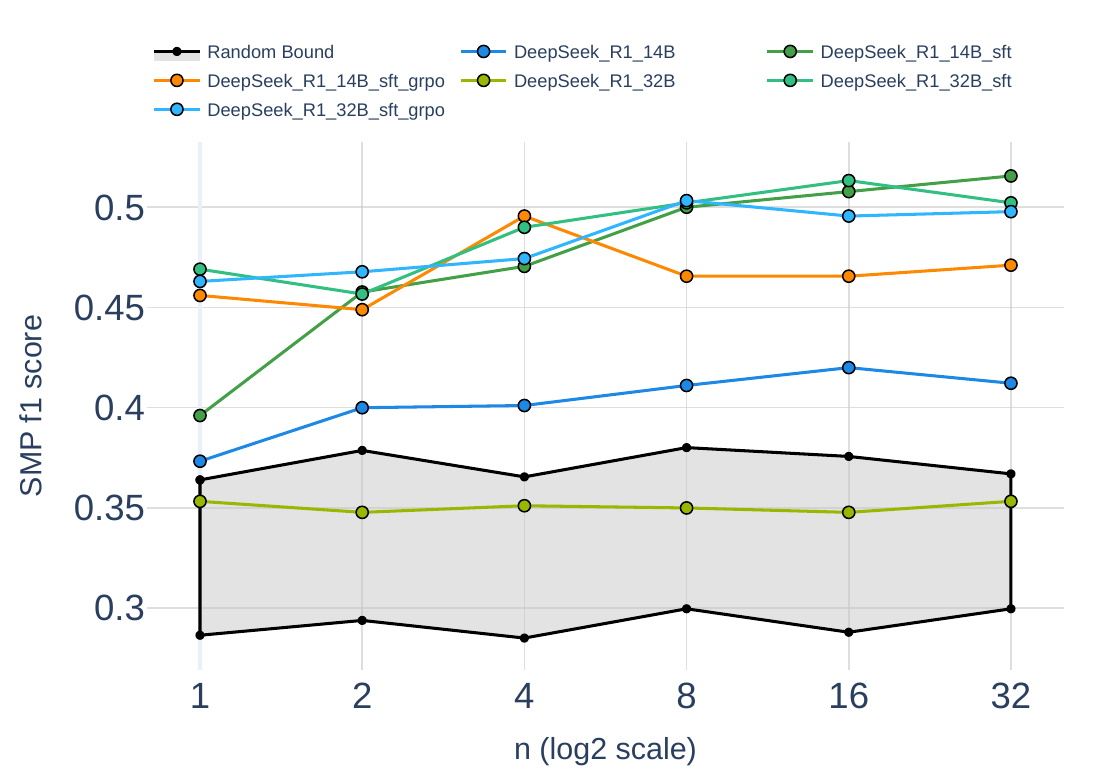}
    \captionof{figure}{\textbf{OOD\_Date} f1 score results on \textbf{Fin-2025[June]}.}
    \label{fig:appendix:ood_date_f1_score_June}
\end{minipage}
\hfill
\begin{minipage}[b]{0.49\textwidth}
    \centering
    \includegraphics[width=\textwidth]{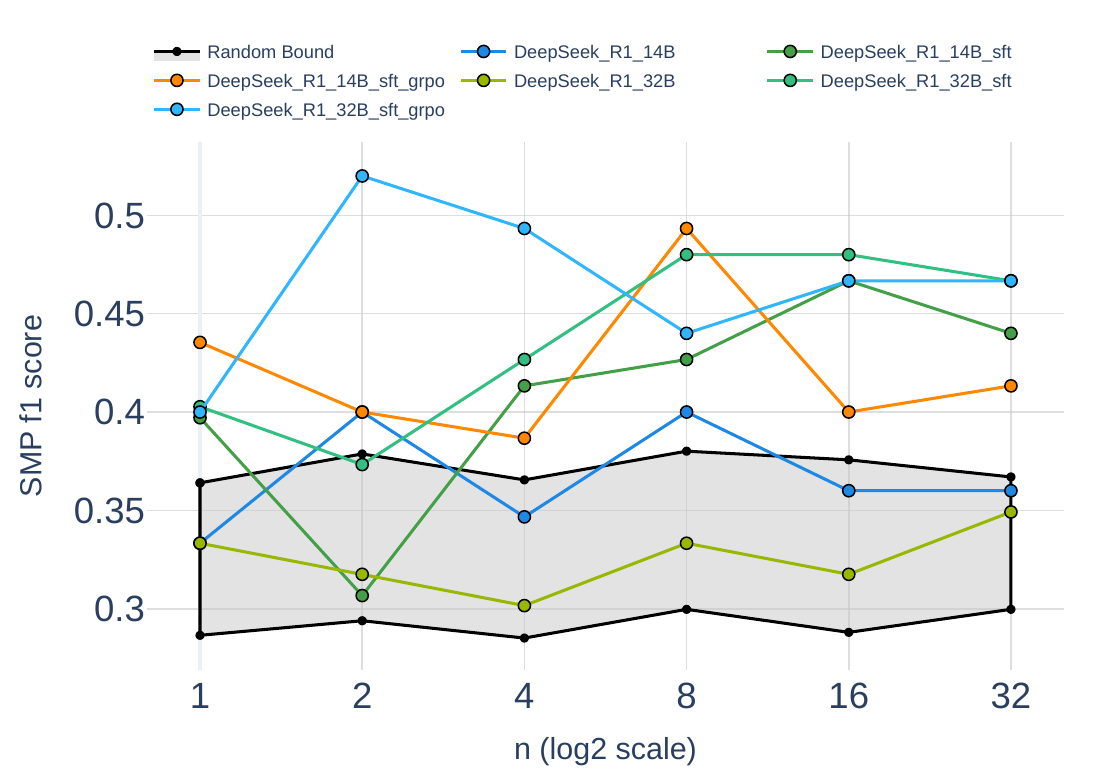}
    \captionof{figure}{\textbf{OOD\_Stock\&Date} f1 score results on \textbf{Fin-20254[June]}.}
    \label{fig:appendix:ood_stock_date_f1_score_June}
\end{minipage}

\subsubsection{Overall OOD Performance}
As shown in Figure~\ref{fig:appendix:overall_f1_score}, which aggregates results from the \textbf{OOD\_Stock}, \textbf{OOD\_Date}, and \textbf{OOD\_Stock\&Date} scenarios, two consistent trends emerge. First, a clear performance hierarchy is observed: 32B models consistently outperform their 14B counterparts, and optimization further improves results (SFT $>$ Base; SFT+GRPO $>$ SFT). Among all configurations, the \textbf{DeepSeek\_R1\_32B\_SFT\_GRPO} model achieves the highest F1 score of approximately 0.44 at \( n=32 \), followed by the 32B SFT and Base models. Importantly, all variants surpass the Random Bound baseline (F1 $<$ 0.30), demonstrating the benefits of both scaling and optimization. Second, all models exhibit \emph{monotonic gains} as \( n \) increases. For instance, DeepSeek\_R1\_32B\_SFT\_GRPO improves from about 0.36 at \( n=1 \) to 0.44 at \( n=32 \), corresponding to a relative gain of roughly 22\%. In contrast, the 14B baseline improves only from 0.28 to 0.34 over the same range. These results confirm that repeated sampling consistently enhances performance, with larger and optimized models benefiting the most.

\subsubsection{OOD\_Stock Scenario (Unseen Stocks)}
Figure~\ref{fig:appendix:ood_stock_f1_score} evaluates performance when predicting trends for \textbf{stocks unseen during training}. This setting achieves the second-highest peak performance across all scenarios, with DeepSeek\_R1\_32B\_SFT\_GRPO reaching about 0.45 at \( n=32 \). The performance gap between 32B and 14B optimized models widens to roughly 0.07 at this sampling level, underscoring the importance of parameter scale for generalizing to new stocks. Moreover, F1 scores rise sharply as \( n \) increases. For example, the 32B SFT model improves from 0.37 at \( n=1 \) to 0.43 at \( n=32 \), a gain of around 16\%. These findings suggest that repeated sampling effectively reduces noise in trend estimation for unfamiliar stocks, thereby mitigating the distribution shift.

\subsubsection{OOD\_Date Scenario (Unseen Time Periods)}
In the \textbf{OOD\_Date} setting (Figure~\ref{fig:appendix:ood_date_f1_score}), where models are tested on unseen time periods such as December 2024, performance is notably weaker. This scenario yields the lowest peak among all four settings, with the best-performing model (32B SFT+GRPO) reaching only about 0.42 at \( n=32 \). Even with maximum scaling, the gains are modest: DeepSeek\_R1\_32B\_SFT\_GRPO improves by only 0.08 (from 0.34 to 0.42) as \( n \) increases from 1 to 32, which is smaller than the gains observed in stock-based shifts. These results indicate that temporal distribution shifts, likely caused by market regime changes, are more difficult to address with repeated sampling alone.

\subsubsection{OOD\_Stock\&Date Scenario (Dual Distribution Shifts)}
The most challenging case, involving \textbf{both unseen stocks and unseen time periods}, is shown in Figure~\ref{fig:appendix:ood_stock_date_f1_score}. Surprisingly, this dual-shift setting achieves the \emph{highest overall peak performance}. DeepSeek\_R1\_32B\_SFT\_GRPO reaches approximately 0.50 at \( n=32 \), surpassing even the OOD\_Stock scenario. The scaling effect is particularly strong: the model improves by about 0.12 (from 0.38 to 0.50) when \( n \) increases from 1 to 32. Moreover, the performance gap between the 32B optimized model and the 14B baseline widens substantially with larger \( n \). These results highlight a synergy between model scale and repeated sampling, suggesting that larger models are especially capable of leveraging additional samples to disentangle stock-specific volatility from time-dependent shifts.

\subsubsection{Summary of Scaling-Up Ability Across OOD Scenarios}
Taken together, the results provide several key insights. First, repeated sampling proves to be a universally valid strategy, as all models improve with increasing \( n \) across every OOD setting. Second, the magnitude of gains is scenario-dependent: the largest improvements occur in the dual OOD\_Stock\&Date scenario (12--22\% increase), while the OOD\_Date scenario yields the smallest gains (8--15\%), reflecting the particular difficulty of temporal shifts. Third, the optimal configuration for scaling-up ability is the \textbf{DeepSeek\_R1\_32B\_SFT\_GRPO} model at \( n=32 \), which achieves F1 scores of roughly 0.50 (OOD\_Stock\&Date), 0.45 (OOD\_Stock), 0.44 (Overall OOD), and 0.42 (OOD\_Date), consistently outperforming all other settings by 5--15\%. Finally, the results underscore the \textbf{synergy between model scale, optimization, and repeated sampling}: larger models not only benefit more from optimization but also extract greater value from additional samples, with the effect most pronounced in dual-shift scenarios.


\subsection{Detailed Results on BizFinBench} \label{sec:appendix:detailed_results:bizfinbench}

We present the detailed results of various large language models on the BizFinBench benchmark~\citep{lu2025bizfinbench0}, as summarized in Table~\ref{table:bizfinbench_detailed_results}. The evaluation encompasses a range of tasks, including financial entity recognition, financial question answering, financial text classification, and more. The results are color-coded to highlight the top three performers for each task: \colorbox{golden}{golden} indicates the top-performing model, \colorbox{lightblue}{silver} represents the second-best result, and \colorbox{lightgreen}{bronze} denotes the third-best performance.

\begin{table*}[htbp]
  \centering
  \caption{Performance Comparison of Large Language Models on BizFinBench~\citep{lu2025bizfinbench0}. The models are evaluated across multiple tasks, with results color-coded to represent the top three performers for each task: \colorbox{golden}{golden} indicates the top-performing model, \colorbox{lightblue}{silver} represents the second-best result, and \colorbox{lightgreen}{bronze} denotes the third-best performance.}
  \resizebox{\textwidth}{!}{%
    \begin{tabular}{lrrrrrrrrrr}
    \toprule
    \multicolumn{1}{c}{Model} & \multicolumn{1}{c}{AEA} & \multicolumn{1}{c}{FNC} & \multicolumn{1}{c}{FTR} & \multicolumn{1}{c}{FTU} & \multicolumn{1}{c}{FQA} & \multicolumn{1}{c}{FDD} & \multicolumn{1}{c}{ER} & \multicolumn{1}{c}{SP} & \multicolumn{1}{c}{FNER}  & \multicolumn{1}{c}{Average}\\
    \midrule
    \multicolumn{10}{c}{Close-Source LLMs} \\
    ChatGPT-o3 & \cellcolor{lightblue}86.23 & 61.30     & \cellcolor{lightblue}75.36      & \cellcolor{golden}89.15    &   \cellcolor{lightblue}91.25    &  \cellcolor{lightgreen}98.55     & 44.48     & 53.27     & 65.13    & \cellcolor{golden}73.86 \\
    ChatGPT-o4-mini & \cellcolor{lightgreen}85.62 & 60.10     & 71.23      & 74.40    &   90.27    &  95.73     & \cellcolor{golden}47.67     & 52.32     & 64.24    & 71.29 \\
    GPT-4o & 79.42 & 56.51     & \cellcolor{golden}76.20      &   82.37    &   87.79    &  \cellcolor{golden}98.84     & 45.33     & 54.33     & 65.37    & \cellcolor{lightgreen}71.80 \\
    Gemini-2.0-Flash & \cellcolor{golden}86.94 & 62.67    & 73.97    &   82.55   &  90.29    & \cellcolor{lightblue}98.62   & 22.17     & \cellcolor{lightgreen}56.14     & 54.43    & 69.75 \\
    Claude-3.5-Sonnet & 84.68 & 63.18    & 42.81     &  \cellcolor{lightblue}88.05     &  87.35     &  96.85     & 16.67     & 47.60     & 63.09    & 65.59 \\
    \midrule
    \multicolumn{10}{c}{Open-Weight LLMs} \\
    Qwen2.5-7B-Instruct & 73.87 & 32.88     & 39.38     &  79.03     &  83.34     &   78.93    & 37.50     & 51.91     & 30.31    & 56.35 \\
    Qwen2.5-72B-Instruct & 69.27 & 54.28    & 70.72     &   85.29    &  87.79     & 97.43    & 35.33     & 55.13     & 54.02    & 67.70 \\
    Qwen2.5-VL-3B & 53.85 & 15.92     & 17.29     &  8.95     &  81.60     &  59.44    & 39.50     & 52.49     & 21.57    & 38.96 \\
    Qwen2.5-VL-7B & 73.87 & 32.71     &  40.24    &  77.85     &  83.94     &  77.41    & 38.83     & 51.91     & 33.40    & 56.68 \\
    Qwen2.5-VL-14B & 37.12   & 41.44     & 53.08     &   82.07    &  84.23     &  7.97    & 37.33     & 54.93     & 47.47    & 49.52 \\
    Qwen2.5-VL-32B & 76.79 & 50.00     & 62.16     &  83.57     &  85.30     &  95.95    & 40.50     & 54.93     & 68.36    & 68.62 \\
    Qwen2.5-VL-72B & 69.55 & 54.11     & 69.86     &  85.18     &  87.37     &  97.34    & 35.00     & 54.94     & 54.41    & 67.53 \\
    Qwen3-1.7B & 77.40 & 35.80 & 33.40 & 75.82 & 73.81 & 78.62 & 22.40 & 48.53 & 11.23 & 50.78 \\
    Qwen3-4B & 83.60 & 47.40 & 50.00 & 78.19 & 82.24 & 80.16 & 42.20 & 50.51 & 25.19 & 59.94 \\
    Qwen3-14B & 84.20 & 58.20 & 65.80 & 82.19 & 84.12 & 92.91 & 33.00 & 52.31 & 50.70 & 67.05 \\
    Qwen3-32B & 83.80 & 59.60 & 64.60 & 85.12 & 85.43 & 95.37 & 39.00 & 52.26 & 49.19 & 68.26 \\
    QwQ-32B & 84.02 & 52.91    & 64.90     &  84.81     &  89.60     &  94.20     & 34.50     & \cellcolor{lightblue}56.68     & 30.27    & 65.77 \\
    Xuanyuan3-70B & 12.14 & 19.69     & 15.41     &  80.89  &  86.51     &  83.90     & 29.83     & 52.62     & 37.33    & 46.48 \\
    Llama-3.1-8B-Instruct & 73.12 & 22.09    & 2.91      &  77.42     &  76.18     &  69.09     & 29.00     & 54.21     & 36.56    & 48.95 \\
    Llama-3.1-70B-Instruct & 16.26 & 34.25    & 56.34     &  80.64     &   79.97    &  86.90     & 33.33     & \cellcolor{golden}62.16     & 45.95    & 55.09 \\
    Llama 4 Scout & 73.60 & 45.80 & 44.20 & 85.02 & 85.21 & 92.32 & 25.60 & 55.76 & 43.00 & 61.17 \\
    DeepSeek-V3 (671B) & 74.34 & 61.82    & 72.60     &  86.54     &  \cellcolor{lightgreen}91.07     & 98.11      & 32.67     & 55.73     & \cellcolor{lightblue}71.24    & 71.57 \\
    DeepSeek-R1 (671B) & 80.36 & \cellcolor{lightgreen}64.04   & \cellcolor{lightgreen}75.00     &  81.96     &  \cellcolor{golden}91.44     & 98.41      & 39.67     & 55.13     &  \cellcolor{golden}71.46   & \cellcolor{lightblue}73.05 \\
    DeepSeek\_R1\_14B\_Instruct & 71.33 & 44.35 & 50.45     & 81.96    & 85.52     & 92.81   & 39.50     & 50.20     & 52.76    & 59.49 \\
    DeepSeek\_R1\_32B\_Instruct & 73.68 & 51.20 & 50.86     & 83.27     & 87.54     & 97.81     & 41.50     & 53.92     & 56.80    & 66.29 \\
    \midrule
    \multicolumn{11}{c}{Our LLMs} \\
    \cellcolor{Background1}DeepSeek\_R1\_14B\_SFT~\tnote{2}       & \cellcolor{Background1}80.63 & \cellcolor{Background1}51.67 & \cellcolor{Background1}52.61 & \cellcolor{Background1}83.53 & \cellcolor{Background1}89.05 & \cellcolor{Background1}96.72 & \cellcolor{Background1}36.68 & \cellcolor{Background1}50.43 & \cellcolor{Background1}50.85 & \cellcolor{Background1}65.36 \\
    $\quad$ 14B $\Delta_{\text{Instruct}}(\text{SFT})$            & \textcolor{GREEN}{+9.25} & \textcolor{GREEN}{+7.28} & \textcolor{GREEN}{+2.19} & \textcolor{GREEN}{+1.53} & \textcolor{GREEN}{+3.42} & \textcolor{GREEN}{+3.85} & \textcolor{RED}{-2.93} & \textcolor{GREEN}{+0.24} & \textcolor{RED}{-1.92} & \textcolor{GREEN}{+5.83} \\
    \cellcolor{Background1}DeepSeek\_R1\_14B\_SFT\_GRPO~\tnote{2} & \cellcolor{Background1}81.46 & \cellcolor{Background1}52.41 & \cellcolor{Background1}53.47 & \cellcolor{Background1}83.57 & \cellcolor{Background1}89.02 & \cellcolor{Background1}95.58 & \cellcolor{Background1}36.83 & \cellcolor{Background1}54.06 & \cellcolor{Background1}51.24 & \cellcolor{Background1}66.92 \\
    $\quad$ 14B $\Delta_{\text{Instruct}}(\text{SFT\_GRPO})$      & \textcolor{GREEN}{+10.03} & \textcolor{GREEN}{+8.09} & \textcolor{GREEN}{+2.91} & \textcolor{GREEN}{+1.64} & \textcolor{GREEN}{+3.45} & \textcolor{GREEN}{+2.63} & \textcolor{RED}{-2.74} & \textcolor{GREEN}{+3.82} & \textcolor{RED}{-1.53} & \textcolor{GREEN}{+7.46} \\
    $\quad$ 14B $\Delta_{\text{SFT}}(\text{SFT\_GRPO})$           & \textcolor{GREEN}{+0.82} & \textcolor{GREEN}{+0.85} & \textcolor{GREEN}{+0.81} & \textcolor{GREEN}{+0.06} & 0.00 & \textcolor{RED}{-1.23} & \textcolor{GREEN}{+0.25} & \textcolor{GREEN}{+3.63} & \textcolor{GREEN}{+0.42} & \textcolor{GREEN}{+1.53} \\
    \cellcolor{Background1}DeepSeek\_R1\_32B\_SFT                 & \cellcolor{Background1}80.45 & \cellcolor{lightblue}66.42 & \cellcolor{Background1}63.28 & \cellcolor{lightgreen}86.88 & \cellcolor{Background1}88.43 & \cellcolor{Background1}93.76 & \cellcolor{lightblue}46.05 & \cellcolor{Background1}55.27 & \cellcolor{lightgreen}68.41 & \cellcolor{Background1}70.08 \\
    $\quad$ 32B $\Delta_{\text{Instruct}}(\text{SFT})$            & \textcolor{GREEN}{+6.75} & \textcolor{GREEN}{+15.23} & \textcolor{GREEN}{+12.37} & \textcolor{GREEN}{+3.64} & \textcolor{GREEN}{+0.83} & \textcolor{RED}{-4.14} & \textcolor{GREEN}{+4.52} & \textcolor{GREEN}{+1.25} & \textcolor{GREEN}{+11.63} & \textcolor{GREEN}{+3.75} \\
    \cellcolor{Background1}DeepSeek\_R1\_32B\_SFT\_GRPO           & \cellcolor{Background1}80.67 & \cellcolor{golden}66.83 & \cellcolor{Background1}64.45 & \cellcolor{Background1}86.79 & \cellcolor{Background1}88.52 & \cellcolor{Background1}91.26 & \cellcolor{lightgreen}45.68 & \cellcolor{Background1}54.83 & \cellcolor{Background1}67.75 & \cellcolor{Background1}70.44 \\
    $\quad$ 32B $\Delta_{\text{Instruct}}(\text{SFT\_GRPO})$      & \textcolor{GREEN}{+6.95} & \textcolor{GREEN}{+15.62} & \textcolor{GREEN}{+13.57} & \textcolor{GREEN}{+3.55} & \textcolor{GREEN}{+0.93} & \textcolor{RED}{-6.64} & \textcolor{GREEN}{+4.13} & \textcolor{GREEN}{+0.85} & \textcolor{GREEN}{+10.93} & \textcolor{GREEN}{+4.12} \\
    $\quad$ 32B $\Delta_{\text{SFT}}(\text{SFT\_GRPO})$           & \textcolor{GREEN}{+0.23} & \textcolor{GREEN}{+0.42} & \textcolor{GREEN}{+1.23} & \textcolor{RED}{-0.06} & \textcolor{GREEN}{+0.13} & \textcolor{RED}{-2.53} & \textcolor{RED}{-0.42} & \textcolor{RED}{-0.43} & \textcolor{RED}{-0.72} & \textcolor{GREEN}{+0.33} \\
    \bottomrule
    \end{tabular}%
    }
  \label{table:bizfinbench_detailed_results}%
\end{table*}%

\section{Case Study} \label{sec:appendix:case_study}

\subsection{Case study on zero-shot CoT prompting} \label{sec:appendix:case_study:cot_prompting}

\begin{figure*}[htbp]
	\centering
	\includegraphics[width=1\textwidth]{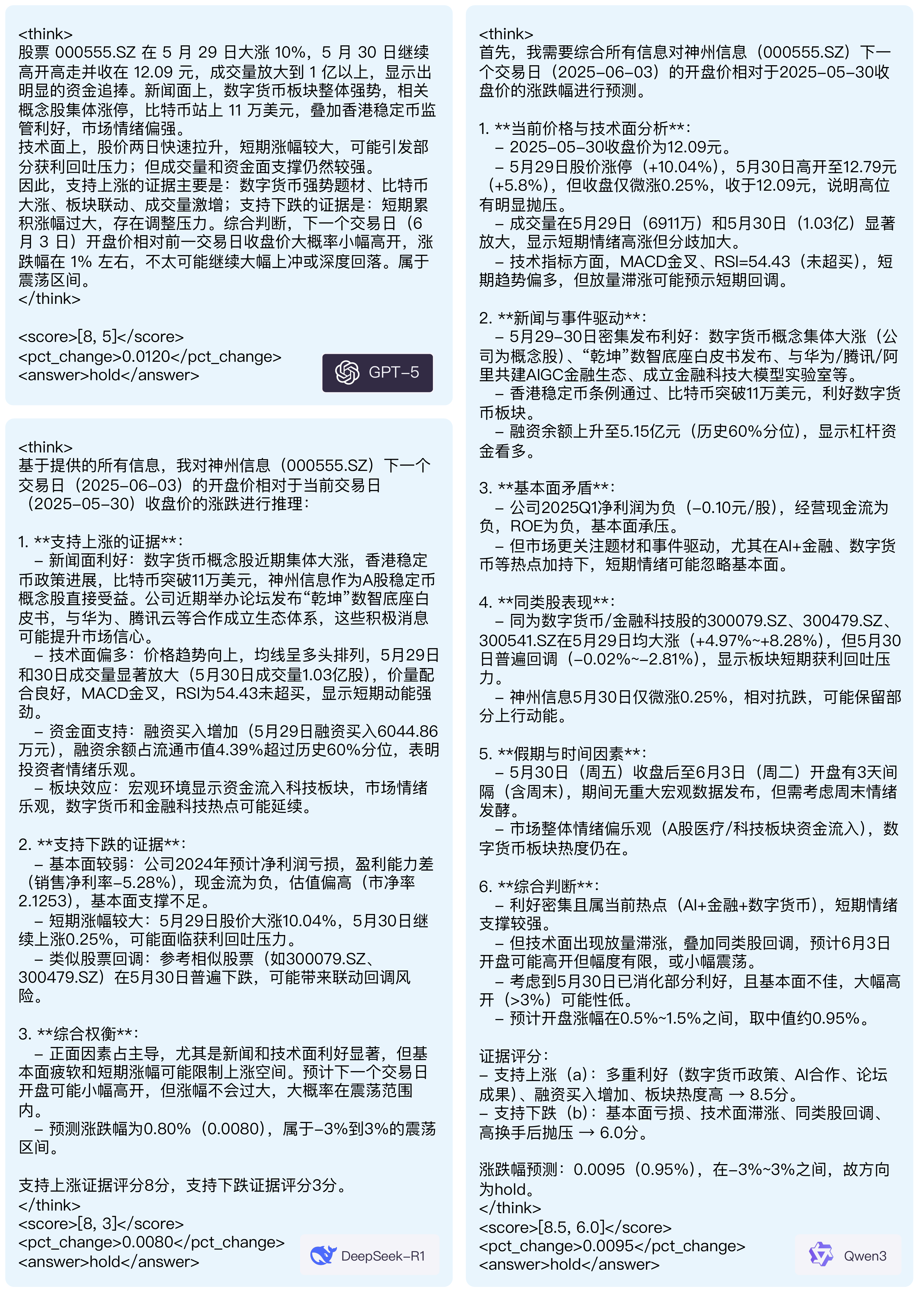}
	\caption{Zero-shot responses. The model struggles to provide accurate and relevant answers without the step-by-step reasoning process enabled by CoT prompting.}
	\label{fig:appendix:case_study:response_no_cot}
\end{figure*}

\begin{figure*}[htbp]
	\centering
	\includegraphics[width=1\textwidth]{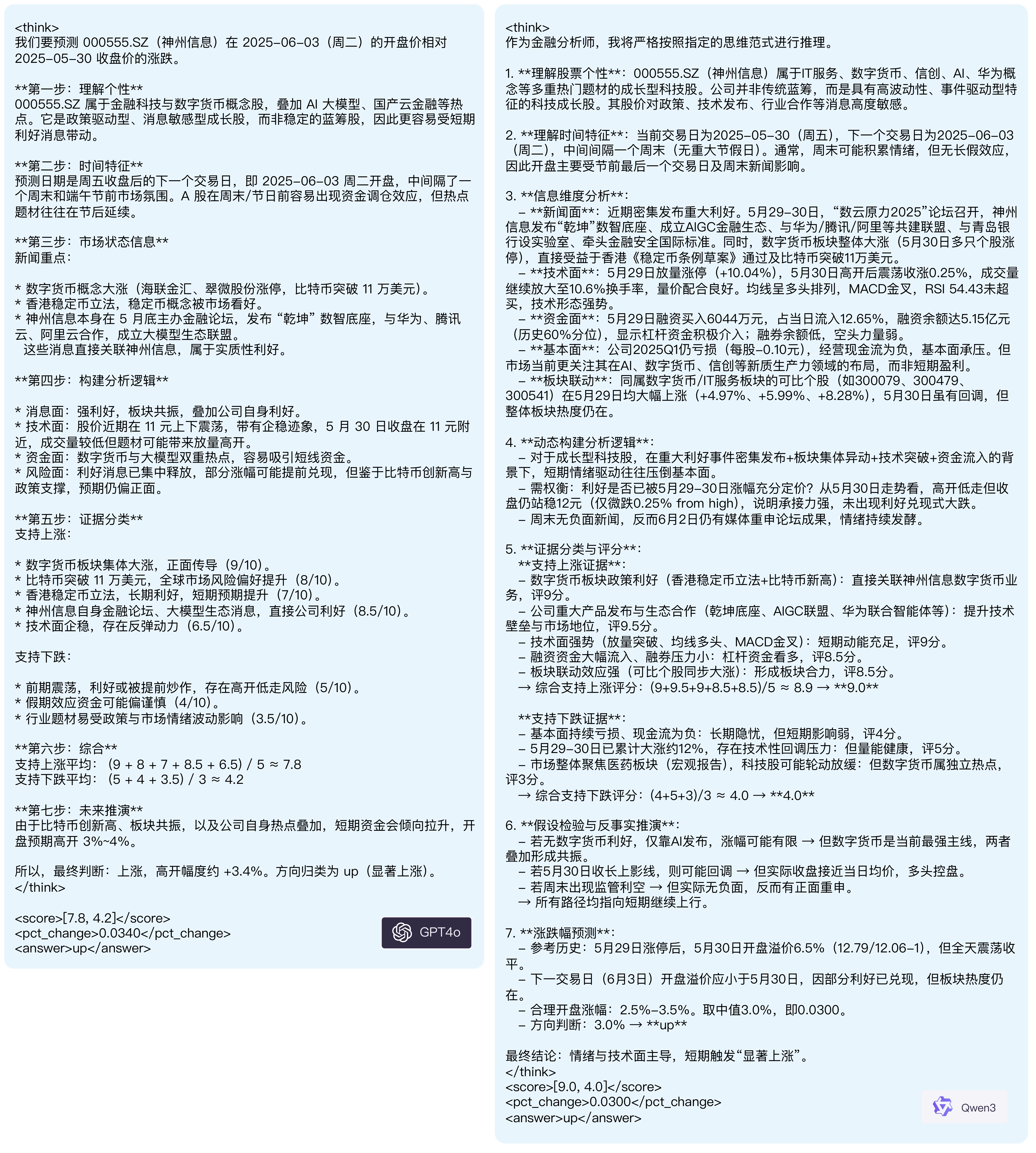}
	\caption{CoT Prompting responses. The model demonstrates improved reasoning and accuracy when guided through a step-by-step thought process.}
	\label{fig:appendix:case_study:response_with_cot}
\end{figure*}

We present a case study comparing the responses of GPT4o, DeepSeek-R1 and Qwen3 under two different prompting strategies: zero-shot prompting without Chain-of-Thought (CoT) and zero-shot prompting with CoT. The results are illustrated in Figures~\ref{fig:appendix:case_study:response_no_cot} and~\ref{fig:appendix:case_study:response_with_cot}.

We observe that without CoT prompting (Figure~\ref{fig:appendix:case_study:response_no_cot}), the models struggle to provide accurate and relevant answers. Their responses often lack depth and fail to address the nuances of the questions. In contrast, with CoT prompting (Figure~\ref{fig:appendix:case_study:response_with_cot}), the models demonstrate improved reasoning and accuracy. The step-by-step thought process enabled by CoT helps the models to break down complex questions and arrive at more precise answers.

\subsection{Case study on cold-started responses} \label{sec:appendix:case_study:response}

\begin{figure*}[htbp]
	\centering
	\includegraphics[width=1\textwidth]{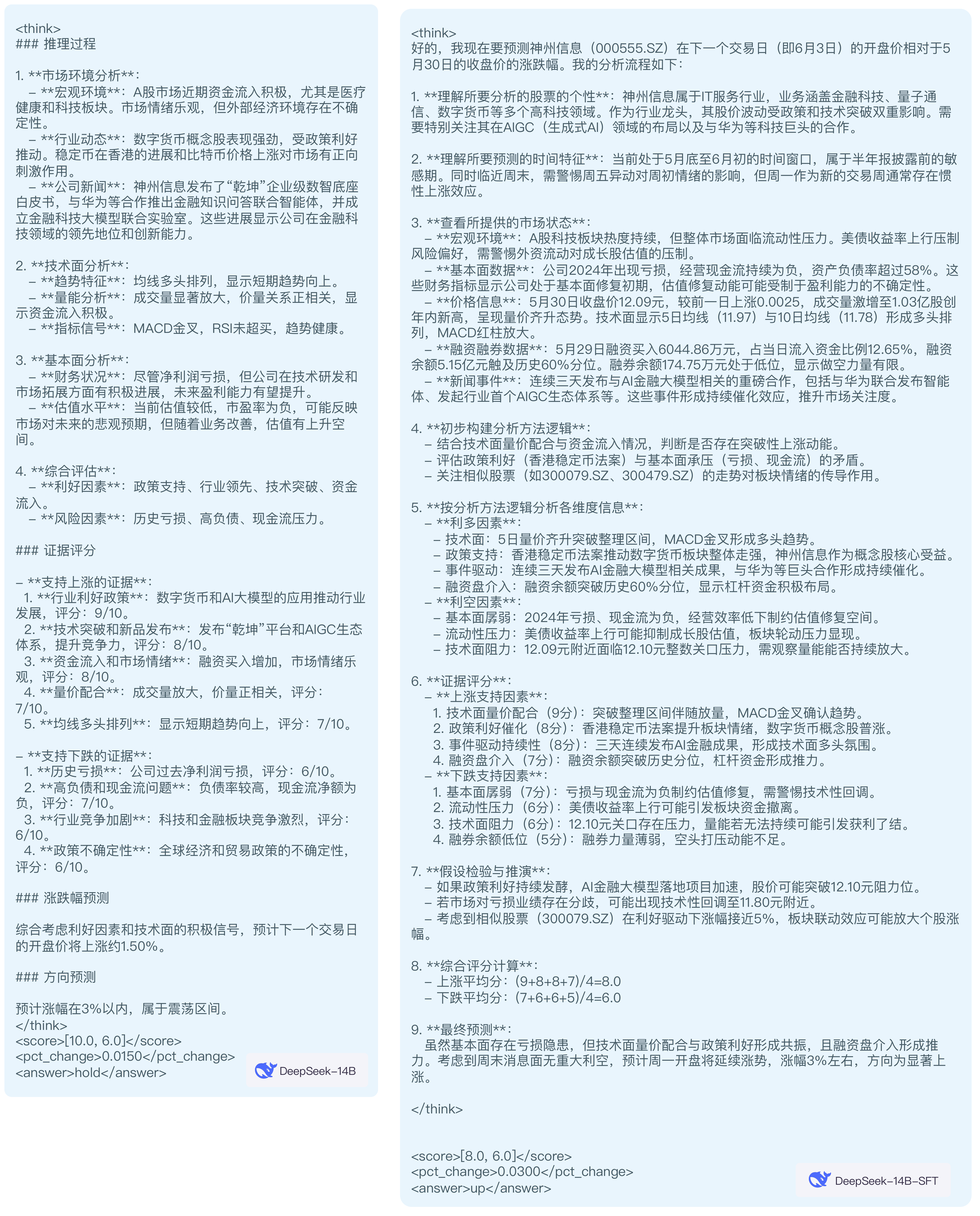}
	\caption{Case study on cold-started responses. Left is the original response, and right is the cold-started response. The cold-started response is more concise and to the point, avoiding unnecessary elaboration.}
	\label{fig:appendix:case_study:response_cold_started}
\end{figure*}

We present a case study comparing the original response and the cold-started response generated by DeepSeek\_R1\_14B\_Instruct and DeepSeek\_R1\_14B\_SFT. The results are illustrated in Figure~\ref{fig:appendix:case_study:response_cold_started}.
Before applying cold-starting techniques, the original response tends to be verbose and includes unnecessary elaboration. After cold-starting, the response becomes more concise and is much longer, which indicates the cold-started model is leveraging its reasoning capabilities to provide a more comprehensive answer.
This demonstrates the effectiveness of cold-starting in enhancing the clarity and relevance of model-generated responses.

\end{document}